\documentclass[lettersize,journal]{IEEEtran}

\usepackage{lineno,hyperref}
\usepackage{amsmath}
\usepackage{amsfonts}
\usepackage{amssymb}
\usepackage{algorithmic}
\usepackage{algorithm}
\usepackage{array}
\usepackage[english]{babel}
\usepackage{balance}
\usepackage[justification=centering]{caption}
\usepackage{colortbl}
\usepackage{csquotes}
\usepackage{enumitem}
\usepackage{graphicx}
\usepackage[utf8x]{inputenc}
\usepackage{multirow}
\usepackage{pgfplots}
\usepgfplotslibrary{fillbetween}
\pgfplotsset{compat=newest}
\usepgfplotslibrary{statistics} 
\usepackage{pgfplotstable}
\usepackage{seqsplit}
\usepackage{stfloats}
\usepackage{subcaption}
\usepackage{tabularx}
\usepackage{textcomp}
\usepackage{tikz}
\usetikzlibrary{arrows,arrows.meta,chains,decorations.markings,patterns,positioning,shapes.geometric,scopes,shadows,shadows.blur,trees}
\usepackage{url}
\usepackage{verbatim}
\usepackage{xcolor}

\def\BibTeX{{\rm B\kern-.05em{\sc i\kern-.025em b}\kern-.08em
    T\kern-.1667em\lower.7ex\hbox{E}\kern-.125emX}}

\newcommand\algorithmicprocedure{\textbf{procedure}}
\newcommand{\algorithmicendprocedure}{\algorithmicend\ \algorithmicprocedure}
\makeatletter
\newcommand\PROCEDURE[3][default]{%
	\ALC@it
	\algorithmicprocedure\ \textsc{#2}(#3)%
	\ALC@com{#1}%
	\begin{ALC@prc}%
	}
	\newcommand\ENDPROCEDURE{%
	\end{ALC@prc}%
	\ifthenelse{\boolean{ALC@noend}}{}{%
		\ALC@it\algorithmicendprocedure
	}%
}
\newenvironment{ALC@prc}{\begin{ALC@g}}{\end{ALC@g}}
\makeatother

\begin{document}

\title{From Google Gemini to OpenAI Q* (Q-Star): A Survey of Reshaping the Generative Artificial Intelligence (AI) Research Landscape}
\author{Timothy R. McIntosh, Teo Susnjak, Tong Liu, Paul Watters, \IEEEmembership{Senior Member, IEEE}, and Malka N. Halgamuge, \IEEEmembership{Senior Member, IEEE}
\thanks{Manuscript received \protect\today. \textit{(Corresponding author: Timothy R. McIntosh.)} }

\thanks{Timothy McIntosh is with Academies Australasia Polytechnic, Melbourne, VIC 3000, Australia (e-mail: t.mcintosh@aapoly.edu.au).}
\thanks{Teo Susnjak and Tong Liu are with Massey University, Auckland 0632, New Zealand (e-mail: t.liu@massey.ac.nz; t.susnjak@massey.ac.nz).}
\thanks{Paul Watters is with Cyberstronomy Pty Ltd, Ballarat, VIC 3350, Australia (e-mail: ceo@cyberstronomy.com).}
\thanks{Malka N. Halgamuge is with RMIT University, Melbourne, VIC 3000, Australia (e-mail: malka.halgamuge@rmit.edu.au).}
}

\markboth{Journal of \LaTeX\ Class Files,~Vol.~1, No.~1, December~2023}%
{McIntosh \MakeLowercase{\textit{et al.}}: From Google Gemini to OpenAI Q* (Q-Star)}

\maketitle

\begin{abstract}
	This comprehensive survey explored the evolving landscape of generative Artificial Intelligence (AI), with a specific focus on the transformative impacts of Mixture of Experts (MoE), multimodal learning, and the speculated advancements towards Artificial General Intelligence (AGI). It critically examined the current state and future trajectory of generative Artificial Intelligence (AI), exploring how innovations like Google's Gemini and the anticipated OpenAI Q* project are reshaping research priorities and applications across various domains, including an impact analysis on the generative AI research taxonomy. It assessed the computational challenges, scalability, and real-world implications of these technologies while highlighting their potential in driving significant progress in fields like healthcare, finance, and education. It also addressed the emerging academic challenges posed by the proliferation of both AI-themed and AI-generated preprints, examining their impact on the peer-review process and scholarly communication. The study highlighted the importance of incorporating ethical and human-centric methods in AI development, ensuring alignment with societal norms and welfare, and outlined a strategy for future AI research that focuses on a balanced and conscientious use of MoE, multimodality, and AGI in generative AI.
\end{abstract}



\section{Introduction}
\label{sec:Introduction}
\IEEEPARstart{T}{he} historical context of AI, tracing back to Alan Turing's \enquote{Imitation Game} \cite{turing1950computing}, early computational theories \cite{mcdermott1976artificial,minsky1961steps}, and the development of the first neural networks and machine learning \cite{lecun2015deep,minsky1969introduction,rumelhart1986learning}, has set the foundation for today's advanced models. This evolution, accentuated by crucial moments such as the rise of deep learning and reinforcement learning, has been vital in shaping the contemporary trends in AI, including the sophisticated Mixture of Experts (MoE) models and multimodal AI systems, illustrating the field's dynamic and continuously evolving character. These advancements are a testament to the dynamic and ever-evolving nature of AI technology. The evolution of Artificial Intelligence (AI) has witnessed a crucial turn with the advent of Large Language Models (LLMs), notably ChatGPT, developed by OpenAI, and the recent unveiling of Google's Gemini \cite{lee2023multimodality,maddigan2023chat2vis}. This technology has not only revolutionized the industry and academia, but has also reignited critical discussions concerning AI consciousness and its potential threats to humanity \cite{mcintosh2023culturally,morris2023levels,schuett2023towards}. The development of such advanced AI systems, including notable competitors like Anthropic's Claude, and now Gemini, which demonstrates several advances over previous models like GPT-3 and Google's own LaMDA, has reshaped the research landscape. Gemini's ability to learn from two-way conversations and its \enquote{spike-and-slab} attention method, which allows it to focus on relevant parts of the context during multi-turn conversations, represents a significant leap in developing models that are better equipped for multidomain conversational applications\footnote{https://deepmind.google/technologies/gemini/}. These innovations in LLMs, including the mixture-of-experts methods employed by Gemini, signal a move towards models that can handle a diversity of inputs and foster multimodal approaches. Amidst this backdrop, speculations of an OpenAI project known as Q* (Q-Star) have surfaced, allegedly combining the power of LLMs with sophisticated algorithms such as Q-learning and A* (A-Star algorithm), further contributing to the dynamic research environment\footnote{https://www.forbes.com/sites/lanceeliot/2023/11/26/about-that-mysterious-ai-breakthrough-known-as-q-by-openai-that-allegedly-attains-true-ai-or-is-on-the-path-toward-artificial-general-intelligence-agi}.

\subsection{Changing AI Research Popularity}
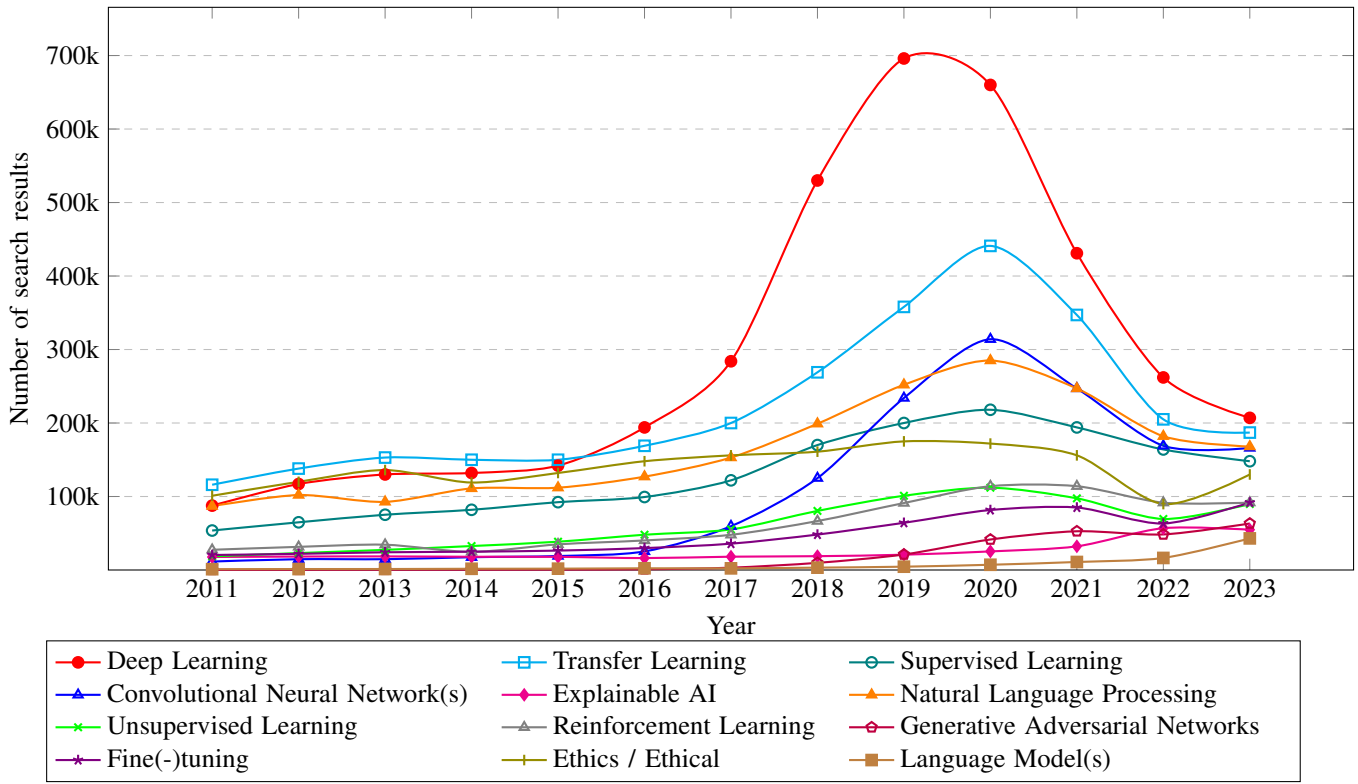
\begin{figure*}[t!]
	\begin{tikzpicture}
		\centering
		\normalsize
		\begin{axis}[
			clip=false,
			xlabel=Year,
			ymajorgrids, tick align=inside,
			major grid style=dashed,
			height=0.5*\textwidth, 
			width=\textwidth,
			legend pos=north west,
			legend style={cells={align=left}},
			xtick={2011,2012,2013,2014,2015,2016,2017,2018,2019,2020,2021,2022,2023},	
			xticklabels={2011,2012,2013,2014,2015,2016,2017,2018,2019,2020,2021,2022,2023},
			ymin=0,
			ytick={100000,200000,300000,400000,500000,600000,700000},	
			yticklabels={100k,200k,300k,400k,500k,600k,700k},
			scaled y ticks=false,
			legend cell align=left,
			legend style={
				cells={align=left},
				at={(-0.05,-0.125)},
				anchor=north west,
				draw=black,
				fill=white,
				align=left,
				legend columns=3 
			},
			ylabel={Number of search results}]

			\addplot[smooth,thick,color=red,mark=*] coordinates {
				(2011,87600) (2012,117000) (2013,130000) (2014,132000) (2015,142000) (2016,194000) (2017,284000) (2018,530000) (2019,696000) (2020,660000) (2021,431000) (2022,262000) (2023,207000)
			};\addlegendentry{Deep Learning~~~}
			
			\addplot[smooth,thick,color=cyan,mark=square] coordinates {
				 (2011,116000) (2012,138000) (2013,153000) (2014,150000) (2015,150000) (2016,169000) (2017,200000) (2018,269000) (2019,358000) (2020,441000) (2021,347000) (2022,205000) (2023,187000)
			};\addlegendentry{Transfer Learning~~~}
			
			\addplot[smooth,thick,color=teal,mark=o] coordinates {
				 (2011,53700) (2012,64900) (2013,75200) (2014,81900) (2015,92300) (2016,99300) (2017,122000) (2018,170000) (2019,200000) (2020,218000) (2021,194000) (2022,164000) (2023,148000)
			};\addlegendentry{Supervised Learning~~~}
			
			\addplot[smooth,thick,color=blue,mark=triangle] coordinates {
				 (2011,11500) (2012,14700) (2013,14700) (2014,17400) (2015,19000) (2016,25300) (2017,59600) (2018,125000) (2019,234000) (2020,314000) (2021,247000) (2022,169000) (2023,166000)
			};\addlegendentry[align=left]{Convolutional Neural Network(s)~~~}
			
			\addplot[smooth,thick,color=magenta,mark=diamond*] coordinates {
				 (2011,17700) (2012,18300) (2013,18500) (2014,18000) (2015,17900) (2016,16200) (2017,18100) (2018,18800) (2019,20800) (2020,25300) (2021,32200) (2022,56900) (2023,54800)
			};\addlegendentry[align=left]{Explainable AI~~~}
					
			\addplot[smooth, thick, color=orange, mark=triangle*] coordinates {
				(2011,86900) (2012,102000) (2013,92700) (2014,111000) (2015,112000) (2016,127000) (2017,153000) (2018,199000) (2019,252000) (2020,285000) (2021,247000) (2022,182000) (2023,168000)
			};\addlegendentry{Natural Language Processing~~~}
			
			\addplot[smooth, thick, color=green, mark=x] coordinates {
				 (2011,18600) (2012,23000) (2013,27400) (2014,32600) (2015,38500) (2016,48000) (2017,55200) (2018,80400) (2019,101000) (2020,112000) (2021,97500) (2022,69400) (2023,89800)
			};\addlegendentry{Unsupervised Learning~~~}
			
			\addplot[smooth, thick, color=gray, mark=triangle] coordinates {
				(2011,27500) (2012,31500) (2013,34400) (2014,24700) (2015,34900) (2016,40200) (2017,47900) (2018,66600) (2019,91100) (2020,114000) (2021,114000) (2022,91600) (2023,91700)
			};
			\addlegendentry{Reinforcement Learning~~~}
			
			\addplot[smooth, thick, color=purple, mark=pentagon] coordinates {
			(2011,242) (2012,293) (2013,352) (2014,421) (2015,523) (2016,970) (2017,3170) (2018,9720) (2019,21100) (2020,41500) (2021,52700) (2022,48400) (2023,63300)
			};\addlegendentry{Generative Adversarial Networks~~~}

			\addplot[smooth, thick, color=violet, mark=star] coordinates {
				(2011,20400) (2012,22000) (2013,24100) (2014,24900) (2015,26400) (2016,29800) (2017,35800) (2018,48200) (2019,64200) (2020,81600) (2021,85000) (2022,63200) (2023,91800)
			};
			\addlegendentry{Fine(-)tuning~~~}
			
			\addplot[smooth, thick, color=olive, mark=|] coordinates {
				(2011,101000) (2012,120000) (2013,136000) (2014,119000) (2015,132000) (2016,148000) (2017,156000) (2018,161000) (2019,175000) (2020,172000) (2021,156000) (2022,89700) (2023,130000)
			};
			\addlegendentry{Ethics / Ethical~~~}
			
			\addplot[smooth, thick, color=brown, mark=square*] coordinates {
				(2011,1090) (2012,1300) (2013,1320) (2014,1670) (2015,1810) (2016,2320) (2017,2240) (2018,3200) (2019,4500) (2020,7130) (2021,10900) (2022,16400) (2023,43000)
			};
			\addlegendentry{Language Model(s)~~~}

		\end{axis}
	\end{tikzpicture}
	\caption{Number of search results on Google Scholar with different keywords by year \protect\footnotemark}
	\label{fig:SearchResults}
\end{figure*}

As the field of LLMs continues to evolve, exemplified by innovations such as Gemini and Q*, a multitude of studies have surfaced with the aim of charting future research paths, which have varied from identifying emerging trends to highlighting areas poised for swift progress. The dichotomy of established methods and early adoption is evident, with \enquote{hot topics} in LLM research increasingly shifting towards multimodal capabilities and conversation-driven learning, as demonstrated by Gemini. The propagation of preprints has expedited knowledge sharing, but also brings the risk of reduced academic scrutiny. Issues like inherent biases, noted by Retraction Watch, along with concerns about plagiarism and forgery, present substantial hurdles \cite{shuai2017multidimensional}. The academic world, therefore, stands at an intersection, necessitating a unified drive to refine research directions in light of the fast-paced evolution of the field, which appears to be partly traced through the changing popularity of various research keywords over time. The release of generative models like GPT and the widespread commercial success of ChatGPT have been influential. As depicted in Figure~\ref{fig:SearchResults}, the rise and fall of certain keywords appear to have correlated with significant industry milestones, such as the release of the ``Transformer'' model in 2017 \cite{vaswani2017attention}, the GPT model in 2018 \cite{radford2018improving}, and the commercial ChatGPT-3.5 in December 2022. For instance, the spike in searches related to \enquote{Deep Learning} coincides with the breakthroughs in neural network applications, while the interest in \enquote{Natural Language Processing} surges as models like GPT and LLaMA redefine what's possible in language understanding and generation. The enduring attention to \enquote{Ethics / Ethical} in AI research, despite some fluctuations, reflects the continuous and deep-rooted concern for the moral dimensions of AI, underscoring that ethical considerations are not merely a reactionary measure, but an integral and persistent dialogue within the AI discussion \cite{huang2022overview}. 

It is academically intriguing to postulate whether these trends signify a causal relationship, where technological advancements drive research focus, or if the burgeoning research itself propels technological development. This paper also explores the profound societal and economic impacts of AI advancements. We examine how AI technologies are reshaping various industries, altering employment landscapes, and influencing socio-economic structures. This analysis highlights both the opportunities and challenges posed by AI in the modern world, emphasizing its role in driving innovation and economic growth, while also considering the ethical implications and potential for societal disruption. Future studies could yield more definitive insights, yet the synchronous interplay between innovation and academic curiosity remains a hallmark of AI's progress.

\footnotetext{The legend entries correspond to the keywords used in the search query, which is constructed as: \textit{\enquote{(AI OR artificial OR (machine learning) OR (neural network) OR computer OR software) AND ([specific keyword])}}.}

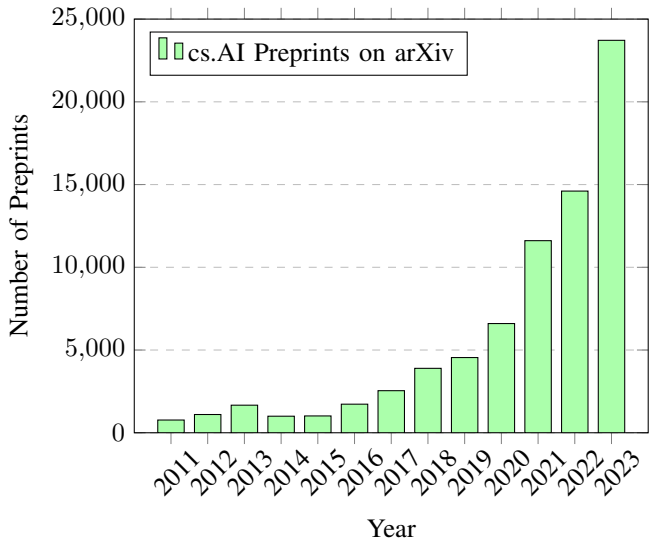
\begin{figure}[t!]
	\begin{tikzpicture}
		\begin{axis}[
			ybar,
			xlabel={Year},
			ylabel={Number of Preprints},
			xmin=2010, xmax=2024,
			ymin=0, 
			ymax=25000, 
			xtick={2011,2012,2013,2014,2015,2016,2017,2018,2019,2020,2021,2022,2023},
			xticklabels={2011,2012,2013,2014,2015,2016,2017,2018,2019,2020,2021,2022,2023},
			x tick label style={rotate=45, anchor=center,xshift=-6pt,yshift=-9pt},
			scaled y ticks=false,
			legend pos=north west,
			ymajorgrids=true,
			grid style=dashed,
			width=0.95\columnwidth,
			height=0.8\columnwidth
			]
			\addplot[fill=green!33] coordinates {
				(2011,770) 
				(2012,1103)
				(2013,1667)
				(2014,1001)
				(2015,1017)
				(2016,1728)
				(2017,2543)
				(2018,3894)
				(2019,4545)
				(2020,6600)
				(2021,11609)
				(2022,14606)
				(2023,23717)
			};
			\legend{cs.AI Preprints on arXiv}
		\end{axis}
	\end{tikzpicture}
	\caption{Annual number of preprints posted under the cs.AI category on arXiv.org}
	\label{fig:arxiv-preprints}
\end{figure}

Meanwhile, the exponential increase in the number of preprints posted on arXiv under the Computer Science $>$ Artificial Intelligence (cs.AI) category, as illustrated in Figure~\ref{fig:arxiv-preprints}, appears to signify a paradigm shift in research dissemination within the AI community. While the rapid distribution of findings enables swift knowledge exchange, it also raises concerns regarding the validation of information. The surge in preprints may lead to the propagation of unvalidated or biased information, as these studies do not undergo the rigorous scrutiny and potential retraction typical of peer-reviewed publications \cite{besanccon2021open,triggle2022requiem}. This trend underlines the need for careful consideration and critique in the academic community, especially given the potential for such unvetted studies to be cited and their findings propagated.

\subsection{Objectives}
The impetus for this investigation is the official unveiling of Gemini and the speculative discourse surrounding Q* project, which prompts a timely examination of the prevailing currents in generative AI research. This paper specifically contributes to the understanding of how MoE, multimodality, and Artificial General Intelligence (AGI) are impacting generative AI models, offering detailed analysis and future directions for each of these three key areas. This study does not aim to perpetuate conjecture about the unrevealed Q-Star initiative, but rather to critically appraise the potential for obsolescence or insignificance in extant research themes, whilst concurrently delving into burgeoning prospects within the rapidly transforming LLM panorama. This inquiry is reminiscent of the obsolete nature of encryption-centric or file-entropy-based ransomware detection methodologies, which have been eclipsed by the transition of ransomware collectives towards data theft strategies utilizing varied attack vectors, relegating contemporary studies on crypto-ransomware to the status of latecomers \cite{mcintosh2021ransomware,mcintosh2023harnessing}. Advances in AI are anticipated to not only enhance capabilities in language analysis and knowledge synthesis but also to pioneer in areas like Mixture of Experts (MoE) \cite{bao2022vlmo,du2022glam,masoudnia2014mixture,riquelme2021scaling,yuksel2012twenty,zhang2019learning}, multimodality \cite{martin2022multimodality,sun2023generative,wei2022mvp,wu2023menet,ye2023mplug}, and Artificial General Intelligence (AGI) \cite{lagrandeur2021safe,mclean2023risks,morris2023levels,schuett2023towards}, and has already heralded the obsolescence of conventional, statistics-driven natural language processing techniques in many domains \cite{maddigan2023chat2vis}. Nonetheless, the perennial imperative for AI to align with human ethics and values persists as a fundamental tenet \cite{dwivedi2021artificial,gabriel2020artificial,shaban2022applied}, and the conjectural Q-Star initiative offers an unprecedented opportunity to instigate discourse on how such advancements might reconfigure the LLM research topography. Within this milieu, insights from Dr. Jim Fan (senior research scientist \& lead of AI agents at NVIDIA) on Q*, particularly concerning the amalgamation of learning and search algorithms, furnish an invaluable perspective on the prospective technical construct and proficiencies of such an undertaking\footnote{\url{https://twitter.com/DrJimFan/status/1728100123862004105}}. Our research methodology involved a structured literature search using key terms like `Large Language Models' and `Generative AI'. We utilized filters across several academic databases such as IEEE Xplore, Scopus, ACM Digital Library, ScienceDirect, Web of Science, and ProQuest Central, tailored to identify relevant articles published in the timeframe from 2017 (the release of the \enquote{Transformer} model) to 2023 (the writing time of this manuscript). This paper aspires to dissect the technical ramifications of Gemini and Q*, probing how they (and similar technologies whose emergence is now inevitable) may transfigure research trajectories and disclose new vistas in the domain of AI. In doing so, we have pinpointed three nascent research domains—MoE, multimodality, and AGI—that stand to reshape the generative AI research landscape profoundly. This investigation adopts a survey-style approach, systematically mapping out a research roadmap that synthesizes and analyzes the current and emergent trends in generative AI.

The major contributions of this study is as follows:
\begin{enumerate}
	\item Detailed examination of the evolving landscape in generative AI, emphasizing the advancements and innovations in technologies like Gemini and Q*, and their wide-ranging implications within the AI domain.
	\item Analysis of the transformative effect of advanced generative AI systems on academic research, exploring how these developments are altering research methodologies, setting new trends, and potentially leading to the obsolescence of traditional approaches.
	\item Thorough assessment of the ethical, societal, and technical challenges arising from the integration of generative AI in academia, underscoring the crucial need for aligning these technologies with ethical norms, ensuring data privacy, and developing comprehensive governance frameworks.
\end{enumerate}

The rest of this paper is organized as follows: Section \ref{sec:background} explores the historical development of Generative AI. Section \ref{sec:generative-ai-taxonomy} presents a taxonomy of current Generative AI research. Section \ref{sec:innovative-horizon-moe} explores the Mixture of Experts (MoE) model architecture, its innovative features, and its impact on transformer-based language models. Section \ref{sec:speculated-capabilities-q-star} discusses the speculated capabilities of the Q* project. Section \ref{sec:projected-capabilities-agi} discusses the projected capabilities of AGI. Section \ref{sec:impact-analysis-llm} examines the impact of recent advancements on the Generative AI research taxonomy. Section \ref{sec:emergent-research-priorities-agi} identifies emerging research priorities in Generative AI. Section \ref{sec:academic-challenges-preprints} discusses the academic challenges of the rapid surge of preprints in AI. The paper concludes in Section \ref{sec:conclusions}, summarizing the overall effects of these developments in generative AI.

\section{Background: Evolution of Generative AI}
\label{sec:background}
The ascent of Generative AI has been marked by significant milestones, with each new model paving the way for the next evolutionary leap. From single-purpose algorithms to LLMs like OpenAI's ChatGPT and the latest multimodal systems, the AI landscape has been transformed, while countless other fields have been disrupted.

\subsection{The Evolution of Language Models}
\label{subsec:evolution-language-models}
\begin{figure}[t!]
	\centering
	\begin{tikzpicture}[node distance=1cm, auto, decoration={markings,mark=at position 1 with {\arrow{latex}}}]
		
		\draw[thick, postaction={decorate}] (0,0) -- (0,-6);
		
		\node at (0,0) [right] {1980s: Statistical Models (n-grams)};
		\node at (0,-1) [right] {1990s: Adoption in NLP, n-gram Usage};
		\node at (0,-2) [right] {1997: Introduction of LSTMs};
		\node at (0,-3) [right] {2000s: LSTMs in Text/Voice Processing};
		\node at (0,-4) [right] {2010s: Deep Learning Era, GPT, BERT};
		\node at (0,-5) [right] {2020s: LLaMA, Gemini; ChatGPT Launch};
		
		\foreach \y in {0,-1,-2,-3,-4,-5} {
			\draw[fill=black] (0,\y) circle (2pt);
		}
		
	\end{tikzpicture}
	\caption{Timeline of Key Developments in Language Model Evolution}
	\label{fig:language-model-timeline}
\end{figure}
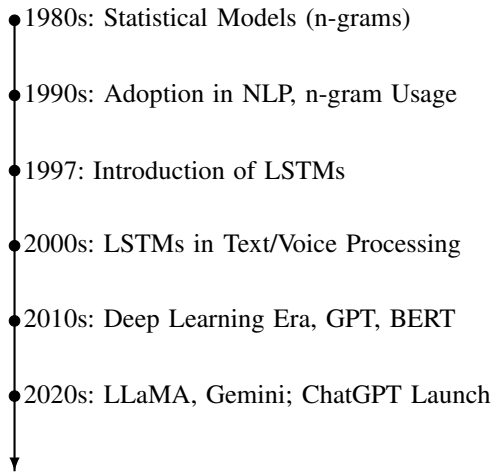

Language models have undergone a transformative journey (Fig. \ref{fig:language-model-timeline}), evolving from rudimentary statistical methods to the complex neural network architectures that underpin today's LLMs \cite{ji2023survey,min2023recent}. This evolution has been driven by a relentless quest for models that more accurately reflect the nuances of human language, as well as the desire to push the boundaries of what machines can understand and generate \cite{ji2023survey,li2023halueval,min2023recent}. However, this rapid advancement has not been without its challenges. As language models have grown in capability, so too have the ethical and safety concerns surrounding their use, prompting a reevaluation of how these models are developed and the purposes for which they are employed \cite{ji2023survey,weidinger2021ethical,zhiheng2023safety}.

\subsubsection{Language Models as Precursors}
\label{subsubsec:language-models-precursors}
The inception of language modeling can be traced to the statistical approaches of the late 1980s, a period marked by a transition from rule-based to machine learning algorithms in Natural Language Processing (NLP) \cite{brown1992class,katz1987estimation,kneser1995improved,kuhn1990cache,ney1994structuring}. Early models, primarily $n$-gram based, calculated the probability of word sequences in a corpus, thus providing a rudimentary understanding of language structure \cite{brown1992class}. Those models, simplistic yet groundbreaking, laid the groundwork for future advances in language understanding. With the increase of computational power, the late 1980s witnessed a revolution in NLP, pivoting towards statistical models capable of `soft' probabilistic decisions, as opposed to the rigid, `handwritten' rule-based systems that dominated early NLP systems \cite{kneser1995improved}. IBM's development of complicated statistical models throughout this period signified the growing importance and success of these approaches. In the subsequent decade, the popularity and applicability of statistical models surged, proving invaluable in managing the flourishing flow of digital text. The 1990s saw statistical methods firmly established in NLP research, with n-grams becoming instrumental in numerically capturing linguistic patterns. The introduction of Long Short-Term Memory (LSTM) networks in 1997 \cite{hochreiter1997long}, and their application to voice and text processing a decade later \cite{nammous2019natural,wei2017research,yao2018improved}, marked a significant milestone, leading to the current era where neural network models represent the cutting edge of NLP research and development.

\subsubsection{Large Language Models: Technical Advancement and Commercial Success}
\label{subsubsec:llms-expansion-limits}
The advent of deep learning has revolutionized the field of NLP, leading to the development of LLMs like GPT, BERT, and notably, OpenAI's ChatGPT. Recent models such as GPT-4 and LLaMA have pushed the boundaries by integrating sophisticated techniques like transformer architectures and advanced natural language understanding, illustrating the rapid evolution in this field \cite{min2023recent}. These models represent a significant leap in NLP capabilities, leveraging vast computational resources and extensive datasets to achieve new heights in language understanding and generation \cite{min2023recent,ouyang2022training}. ChatGPT has shown impressive conversational skills and contextual understanding with a broad spectrum of functional uses in many areas, as evidenced by its technical and commercial success, including rapid adoption by over 100 million users shortly after launch, which underscores a robust market demand for natural language AI and has catalyzed interdisciplinary research into its applications in sectors like education, healthcare, and commerce \cite{maddigan2023chat2vis,ouyang2022training,susnjak2023beyond,susnjak2023towards,yang2023large}. In education, ChatGPT offers innovative approaches to personalized learning and interactive teaching \cite{baidoo2023education,susnjak2023beyond,susnjak2022chatgpt,tlili2023if}, while in commerce, it revolutionizes customer service and content creation \cite{alafnan2023chatgpt,george2023review}. The widespread use of ChatGPT, Google Bard, Anthropic Claude and similar commercial LLMs has reignited important debates in the field of AI, particularly concerning AI consciousness and safety, as its human-like interaction capabilities raise significant ethical questions and highlight the need for robust governance and safety measures in AI development \cite{hadfield2023regulatory,lagrandeur2021safe,mclean2023risks,schuett2023towards}. Such influence appears to extend beyond its technical achievements, shaping cultural and societal discussions about the role and future of AI in our world.

The advancements in LLMs, including the development of models like GPT and BERT, have paved the way for the conceptualization of Q*. Specifically, the scalable architecture and extensive training data that characterize these models are foundational to the proposed capabilities of Q*. The success of ChatGPT in contextual understanding and conversational AI, for example, informs the design principles of Q*, suggesting a trajectory towards more sophisticated, context-aware, and adaptive language processing capabilities. Similarly, the emergence of multimodal systems like Gemini, capable of integrating text, images, audio, and video, reflects an evolutionary path that Q* could extend, combining the versatility of LLMs with advanced learning and pathfinding algorithms for a more holistic AI solution.

\subsubsection{Fine-tuning, Hallucination Reduction, and Alignment in LLMs}
\label{subsubsec:fine-tuning-alignment}
The advancement of LLMs has underlined the significance of fine-tuning \cite{bakker2022fine,hu2023llm,liu2022few,zheng2023learn}, hallucination reduction \cite{manakul2023selfcheckgpt,martino2023knowledge,yao2023llm,zhang2023siren}, and alignment \cite{ji2023beavertails,liu2023trustworthy,wang2023aligning,sun2023principle,wolf2023fundamental}. These aspects are crucial in enhancing the functionality and reliability of LLMs. Fine-tuning, which involves adapting pre-trained models to specific tasks, has seen significant progress: techniques like prompt-based and few-shot learning \cite{dang2022prompt,ma2021template,qin2021lfpt5,wang2022trustworthy}, alongside supervised fine-tuning on specialized datasets \cite{bakker2022fine,fan2023grammargpt,liga2023fine,liu2023improving}, have enhanced the adaptability of LLMs in various contexts, but challenges remain, particularly in bias mitigation and the generalization of models across diverse tasks \cite{bakker2022fine,talat2022you,wolf2023fundamental}. Hallucination reduction is a persistent challenge in LLMs, characterized by the generation of confident but factually incorrect information \cite{ji2023survey}. Strategies such as confidence penalty regularization during fine-tuning have been implemented to mitigate overconfidence and improve accuracy \cite{liu2023hessian,lu2021confidence,pereyra2017regularizing}. Despite these efforts, the complexity of human language and the breadth of topics make completely eradicating hallucinations a daunting task, especially in culturally sensitive contexts \cite{ji2023survey,mcintosh2023culturally}. Alignment, ensuring LLM outputs are congruent with human values and ethics, is an area of ongoing research. Innovative approaches, from constrained optimization \cite{chen2022redeeming,jiang2022stable,kachuee2022constrained,song2023surrogate,yu2022structure}, to different types of reward modeling \cite{butlin2021ai,faal2023reward,leike2018scalable,li2023tool}, aim to embed human preferences within AI systems. While advancements in fine-tuning, hallucination reduction, and alignment have propelled LLMs forward, these areas still present considerable challenges. The complexity of aligning AI with the diverse spectrum of human ethics and the persistence of hallucinations, particularly on culturally sensitive topics, highlight the need for continued interdisciplinary research in the development and application of LLMs \cite{mcintosh2023culturally}.

\subsubsection{Mixture of Experts: A Paradigm Shift}
\label{subsubsec:moe-paradigm-shift}
The adoption of the MoE architecture in LLMs marks a critical evolution in AI technology. This innovative approach, exemplified by advanced models like Google's Switch Transformer\footnote{https://huggingface.co/google/switch-c-2048} and MistralAI
s Mixtral-8x7B\footnote{https://huggingface.co/mistralai/Mixtral-8x7B-v0.1}, leverages multiple transformer-based expert modules for dynamic token routing, enhancing modeling efficiency and scalability. The primary advantage of MoE lies in its ability to handle vast parameter scales, reducing memory footprint and computational costs significantly \cite{barreto2023generative,chen2023octavius,dun2023sweeping,naveed2023comprehensive,xue2023repeat}. This is achieved through model parallelism across specialized experts, allowing the training of models with trillions of parameters, and its specialization in handling diverse data distributions enhances its capability in few-shot learning and other complex tasks \cite{chen2023octavius,dun2023sweeping}. To illustrate the practicality of MoE, consider its application in healthcare. For example, an MoE-based system could be used for personalized medicine, where different `expert' modules specialize in various aspects of patient data analysis, including genomics, medical imaging, and electronic health records. This approach could significantly enhance diagnostic accuracy and treatment personalization. Similarly, in finance, MoE models can be deployed for risk assessment, where experts analyze distinct financial indicators, market trends, and regulatory compliance factors.

Despite its benefits, MoE confronts challenges in dynamic routing complexity \cite{nowaz2023patch,peng2023sparse,santos2023memory,wang2023language,zhao2023sparse}, expert imbalance \cite{huang2023multi,liu2023diversifying,wang2023prophet,yao2023enhancing}, and probability dilution \cite{xiao2021adversarial}, and such technical hurdles demand sophisticated solutions to fully harness MoE's potential. Moreover, while MoE may offer performance gains, it does not inherently solve ethical alignment issues in AI \cite{agbese2023implementing,chen2022towards,zhou2022mixture}. The complexity and specialization of MoE models can obscure the decision-making processes, complicating efforts to ensure ethical compliance and alignment with human values \cite{agbese2023implementing,guha2023ai}. Although the paradigm shift to MoE signifies a major leap in LLM development, offering significant scalability and specialization advantages, ensuring the safety, ethical alignment, and transparency of these models remains a paramount concern. The MoE architecture, while technologically advanced, entails continued interdisciplinary research and governance to align AI with broader societal values and ethical standards.

\subsection{Multimodal AI and the Future of Interaction}
\label{subsec:multimodal-future}
The advent of multimodal AI marks a transformative era in AI development, revolutionizing how machines interpret and interact with a diverse array of human sensory inputs and contextual data.

\subsubsection{Gemini: Redefining Benchmarks in Multimodality}
\label{subsubsec:gemini-redefining-benchmarks}
Gemini, a pioneering multimodal conversational system, marks a significant shift in AI technology by surpassing traditional text-based LLMs like GPT-3 and even its multimodal counterpart, ChatGPT-4. Gemini's architecture has been designed to incorporate the processing of diverse data types such as text, images, audio, and video, a feat facilitated by its unique multimodal encoder, cross-modal attention network, and multimodal decoder \cite{gemini2023gemini}. The architectural core of Gemini is its dual-encoder structure, with separate encoders for visual and textual data, enabling sophisticated multimodal contextualization \cite{gemini2023gemini}. This architecture is believed to surpass the capabilities of single-encoder systems, allowing Gemini to associate textual concepts with image regions and achieve a compositional understanding of scenes \cite{gemini2023gemini}. Furthermore, Gemini integrates structured knowledge and employs specialized training paradigms for cross-modal intelligence, setting new benchmarks in AI \cite{gemini2023gemini}. In \cite{gemini2023gemini}, Google has claimed and demonstrated that Gemini distinguishes itself from ChatGPT-4 through several key features:

\begin{itemize}
	\item \textit{Breadth of Modalities:} Unlike ChatGPT-4, which primarily focuses on text, documents, images, and code, Gemini handles a wider range of modalities including audio, and video. This extensive range allows Gemini to tackle complex tasks and understand real-world contexts more effectively.
	
	\item \textit{Performance:} Gemini Ultra excels in key multimodality benchmarks, notably in massive multitask language understanding (MMLU) which encompasses a diverse array of domains like science, law, and medicine, outperforming ChatGPT-4.
	
	\item \textit{Scalability and Accessibility:} Gemini is available in three tailored versions – Ultra, Pro, and Nano – catering to a range of applications from data centers to on-device tasks, a level of flexibility not yet seen in ChatGPT-4.
	
	\item \textit{Code Generation:} Gemini's proficiency in understanding and generating code across various programming languages is more advanced, offering practical applications beyond ChatGPT-4's capabilities.
	
	\item \textit{Transparency and Explainability:} A focus on explainability sets Gemini apart, as it provides justifications for its outputs, enhancing user trust and understanding of the AI's reasoning process.
\end{itemize}

Despite these advancements, Gemini's real-world performance in complex reasoning tasks that require integration of commonsense knowledge across modalities remains to be thoroughly evaluated.

\subsubsection{Technical Challenges in Multimodal Systems}
\label{subsubsec:technical-challenges-advancements}
The development of multimodal AI systems faces several technical hurdles, including creating robust and diverse datasets, managing scalability, and enhancing user trust and system interpretability \cite{acosta2022multimodal,qi2023limitation,xu2020applications}. Challenges like data skew and bias are prevalent due to data acquisition and annotation issues, which requires effective dataset management by employing strategies such as data augmentation, active learning, and transfer learning \cite{acosta2022multimodal,birhane2021multimodal,talat2022you,xu2020applications}. A significant challenge is the computational demands of processing various data streams simultaneously, requiring powerful hardware and optimized model architectures for multiple encoders \cite{li2021multimodal,zhang2020multimodal}. Advanced algorithms and multimodal attention mechanisms are needed to balance attention across different input media and resolve conflicts between modalities, especially when they provide contradictory information \cite{qiao2022initial,stewart2021multimodal,zhang2020multimodal}. Scalability issues, due to the extensive computational resources needed, are exacerbated by limited high-performance hardware availability \cite{xue2023ulip,yan2022scalability}. There is also a pressing need for calibrated multimodal encoders for compositional scene understanding and data integration \cite{stewart2021multimodal}. Refining evaluation metrics for these systems is necessary to accurately assess performance in real-world tasks, calling for comprehensive datasets and unified benchmarks, and for enhancing user trust and system interpretability through explainable AI in multimodal contexts. Addressing these challenges is vital for the advancement of multimodal AI systems, enabling seamless and intelligent interaction aligned with human expectations.

\subsubsection{Multimodal AI: Beyond Text in Ethical and Social Contexts}
\label{subsubsec:multimodal-ethical-social-contexts}
The expansion of multimodal AI systems introduces both benefits and complex ethical and social challenges that extend beyond those faced by text-based AI. In commerce, multimodal AI can transform customer engagement by integrating visual, textual, and auditory data \cite{liu2022artificial,rahman2023new,sachdeva2023rank2tell}. For autonomous vehicles, multimodality can enhance safety and navigation by synthesizing data from various sensors, including visual, radar, and Light Detection and Ranging (LIDAR) \cite{cui2023survey,sachdeva2023rank2tell,temsamani2022multimodal}. Still, DeepFake technology's ability to generate convincingly realistic videos, audio, and images is a critical concern in multimodality, as it poses risks of misinformation and manipulation that significantly impact public opinion, political landscapes, and personal reputations, thereby compromising the authenticity of digital media and raising issues in social engineering and digital forensics where distinguishing genuine from AI-generated content becomes increasingly challenging \cite{lee2022something,muppalla2023integrating}. Privacy concerns are amplified in multimodal AI due to its ability to process and correlate diverse data sources, potentially leading to intrusive surveillance and profiling, which raises questions about the consent and rights of individuals, especially when personal media is used without permission for AI training or content creation \cite{acosta2022multimodal,kumar2022privacy,marchang2022assistive}. Moreover, multimodal AI can propagate and amplify biases and stereotypes across different modalities, and if unchecked, this can perpetuate discrimination and social inequities, making it imperative to address algorithmic bias effectively \cite{pena2023human,wolfe2022american,wolfe2023contrastive}. The ethical development of multimodal AI systems requires robust governance frameworks focusing on transparency, consent, data handling protocols, and public awareness, when ethical guidelines must evolve to address the unique challenges posed by these technologies, including setting standards for data usage and safeguarding against the nonconsensual exploitation of personal information \cite{afshar2022development,alwahaby2022evidence}. Additionally, the development of AI literacy programs will be crucial in helping society understand and responsibly interact with multimodal AI technologies \cite{acosta2022multimodal,afshar2022development}. As the field progresses, interdisciplinary collaboration will be key in ensuring these systems are developed and deployed in a manner that aligns with societal values and ethical principles \cite{acosta2022multimodal}.

\subsection{Speculative Advances and Chronological Trends}
\label{subsec:speculative-chronological}
In the dynamic landscape of AI, the speculative capabilities of the Q* project, blending LLMs, Q-learning, and A* (A-Star algorithm), embodies a significant leap forward. This section explores the evolutionary trajectory from game-centric AI systems to the broad applications anticipated with Q*.

\subsubsection{From AlphaGo's Groundtruth to Q-Star's Exploration}
\label{subsubsec:alphago-qstar}
The journey from AlphaGo, a game-centric AI, to the conceptual Q-Star project represents a significant paradigm shift in AI. AlphaGo's mastery in the game of Go highlighted the effectiveness of deep learning and tree search algorithms within well-defined rule-based environments, underscoring the potential of AI in complex strategy and decision-making \cite{miao2023dao,rong2022roadmap}. Q-Star, however, is speculated to move beyond these confines, aiming to amalgamate the strengths of reinforcement learning (as seen in AlphaGo), with the knowledge, NLG, creativity and versatility of LLMs, and the strategic efficiency of pathfinding algorithms like A*. This blend, merging pathfinding algorithms and LLMs, could enable AI systems to transcend board game confines and, with Q-Star's natural language processing, interact with human language, enabling nuanced interactions and marking a leap towards AI adept in both structured tasks and complex human-like communication and reasoning. Moreover, the incorporation of Q-learning and A* algorithms would enable Q-Star to optimize decision paths and learn from its interactions, making it more adaptable and intelligent over time. The combination of these technologies could lead to AI that is not only more efficient in problem-solving but also creative and insightful in its approach. This speculative advancement from the game-focused power of AlphaGo to the comprehensive potential of Q-Star illustrates the dynamic and ever-evolving nature of AI research, and opens up possibilities for AI applications that are more integrated with human life and capable of handling a broader range of tasks with greater autonomy and sophistication.

\subsubsection{Bridging Structured Learning with Creativity}
\label{subsubsec:bridging-creativity}
The anticipated Q* project, blending Q-learning and A* algorithms with the creativity of LLMs, embodies a groundbreaking step in AI, potentially surpassing recent innovations like Gemini. The fusion suggested in Q* points to an integration of structured, goal-oriented learning with generative, creative capabilities, a combination that could transcend the existing achievements of Gemini. While Gemini represents a significant leap in multimodal AI, combining various forms of data inputs such as text, images, audio, and video, Q* is speculated to bring a more profound integration of creative reasoning and structured problem-solving. This would be achieved by merging the precision and efficiency of algorithms like A* with the learning adaptability of Q-learning, and the complex understanding of human language and context offered by LLMs. Such an integration could enable AI systems to not only process and analyze complex multimodal data but also to autonomously navigate through structured tasks while engaging in creative problem-solving and knowledge generation, mirroring the multifaceted nature of human cognition. The implications of this potential advancement are vast, suggesting applications that span beyond the capabilities of current multimodal systems like Gemini. By aligning the deterministic aspects of traditional AI algorithms with the creative and generative potential of LLMs, Q* could offer a more holistic approach to AI development. This could bridge the gap between the logical, rule-based processing of AI and the creative, abstract thinking characteristic of human intelligence. The anticipated unveiling of Q*, merging structured learning techniques and creative problem-solving in a singular, advanced framework, holds the promise of not only extending but also significantly surpassing the multimodal capabilities of systems like Gemini, thus heralding another game-changing era in the domain of generative AI, showcasing its potential as a crucial development eagerly awaited in the ongoing evolution of AI.

\section{The Current Generative AI Research Taxonomy}
\label{sec:generative-ai-taxonomy}
The field of Generative AI is evolving rapidly, which necessitates a comprehensive taxonomy that encompasses the breadth and depth of research within this domain. Detailed in Table~\ref{table:generative-ai-taxonomy}, this taxonomy categorizes the key areas of inquiry and innovation in generative AI, and serves as a foundational framework to understand the current state of the field, guiding through the complexities of evolving model architectures, advanced training methodologies, diverse application domains, ethical implications, and the frontiers of emerging technologies.

\begin{table*}[t]
	\centering
	\caption{Comprehensive Taxonomy of Current Generative AI and LLM Research}
	\label{table:generative-ai-taxonomy}
	\resizebox{\textwidth}{!}{
		\begin{tabular}{|p{2cm}|p{2.5cm}|p{2.5cm}|p{9cm}|}
			\hline
			\textbf{Domain} & \textbf{Subdomain} & \textbf{Key Focus} & \textbf{Description} \\ \hline
			Model Architecture & Transformer Models & Efficiency, Scalability & Optimizing network structures for faster processing and larger datasets. \\ \cline{2-4} 
			& Recurrent Neural Networks & Sequence Processing & Handling sequences of data, like text, for improved contextual understanding. \\ \cline{2-4}
			& Mixture of Experts & Specialization, Efficiency & Leveraging multiple expert modules for enhanced efficiency and task-specific performance. \\ \cline{2-4} 
			& Multimodal Models & Sensory Integration & Integrating text, vision, and audio inputs for comprehensive understanding. \\ \hline
			
			Training Techniques & Supervised Learning & Data Labeling, Accuracy & Using labeled datasets to train models for precise predictions. \\ \cline{2-4} 
			& Unsupervised Learning & Pattern Discovery & Finding patterns and structures from unlabeled data. \\ \cline{2-4} 
			& Reinforcement Learning & Adaptability, Optimization & Training models through feedback mechanisms for optimal decision-making. \\ \cline{2-4} 
			& Transfer Learning & Versatility, Generalization & Applying knowledge gained in one task to different but related tasks. \\ \hline
			
			Application Domains & Natural Language Understanding & Comprehension, Contextualization & Enhancing the ability to understand and interpret human language in context. \\ \cline{2-4} 
			& Natural Language Generation & Creativity, Coherence & Generating coherent and contextually relevant text responses. \\ \cline{2-4}
			& Conversational AI & Interaction, Naturalness & Developing systems for natural and contextually relevant human-computer conversations. \\ \cline{2-4} 
			& Creative AI & Innovation, Artistic Generation & Generating creative content, including text, art, and music. \\ \hline
			
			Compliance and Ethical Considerations & Bias Mitigation & Fairness, Representation & Addressing and reducing biases in AI outputs. \\ \cline{2-4} 
			& Data Security & Data Protection, Confidentiality & Ensuring data confidentiality, integrity and availability security in AI models and outputs. \\ \cline{2-4} 
			& AI Ethics & Fairness, Accountability & Addressing ethical issues such as bias, fairness, and accountability in AI systems. \\ \cline{2-4} 
			& Privacy Preservation & Privacy Compliance, Anonymization & Protecting data privacy in model training and outputs. \\ \hline
			
			Advanced Learning & Self-supervised Learning & Autonomy, Efficiency & Utilizing unlabeled data for model training, enhancing learning efficiency. \\ \cline{2-4}
			& Meta-learning & Rapid Adaptation & Enabling AI models to quickly adapt to new tasks with minimal data. \\ \cline{2-4}
			& Fine Tuning & Domain-Specific Tuning, Personalization & Adapting models to specific domains or user preferences for enhanced relevance and accuracy. \\ \cline{2-4}
			& Human Value Alignment & Ethical Integration, Societal Alignment & Aligning AI outputs with human ethics and societal norms, ensuring decisions are ethically and socially responsible. \\ \hline
			
			Emerging Trends & Multimodal Learning & Integration with Vision, Audio & Combining language models with other sensory data types for richer understanding. \\ \cline{2-4} 
			& Interactive and Cooperative AI & Collaboration, Human-AI Interaction & Enhancing AI's ability to work alongside humans in collaborative tasks. \\ \cline{2-4}
			& AGI Development & Holistic Understanding & Pursuing the development of AI systems with comprehensive, human-like understanding. \\ \cline{2-4}
			& AGI Containment & Safety Protocols, Control Mechanisms & Developing methods to contain and control AGI systems to prevent unintended consequences. \\ 
			 \hline
		\end{tabular}
	}
\end{table*}

\subsection{Model Architectures}
\label{subsec:model-architectures}
Generative AI model architectures have seen significant developments, with four key domains standing out:

\begin{itemize}
	\item \textbf{Transformer Models:} Transformer models have significantly revolutionized the field of AI, especially in NLP, due to their higher efficiency and scalability \cite{gao2022data,peebles2023scalable,pope2023efficiently}. They employ advanced attention mechanisms to achieve enhanced contextual processing, allowing for more subtle understanding and interaction \cite{ding2022convolutional,ding2022novel,wang2022shift}. These models have also made notable strides in computer vision, as evidenced by the development of vision transformers like EfficientViT \cite{cai2023efficientvit,liu2023efficientvit} and YOLOv8 \cite{li2023modified,talaat2023improved,tamang2023enhancing}. These innovations symbolize the extended capabilities of transformer models in areas such as object detection, offering not only improved performance but also increased computational efficiency. 	
	
	\item \textbf{Recurrent Neural Networks (RNNs):} RNNs excel in the realm of sequence modeling, making them particularly effective for tasks involving language and temporal data, as their architecture is specifically designed to process sequences of data, such as text, enabling them to capture the context and order of the input effectively \cite{lu2022battery,onan2022bidirectional,shan2022success,sridhar2022optimal,zhu2022application}. This proficiency in handling sequential information renders them indispensable in applications that require a deep understanding of the temporal dynamics within data, such as natural language tasks and time-series analysis \cite{lin2023segrnn,wei2022extracting}. RNNs' ability to maintain a sense of continuity over sequences is a critical asset in the broader field of AI, especially in scenarios where context and historical data play crucial roles \cite{bonassi2022recurrent}.

	\item \textbf{Mixture of Experts (MoE):} MoE models can significantly enhance efficiency by deploying model parallelism across multiple specialized expert modules, which enables these models to leverage transformer-based modules for dynamic token routing, and to scale to trillions of parameters, thereby reducing both memory footprint and computational costs \cite{chen2023octavius,nowaz2023patch}. MoE models stand out for their ability to divide computational loads among various experts, each specializing in different aspects of the data, which allows for handling vast scales of parameters more effectively, leading to a more efficient and specialized handling of complex tasks \cite{chen2023octavius,du2022glam}.

	\item \textbf{Multimodal Models:} Multimodal models, which integrate a variety of sensory inputs such as text, vision, and audio, are crucial in achieving a comprehensive understanding of complex data sets, particularly transformative in fields like medical imaging \cite{acosta2022multimodal,gemini2023gemini,xu2020applications}. These models facilitate accurate and data-efficient analysis by employing multi-view pipelines and cross-attention blocks \cite{guo2023viewrefer,pan2023baeformer}. This integration of diverse sensory inputs allows for a more nuanced and detailed interpretation of data, enhancing the model's ability to accurately analyze and understand various types of information \cite{xu2023multimodal}. The combination of different data types, processed concurrently, enables these models to provide a holistic view, making them especially effective in applications that require a deep and multifaceted understanding of complex scenarios \cite{acosta2022multimodal,molenaar2023measuring,steyaert2023multimodal,xu2023multimodal}.
	
\end{itemize}

\subsection{Training Techniques}
\label{subsec:training-techniques}

The training of generative AI models leverages four key techniques, each contributing uniquely to the field:

\begin{itemize}
	\item \textbf{Supervised Learning:} Supervised learning, a foundational approach in AI, uses labeled datasets to guide models towards accurate predictions, and it has been integral to various applications, including image recognition and NLP  \cite{rani2023self,schiappa2023self,yu2023self}. Recent advancements have focused on developing sophisticated loss functions and regularization techniques, aimed at enhancing the performance and generalization capabilities of supervised learning models, ensuring they remain robust and effective across a wide range of tasks and data types \cite{bharti2023label,sam2023losses,wang2023t5}. 
	
	\item \textbf{Unsupervised Learning:} Unsupervised learning is essential in AI for uncovering patterns within unlabeled data, a process central to tasks like feature learning and clustering \cite{li2022unsupervised,nancy2022deep}. This method has seen significant advancements with the introduction of autoencoders \cite{an2022ensemble,yan2023hybrid} and Generative Adversarial Networks (GANs) \cite{ayanoglu2022machine,yan2022physical,zhou2023hybrid}, which have notably expanded unsupervised learning's applicability, enabling more sophisticated data generation and representation learning capabilities. Such innovations are crucial for understanding and leveraging the complex structures often inherent in unstructured datasets, highlighting the growing versatility and depth of unsupervised learning techniques.

	\item \textbf{Reinforcement Learning:} Reinforcement learning, characterized by its adaptability and optimization capabilities, has become increasingly vital in decision-making and autonomous systems \cite{ladosz2022exploration,matsuo2022deep}. This training technique has undergone significant advancements, particularly with the development of Deep Q-Networks (DQN) \cite{bertoin2022look,hafiz2022survey,hafiz2023reinforcement} and Proximal Policy Optimization (PPO) algorithms \cite{alagha2022target,hassan20223to,jayant2022model}. These enhancements have been crucial in improving the efficacy and applicability of reinforcement learning, especially in complex and dynamic environments. By optimizing decisions and policies through interactive feedback loops, reinforcement learning has established itself as a crucial tool for training AI systems in scenarios that demand a high degree of adaptability and precision in decision-making \cite{lin2023reinforcement,luo2023human}.

	\item \textbf{Transfer Learning:} Transfer learning emphasizes versatility and efficiency in AI training, allowing models to apply knowledge acquired from one task to different yet related tasks, which significantly reduces the need for large labeled datasets \cite{raza2022designing,siahpour2022novel}. Transfer learning, through the use of pre-trained networks, streamlines the training process by allowing models to be efficiently fine-tuned for specific applications, thereby enhancing adaptability and performance across diverse tasks, and proving particularly beneficial in scenarios where acquiring extensive labeled data is impractical or unfeasible \cite{guo2022transfer,liu2022adaptive}.
	
\end{itemize}

\subsection{Application Domains}
The application domains of Generative AI are remarkably diverse and evolving, encompassing both established and emerging areas of research and application. These domains have been significantly influenced by recent advancements in AI technology and the expanding scope of AI applications.

\begin{itemize}
	\item \textbf{Natural Language Understanding (NLU)}: NLU is central to enhancing the comprehension and contextualization of human language in AI systems, and involves key capabilities such as semantic analysis, named entity recognition, sentiment analysis, textual entailment, and machine reading comprehension \cite{liu2023logiqa,meng2022generating,samant2022framework,weld2022survey}. Advances in NLU have been crucial in improving AI's proficiency in interpreting and analyzing language across a spectrum of contexts, ranging from straightforward conversational exchanges to intricate textual data \cite{liu2023logiqa,samant2022framework,weld2022survey}. NLU is fundamental in applications like sentiment analysis, language translation, information extraction, and more \cite{ajmal2023natural,montejo2022current,vijayan2022language}. Recent advancements have prominently featured large transformer-based models like BERT and GPT-3, which have significantly advanced the field by enabling a deeper and more complex understanding of language subtleties \cite{manning2022human,peng2023gpt}.
	
	\item \textbf{Natural Language Generation (NLG)}: NLG emphasizes the training of models to generate coherent, contextually-relevant, and creative text responses, a critical component in chatbots, virtual assistants, and automated content creation tools \cite{erdem2022neural,ji2023survey,qian2022controllable,rashkin2023measuring}. NLG encompasses challenges such as topic modeling, discourse planning, concept-to-text generation, style transfer, and controllable text generation \cite{ji2023survey,pandey2023natural}. The recent surge in NLG capabilities, exemplified by advanced models like GPT-3, has significantly enhanced the sophistication and nuance of text generation, which enable AI systems to produce text that closely mirrors human writing styles, thereby broadening the scope and applicability of NLG in various interactive and creative contexts \cite{khan2022automatic,susnjak2022chatgpt,susnjak2023beyond}.
	
	\item \textbf{Conversational AI}: This subdomain is dedicated to developing AI systems capable of smooth, natural, and context-aware human-computer interactions, by focusing on dialogue modeling, question answering, user intent recognition, and multi-turn context tracking \cite{dwivedi2023so,fu2022learning,ji2023systematic,wan2023biasasker}. In finance and cybersecurity, AI's predictive analytics have transformed risk assessment and fraud detection, leading to more secure and efficient operations \cite{fu2022learning,mcintosh2023harnessing}. The advancements in this area, demonstrated by large pre-trained models like Meena\footnote{https://neptune.ai/blog/transformer-nlp-models-meena-lamda-chatbots} and BlenderBot\footnote{https://blenderbot.ai}, have significantly enhanced the empathetic and responsive capabilities of AI interactions. These systems not only improve user engagement and satisfaction, but also maintain the flow of conversation over multiple turns, providing coherent, contextually relevant, and engaging experiences \cite{kusal2022ai,xiao2023seeing}.
	
	\item \textbf{Creative AI}: This emerging subdomain spans across text, art, music, and more, pushing the boundaries of AI's creative and innovative potential across various modalities including images, audio, and video, by engaging in the generation of artistic content, encompassing applications in idea generation, storytelling, poetry, music composition, visual arts, and creative writing, and has resulted in commercial success like MidJourney and DALL-E \cite{ko2023large,pearson2023rise,rezwana2023designing}. The challenges in this field involve finding suitable data representations, algorithms, and evaluation metrics to effectively assess and foster creativity \cite{rezwana2023designing,sharma2023generative}. Creative AI serves not only as a tool for automating and enhancing artistic processes, but also as a medium for exploring new forms of artistic expression, enabling the creation of novel and diverse creative outputs \cite{rezwana2023designing}. This domain represents a significant leap in AI's capability to engage in and contribute to creative endeavors, redefining the intersection of technology and art.
\end{itemize}

\subsection{Compliance and Ethical Considerations}
\label{subsec:compliance-ethical-considerations}
As AI technologies rapidly evolve and become more integrated into various sectors, ethical considerations and legal compliance have become increasingly crucial, which requires a focus on developing `Ethical AI Frameworks', a new category in our taxonomy reflecting the trend towards responsible AI development in generative AI \cite{attard2023ethics,gardner2022ethical,huang2022overview,schuett2023three,sloane2022german}. Such frameworks are crucial in ensuring AI systems are built with a core emphasis on ethical considerations, fairness, and transparency, as they address critical aspects such as bias mitigation for fairness, privacy and security concerns for data protection, and AI ethics for accountability, thus responding to the evolving landscape where accountability in AI is of paramount importance \cite{attard2023ethics,huang2022overview}. The need for rigorous approaches to uphold ethical integrity and legal conformity has never been more pressing, reflecting the complexity and multifaceted challenges introduced by the adoption of these technologies \cite{huang2022overview}.

\begin{itemize}
	\item \textbf{Bias Mitigation:} Bias Mitigation in AI systems is a critical endeavor to ensure fairness and representation, which involves not only balanced data collection to avoid skewed perspectives but also involves implementing algorithmic adjustments and regularization techniques to minimize biases \cite{vasconcelos2018modeling,yang2022enhancing}. Continuous monitoring and bias testing are essential to identify and address any biases that may emerge from AI's predictive patterns \cite{schwartz2022towards,yang2022enhancing}. A significant challenge in this area is dealing with intersectional biases \cite{guo2021detecting,kong2022intersectionally,tan2019assessing} and understanding the causal interactions that may contribute to these biases \cite{cheng2021causal,correa2019identification,ghai2022d,yan2020silva}. 
	
	\item \textbf{Data Security:} In AI data security, key requirements and challenges include ensuring data confidentiality, adhering to consent norms, and safeguarding against vulnerabilities like membership inference attacks \cite{bertino2021ai,susanto2021data}. Compliance with stringent legal standards within applicable jurisdictions, such as the General Data Protection Regulation (GDPR) and California Consumer Privacy Act (CCPA), is essential, necessitating purpose limitation and data minimization \cite{dilmaghani2019privacy,mcintosh2022intercepting,mcintosh2023applying}. Additionally, issues of data sovereignty and copyright emphasize the need for robust encryption, access control, and continuous security assessments \cite{hummel2021data,lukings2022data}. These efforts are critical for maintaining the integrity of AI systems and protecting user privacy in an evolving digital landscape.
	
	\item \textbf{AI Ethics:} The field of AI ethics focuses on fairness, accountability, and societal impact, addresses the surge in ethical challenges posed by AI's increasing complexity and potential misalignment with human values, and requires ethical governance frameworks, multidisciplinary collaborations, and technological solutions \cite{attard2023ethics,hickok2021lessons,huang2022overview,zhou2022ai}. Furthermore, AI Ethics involves ensuring traceability, auditability, and transparency throughout the model development lifecycle, employing practices such as algorithmic auditing, establishing ethics boards, and adhering to documentation standards and model cards \cite{kroll2021outlining,zhou2022ai}. However, the adoption of these initiatives remains uneven, highlighting the ongoing need for comprehensive and consistent ethical practices in AI development and deployment \cite{attard2023ethics}.
	
	\item \textbf{Privacy Preservation:} This domain focuses on maintaining data confidentiality and integrity, employing strategies like anonymization and federated learning to minimize direct data exposure, especially when the rise of generative AI poses risks of user profiling \cite{oseni2021security,stahl2018ethics}. Despite these efforts, challenges such as achieving true anonymity against correlation attacks highlight the complexities in effectively protecting against intrusive surveillance \cite{ma2023trusted,song2020analyzing}. Ensuring compliance with privacy laws and implementing secure data handling practices are crucial in this context, demonstrating the continuous need for robust privacy preservation mechanisms.
\end{itemize}

\subsection{Advanced Learning}
Advanced learning techniques, including self-supervised learning, meta-learning, and fine-tuning, are at the forefront of AI research, enhancing the autonomy, efficiency, and versatility of AI models. 

\begin{itemize}
	\item \textbf{Self-supervised Learning:} This method emphasizes autonomous model training using unlabeled data, reducing manual labeling efforts and model biases \cite{misra2020self,yu2023self,zhai2019s4l}. It incorporates generative models like autoencoders and GANs for data distribution learning and original input reconstruction \cite{chen2019self,jenni2018self,patel2021lt}, and also includes contrastive methods such as SimCLR \cite{chen2020simple} and MoCo \cite{he2020momentum}, designed to differentiate between positive and negative sample pairs. Further, it employs self-prediction strategies, inspired by NLP, using techniques like masking for input reconstruction, significantly enhanced by recent Vision Transformers developments \cite{liu2021tera,pang2022masked,yu2023self}. This integration of varied methods highlights self-supervised learning's role in advancing AI's autonomous training capabilities.
	
	\item \textbf{Meta-learning:} Meta-learning, or `learning to learn', centers on equipping AI models with the ability to rapidly adapt to new tasks and domains using limited data samples \cite{hospedales2021meta,vilalta2002perspective}. This technique involves mastering the optimization process and is critical in situations with limited data availability, to ensure models can quickly adapt and perform across diverse tasks, essential in the current data-driven landscape \cite{al2021data,hu2021task}. It focuses on few-shot generalization, enabling AI to handle a wide range of tasks with minimal data, underlining its importance in developing versatile and adaptable AI systems \cite{baik2021meta,chen2021meta,hu2021task,jamal2019task}.
	
	\item \textbf{Fine Tuning:} Involves customizing pre-trained models to specific domains or user preferences, enhancing accuracy and relevance for niche applications \cite{bakker2022fine,behnia2022ew,wei2021finetuned}. Its two primary approaches are end-to-end fine-tuning, which adjusts all weights of the encoder and classifier \cite{kuang2023federatedscope,nguyen2023efficient}, and feature-extraction fine-tuning, where the encoder weights are frozen to extract features for a downstream classifier \cite{engelbach2023fine,nguyen2023fine,zhou2023regionblip}. This technique ensures that generative models are more effectively adapted to specific user needs or domain requirements, making them more versatile and applicable across various contexts.
	
	\item \textbf{Human Value Alignment:} This emerging aspect concentrates on harmonizing AI models with human ethics and values to ensure that their decisions and actions mirror societal norms and ethical standards, involving the integration of ethical decision-making processes and the adaptation of AI outputs to conform with human moral values \cite{arnold2017value,butlin2021ai,gabriel2021challenge}. This is increasingly important in scenarios where AI interacts closely with humans, such as in healthcare, finance, and personal assistants, to ensure that AI systems make decisions that are not only technically sound, but also ethically and socially responsible, which means human value alignment is becoming crucial in developing AI systems that are trusted and accepted by society \cite{butlin2021ai,nyholm2023responsibility}.
	
\end{itemize}

\subsection{Emerging Trends}
Emerging trends in generative AI research are shaping the future of technology and human interaction, and they indicate a dynamic shift towards more integrated, interactive, and intelligent AI systems, driving forward the boundaries of what is possible in the realm of AI. Key developments in this area include:

\begin{itemize}
	\item \textbf{Multimodal Learning:} Multimodal Learning in AI, a rapidly evolving subdomain, focuses on combining language understanding with computer vision and audio processing to achieve a richer, multi-sensory context awareness \cite{qi2023limitation,wu2023next}. Recent developments like Gemini model have set new benchmarks by demonstrating state-of-the-art performance in various multimodal tasks, including natural image, audio, and video understanding, and mathematical reasoning \cite{gemini2023gemini}. Gemini's inherently multimodal design exemplifies the seamless integration and operation across different information types \cite{gemini2023gemini}. Despite the advancements, the field of multimodal learning still confronts ongoing challenges, such as refining the architectures to handle diverse data types more effectively \cite{bayoudh2021survey,hu2019scalable}, developing comprehensive datasets that accurately represent multifaceted information \cite{bayoudh2021survey,rahate2022multimodal}, and establishing benchmarks for evaluating the performance of these complex systems \cite{che2023multimodal,liang2021multibench}. 
	
	\item \textbf{Interactive and Cooperative AI:} This subdomain aims to enhance the capabilities of AI models to collaborate effectively with humans in complex tasks \cite{ashktorab2020human,shaban2022applied}. This trend focuses on developing AI systems that can work alongside humans, thereby improving user experience and efficiency across various applications, including productivity and healthcare \cite{esmaeilzadeh2021patients,nazar2021systematic,rajawat2021robotic}. Core aspects of this subdomain involve advancing AI in areas such as explainability \cite{mohseni2021multidisciplinary}, understanding human intentions and behavior (theory of mind) \cite{buehler2020theory,ccelikok2019interactive}, and scalable coordination between AI systems and humans, a collaborative approach crucial in creating more intuitive and interactive AI systems, capable of assisting and augmenting human capabilities in diverse contexts \cite{dafoe2020open,shaban2022applied}.

	\item \textbf{AGI Development:} AGI, representing the visionary goal of crafting AI systems that emulate the comprehensive and multifaceted aspects of human cognition, is a subdomain focused on developing AI with the capability for holistic understanding and complex reasoning that closely aligns with the depth and breadth of human cognitive abilities \cite{bubeck2023sparks,fei2022towards,mclean2023risks}. AGI is not just about replicating human intelligence, but also involves crafting systems that can autonomously perform a variety of tasks, demonstrating adaptability and learning capabilities akin to those of humans \cite{bubeck2023sparks,fei2022towards}. The pursuit of AGI is a long-term aspiration, continually pushing the boundaries of AI research and development.

	\item \textbf{AGI Containment:} AGI Safety and Containment acknowledges the potential risks associated with highly advanced AI systems, focused on ensuring that these advanced systems are not only technically proficient but also ethically aligned with human values and societal norms \cite{huang2022overview,mclean2023risks,schuett2023towards}. As we progress towards developing superintelligent systems, it becomes crucial to establish rigorous safety protocols and control mechanisms \cite{schuett2023towards}. Key areas of concern include mitigating representational biases, addressing distribution shifts, and correcting spurious correlations within AI models \cite{schuett2023towards,williams2021understanding}. The objective is to prevent unintended societal consequences by aligning AI development with responsible and ethical standards. 
\end{itemize}

\section{Innovative Horizon of MoE}
\label{sec:innovative-horizon-moe}
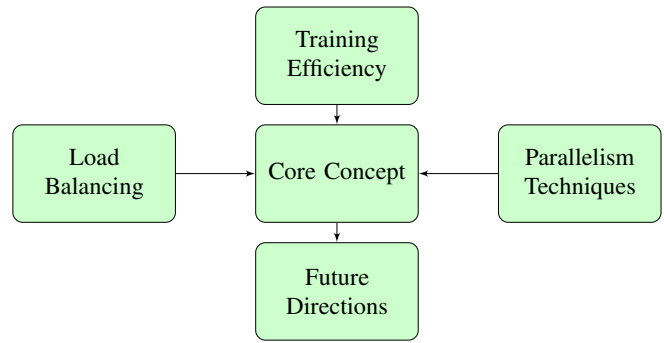
\begin{figure}[t!]
	\centering
	\resizebox{\columnwidth}{!}{
		\begin{tikzpicture}[node distance=1.7cm, auto]
			
			\tikzstyle{block} = [rectangle, draw, fill=green!20, text width=6em, text centered, rounded corners, minimum height=4em]
			\tikzstyle{line} = [draw, -latex']
			
			\node [block] (trainingEfficiency) {Training Efficiency};
			\node [block, below of=trainingEfficiency] (coreConcept) {Core Concept};
			\node [block, left of=coreConcept, node distance=3.5cm] (loadBalancing) {Load Balancing};
			\node [block, right of=coreConcept, node distance=3.5cm] (parallelism) {Parallelism Techniques};
			\node [block, below of=coreConcept] (futureDirections) {Future Directions};
			
			\path [line] (trainingEfficiency) -- (coreConcept);
			\path [line] (loadBalancing) -- (coreConcept);
			\path [line] (parallelism) -- (coreConcept);
			\path [line] (coreConcept) -- (futureDirections);
			
		\end{tikzpicture}
	}
	\caption{Conceptual Diagram of MoE's Innovation}
	\label{fig:moe_innovative_horizon}
\end{figure}

The MoE model architecture represents a pioneering advancement in transformer-based language models, offering unparalleled scalability and efficiency (Fig. \ref{fig:moe_innovative_horizon}). As evidenced by recent models like the 1.6 trillion parameter Switch Transformer \cite{fedus2022switch} and the 8x7B parameter Mixtra \cite{shen2023mixture}, MoE-based designs are rapidly redefining the frontiers of model scale and performance across diverse language tasks.

\subsection{Core Concept and Structure}
\label{subsec:core-concept-structure}
MoE models represent a significant innovation in neural network design, offering enhanced scalability and efficiency in training and inference \cite{rajbhandari2022deepspeed,shen2022se,zhou2022mixture}. At their core, MoE models utilize a sparsity-driven architecture by replacing dense layers with sparse MoE layers comprising multiple expert networks, where each expert is dedicated to a specific subset of the training data or task, and a trainable gating mechanism dynamically allocates input tokens to these experts, thereby optimizing computational resources and effectively adapting to the task's complexity \cite{chen2023octavius,du2022glam,zhou2022mixture}. MoE models demonstrate a substantial advantage in terms of pretraining speed, outperforming dense models \cite{chen2023octavius,rajbhandari2022deepspeed}. However, they face challenges in fine-tuning and require substantial memory for inference due to the necessity of loading all experts into Video Random Access Memory (VRAM) \cite{hwang2023tutel,wang2022adamix,zhou2022mixture}. The structure of MoE involves alternating transformer layers with router layers containing gating networks for expert routing, leading to an architecture that allows significant parameter scaling and advanced specialization in problem-solving \cite{chen2023sparse,du2022glam}. 

A distinguishing characteristic of MoE models is their flexibility in managing large datasets, capable of amplifying model capacity by over a thousand times while only experiencing minor reductions in computational efficiency \cite{hwang2023tutel,zhu2022multi}. The Sparsely-Gated Mixture-of-Experts Layer, a key component of these models, comprises numerous simple feed-forward expert networks and a trainable gating network responsible for expert selection, which can facilitate the dynamic and sparse activation of experts for each input instance, maintaining high computational efficiency \cite{chi2022representation,gupta2022sparsely,zhou2022mixture}.

Recent advancements in MoE models, such as those in the Switch Transformer, have highlighted the significant benefits of intelligent routing, when the router's ability to intelligently route tokens to appropriate experts confers considerable advantages to MoE models, allowing them to scale up model sizes while keeping compute time constant \cite{dikkala2023benefits,dryden2022spatial,you2022speechmoe2}. Experimental evidence suggests that routers learn to route inputs according to data clusters, demonstrating their potential in real-world applications \cite{dikkala2023benefits,hwang2023tutel}. The core concept and structure of MoE models lie in their dynamic routing and specialization capabilities, offering promising avenues for scaling up neural networks and enhancing their efficiency and adaptability in various tasks, but the robustness of the router must be protected against adversarial attacks \cite{hwang2023tutel,puigcerver2022adversarial}.

\subsection{Training and Inference Efficiency}
\label{subsec:training-inference-efficiency}
MoE models, notably Mixtral 8x7B, are renowned for their superior pretraining speed compared to dense models, yet they face hurdles in fine-tuning and demand considerable VRAM for inference, owing to the requirement of loading all experts \cite{hwang2023tutel,wang2022adamix,zhou2022mixture}. Recent advancements in MoE architecture have resulted in notable training cost efficiencies, especially in encoder-decoder models, with evidence showing cost savings of up to fivefold in certain contexts when compared to dense models \cite{du2022glam,hwang2023tutel,puigcerver2022adversarial,rajbhandari2022deepspeed}. Innovations like DeepSpeed-MoE \cite{rajbhandari2022deepspeed} offered new architectural designs and model compression, decreasing the MoE model size by approximately 3.7x and optimizing inference to achieve up to 7.3x better latency and cost efficiency. The progression in distributed MoE training and inference, notably with innovations like Lina \cite{li2023accelerating}, has effectively tackled the all-to-all communication bottleneck by enhancing tensor partitioning, which not only improves all-to-all communication and training step time, but also optimizes resource scheduling during inference, leading to a substantial reduction in training step time by up to 1.73 times and lowering the 95th percentile inference time by an average of 1.63 times compared to existing systems. These developments have marked a crucial shift in the large model landscape, from dense to sparse MoE models, expanding the potential applications of AI by training higher-quality models with fewer resources.

\subsection{Load Balancing and Router Optimization}
\label{subsec:load-balancing-router}
Effective load balancing is essential in MoE models to guarantee a uniform distribution of computational load among experts, with the router network in MoE layers, responsible for selecting the appropriate experts for processing specific tokens, playing a pivotal role in achieving this balance, which is fundamental to the stability and overall performance of MoE models \cite{chi2022representation,hwang2023tutel,shen2022se,wu2022residual,zhou2022mixture}. Developments in router Z-loss regularization techniques plays a crucial role in addressing expert imbalance in MoE models by fine-tuning the gating mechanism, ensuring a more equitable workload distribution across experts and fostering a stable training environment, thereby enhancing model performance and reducing training time and computational overhead \cite{zoph2022designing,zoph2022st}. Concurrently, the integration of expert capacity management strategies, emerges as a crucial approach in MoE models to regulate the processing abilities of individual experts by setting thresholds on the number of tokens each can handle, effectively averting bottlenecks and ensuring a more efficient and streamlined model operation, leading to improved training processes and heightened performance during complex computational tasks \cite{chi2022representation,chow2022mixture,hwang2023tutel}.

\subsection{Parallelism and Serving Techniques}
\label{subsec:parallelism-serving}
Recent developments in MoE models highlighted their efficiency in parallelism and serving techniques, significantly influencing large-scale neural networks. DeepSpeed-MoE, for instance, introduces advanced parallelism modes like data parallelism, tensor-slicing for non-expert parameters, and expert parallelism for expert parameters, enhancing model efficiency, as their approach optimizes both latency and throughput in MoE model inference, offering scalable solutions in production environments using multiple Graphics Processing Unit (GPU) devices \cite{rajbhandari2022deepspeed}. MoE models, versatile in applications like multilingual tasks and coding, demonstrated impressive capabilities in handling complex tasks due to their ensemble-like structure within a single framework \cite{fan2022m3vit,zadouri2023pushing,zhu2022uni}. Notably, models like Mixtral and Switch Transformer, with over 1.6 trillion parameters, achieved computational efficiency equivalent to a 10 billion-parameter dense model, because they benefited from the sublinear scaling of MoE compute versus model size, leading to substantial accuracy gains within fixed compute budgets \cite{du2022glam,hwang2023tutel,rajbhandari2022deepspeed,zhou2022mixture}. Moreover, DeepSpeed-MoE included model compression techniques, reducing model size by up to 3.7x while maintaining accuracy, and an end-to-end MoE training and inference solution, part of the DeepSpeed library, which was instrumental in serving large-scale MoE models with enhanced speed and cost-efficiency \cite{rajbhandari2022deepspeed}. These innovations open new directions in AI, shifting from dense to sparse MoE models, where training and deploying higher-quality models with fewer resources become more widely achievable.

\subsection{Future Directions and Applications}
\label{subsec:future-directions-applications}
Emerging research on MoE architectures could focus on advancing sparse fine-tuning techniques, exploring instruction tuning methods, and improving routing algorithms to fully utilize performance and efficiency gains. As models scale over one billion parameters, MoE represents a paradigm shift for vastly expanding capabilities across scientific, medical, creative, and real-world applications. Frontier work could also aim to refine auto-tuning of hyperparameters during fine-tuning to optimize accuracy, calibration, and safety. MoE research continues to push model scale limits while maintaining specialization for transfer learning. Adaptive sparse access allows coordinating thousands of experts to cooperate on tasks ranging from reasoning to open domain dialogue. Continued analysis of routing mechanisms seeks to balance load across experts and minimize redundant computation. As the AI community further investigates MoE methods at scale, these models hold promise for new breakthroughs in language, code generation, reasoning, and multimodal applications. There is great interest in evaluating implications across education, healthcare, financial analysis, and other fields. Outcomes may yield insights not only into model optimization but also for understanding principles behind combinatorial generalization.

\section{Speculated Capabilities of Q*}
\label{sec:speculated-capabilities-q-star}
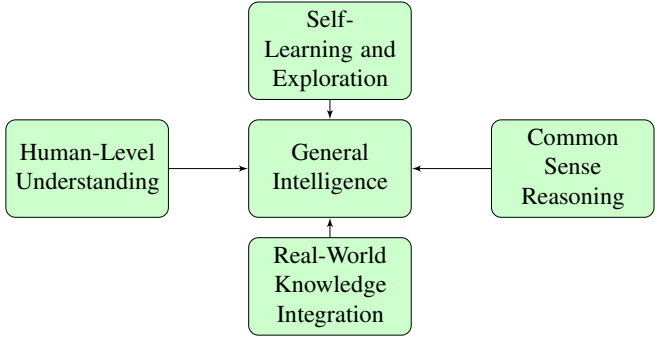
\begin{figure}[t!]
	\centering
	\resizebox{\columnwidth}{!}{
		\begin{tikzpicture}[node distance=1.7cm, auto]
			
			\tikzstyle{block} = [rectangle, draw, fill=green!20, text width=6em, text centered, rounded corners, minimum height=4em]
			\tikzstyle{line} = [draw, -latex']
			
			\node [block] (selfLearning) {Self-Learning and Exploration};
			\node [block, below of=selfLearning] (genIntelligence) {General Intelligence};
			\node [block, left of=genIntelligence, node distance=3.5cm] (humanUnderstanding) {Human-Level Understanding};
			\node [block, right of=genIntelligence, node distance=3.5cm] (commonSense) {Common Sense Reasoning};
			\node [block, below of=genIntelligence] (knowledgeIntegration) {Real-World Knowledge Integration};
			
			\path [line] (selfLearning) -- (genIntelligence);
			\path [line] (humanUnderstanding) -- (genIntelligence);
			\path [line] (commonSense) -- (genIntelligence);
			\path [line] (knowledgeIntegration) -- (genIntelligence) ;
			
		\end{tikzpicture}
	}
	\caption{Conceptual Diagram of Speculated Q* Capabilities}
	\label{fig:q_star_capabilities}
\end{figure}

In the burgeoning realm of AI, the anticipated Q* project stands as a beacon of potential breakthroughs, heralding advancements that could redefine the landscape of AI capabilities (Fig. \ref{fig:q_star_capabilities}).

\subsection{Enhanced General Intelligence}
\label{subsec:enhanced-general-intelligence}
Q*'s development in the arena of general intelligence represents a paradigm shift from specialized to holistic AI, indicating a broadening of the model’s cognitive abilities akin to human intelligence. This advanced form of general intelligence involves integrating diverse neural network architectures and machine learning techniques, enabling the AI to process and synthesize multifaceted information seamlessly. The universal adapter approach, mirroring models like T0, could endow Q* with the capability to rapidly assimilate knowledge from various domains. This method allows Q* to learn adaptable module plugins, enhancing its ability to tackle new data types while preserving existing skills, leading to an AI model that combines narrow specializations into a comprehensive, adaptive, and versatile reasoning system. The corresponding quasi-mathematical formulation can be expressed as:

\begin{equation}
	EGI(Q*) = \bigoplus_{i=1}^{n} (NN_i \odot MLT_i)
\end{equation}

Where:
\begin{itemize}
	\item \(EGI\): \enquote{Enhanced General Intelligence}
	\item \( NN_i \): a diverse set of neural network architectures.
	\item \( MLT_i \): various machine learning techniques.
	\item \( \bigoplus \): the integration of these components.
	\item \( \odot \): a functional interaction between neural networks and machine learning techniques.
\end{itemize}

Such advancements in AI suggest the emergence of an intelligence that not only parallels but potentially exceeds human cognitive flexibility, with far-reaching implications in facilitating cross-disciplinary innovations and complex problem-solving. The speculated capabilities of Q* bring forth complex ethical implications and governance challenges. As AI systems approach higher levels of autonomy and decision-making, it is crucial to establish robust ethical frameworks and governance structures to ensure responsible and transparent AI development. This involves mitigating potential risks associated with advanced AI capabilities, emphasizing the need for comprehensive and dynamic ethical guidelines that evolve in tandem with AI advancements.

\subsection{Advanced Self-Learning and Exploration}
\label{subsec:advanced-self-learning}
In the realm of advanced AI development, Q* is anticipated to represent a significant evolution in self-learning and exploration capabilities. It is speculated to utilize sophisticated Policy Neural Networks (NNs), similar to those in AlphaGo, but with substantial enhancements to handle the complexities of language and reasoning tasks. These networks are expected to employ advanced reinforcement learning techniques like Proximal Policy Optimization (PPO), which stabilizes policy updates and improves sample efficiency, a crucial factor in autonomous learning. The integration of these NNs with cutting-edge search algorithms, potentially including novel iterations of Tree or Graph of Thought, is predicted to enable Q* to autonomously navigate and assimilate complex information. This approach might be augmented with graph neural networks to bolster meta-learning capacities, allowing Q* to rapidly adapt to new tasks and environments while retaining previously acquired knowledge. The corresponding quasi-mathematical formulation can be represented as:

\begin{equation}
	ASLE(Q*) = RL(PNN, SA) \times GNN
\end{equation}

Where:
\begin{itemize}
	\item \(ASLE\): \enquote{Advanced Self-Learning and Exploration}
	\item \( RL \): to reinforcement learning algorithms, particularly Proximal Policy Optimization (PPO).
	\item \( PNN \): Policy Neural Networks, adapted for language and reasoning tasks.
	\item \( SA \): sophisticated search algorithms, like Tree or Graph of Thought.
	\item \( GNN \): the incorporation of Graph Neural Networks for meta-learning.
	\item \( \times \): the cross-functional enhancement of RL with GNN.
\end{itemize}

Such capabilities indicate a model not limited to understanding existing data but equipped to actively seek and synthesize new knowledge, effectively adapting to evolving scenarios without the need for frequent retraining. This signifies a leap beyond current AI models, embedding a level of autonomy and efficiency previously unattained.

\subsection{Superior Human-Level Understanding}
\label{subsec:superior-human-level-understanding}
Q*'s aspiration to achieve superior human-level understanding is speculated to hinge on an advanced integration of multiple neural networks, including a Value Neural Network (VNN), paralleling the evaluative components found in systems like AlphaGo. This network would extend beyond assessing accuracy and relevance in language and reasoning processes, delving into the subtleties of human communication. The model's deep comprehension capabilities may be enhanced by advanced natural language processing algorithms and techniques, such as those found in transformer architectures like DeBERTa. These algorithms would empower Q* to interpret not just the text but also the nuanced socio-emotional aspects such as intent, emotion, and underlying meanings. Incorporating sentiment analysis and natural language inference, Q* could navigate layers of socio-emotional insights, including empathy, sarcasm, and attitude. The corresponding quasi-mathematical formulation can be expressed as:

\begin{equation}
	SHLU(Q*) = \sum_{alg \in NLP} (VNN \oplus alg)
\end{equation}

Where:
\begin{itemize}
	\item \( SHLU \): \enquote{Superior Human-Level Understanding}.
	\item \( VNN \): the Value Neural Network, similar to evaluative components in systems like AlphaGo.
	\item \( NLP \): a set of advanced NLP algorithms.
	\item \( \oplus \): the combination of VNN evaluation with NLP algorithms.
	\item \( alg \): individual algorithms within the NLP set.
\end{itemize}

This level of understanding, surpassing current language models, would position Q* to excel in empathetic, context-aware interactions, thus enabling a new echelon of personalization and user engagement in AI applications.

\subsection{Advanced Common Sense Reasoning}
\label{subsec:advanced-common-sense-reasoning}
Q*'s anticipated development in advanced common sense reasoning is predicted to integrate sophisticated logic and decision-making algorithms, potentially combining elements of symbolic AI and probabilistic reasoning. This integration aims to endow Q* with an intuitive grasp of everyday logic and an understanding akin to human common sense, thus bridging a significant gap between artificial and natural intelligence. Enhancements in Q*'s reasoning abilities might involve graph-structured world knowledge, incorporating physics and social engines similar to those in models like CogSKR. This approach, grounded in physical reality, is expected to capture and interpret the everyday logic often absent in contemporary AI systems. By leveraging large-scale knowledge bases and semantic networks, Q* could effectively navigate and respond to complex social and practical scenarios, aligning its inferences and decisions more closely with human experiences and expectations. The corresponding quasi-mathematical formulation can be represented as:

\begin{equation}
	ACSR(Q*) = LogicAI \odot ProbAI \odot WorldK
\end{equation}

Where:
\begin{itemize}
	\item \( ACSR \): \enquote{Advanced Common Sense Reasoning}.
	\item \( LogicAI \) and \( ProbAI \): symbolic AI and probabilistic reasoning components, respectively.
	\item \( WorldK \): the integration of graph-structured world knowledge.
	\item \( \odot \): the integrated operation of these elements for common sense reasoning.
\end{itemize}

\subsection{Extensive Real-World Knowledge Integration}
\label{subsec:extensive-real-world-knowledge}
Q*'s approach to integrating extensive real-world knowledge is speculated to involve the use of advanced formal verification systems, which would provide a robust basis for validating its logical and factual reasoning. This method, when coupled with sophisticated neural network architectures and dynamic learning algorithms, would enable Q* to engage deeply with the complexities of the real world, transcending conventional AI limitations. Additionally, Q* might employ mathematical theorem proving techniques for validation, ensuring that its reasoning and outputs are not only accurate but also ethically grounded. The incorporation of Ethics classifiers in this process further strengthens its capacity to deliver reliable and responsible understanding and interaction with real-world scenarios. The corresponding quasi-mathematical formulation can be represented as:

\begin{equation}
	ERWKI(Q*) = FVS \otimes NN \otimes LTP \otimes EC
\end{equation}

Where:
\begin{itemize}
	\item \( ERWKI \): \enquote{Extensive Real-World Knowledge Integration}.
	\item \( FVS \): Formal Verification Systems.
	\item \( NN \):  neural network architectures.
	\item \( LTP \): mathematical theorem proving for logical and factual validation.
	\item \( EC \): the incorporation of Ethics classifiers.
	\item \( \otimes \): the comprehensive integration for knowledge synthesis and ethical alignment.
\end{itemize}

Furthermore, the speculated capabilities of Q* have the potential to significantly reshape the job market and labor dynamics. With its advanced functionalities, Q* could automate complex tasks, leading to a shift in job requirements and the emergence of new skill demands. This necessitates a re-evaluation of workforce strategies and educational paradigms, aligning them with the evolving technological landscape and ensuring that the workforce is equipped to interact with and complement these advanced AI systems.

\section{Projected Capabilities of AGI}
\label{sec:projected-capabilities-agi}
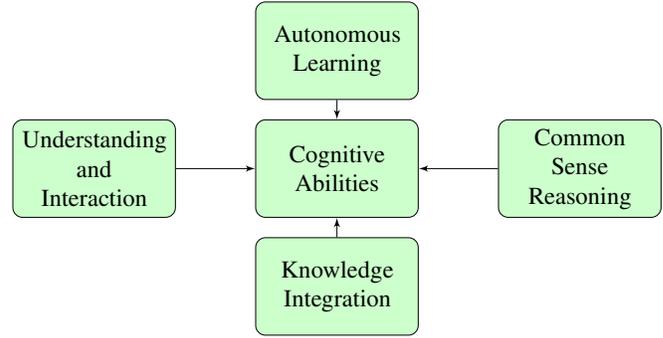
\begin{figure}[t!]
	\centering
	\resizebox{\columnwidth}{!}{
		\begin{tikzpicture}[node distance=1.7cm, auto]
			
			\tikzstyle{block} = [rectangle, draw, fill=green!20, text width=6em, text centered, rounded corners, minimum height=4em]
			\tikzstyle{line} = [draw, -latex']
			
			\node [block] (autonomousLearning) {Autonomous Learning};
			\node [block, below of=autonomousLearning] (cognitiveAbilities) {Cognitive Abilities};
			\node [block, left of=cognitiveAbilities, node distance=3.5cm] (understandingInteraction) {Understanding and Interaction};
			\node [block, right of=cognitiveAbilities, node distance=3.5cm] (commonSense) {Common Sense Reasoning};
			\node [block, below of=cognitiveAbilities] (knowledgeIntegration) {Knowledge Integration};
			
			\path [line] (autonomousLearning) -- (cognitiveAbilities);
			\path [line] (understandingInteraction) -- (cognitiveAbilities);
			\path [line] (commonSense) -- (cognitiveAbilities);
			\path [line]  (knowledgeIntegration) -- (cognitiveAbilities) ;
			
		\end{tikzpicture}
	}
	\caption{Conceptual Diagram of Projected AGI Capabilities}
	\label{fig:agi_capabilities}
\end{figure}

AGI stands as a transformative leap in AI, endeavoring to mirror human cognitive abilities in a software paradigm (Fig. \ref{fig:agi_capabilities}). AGI's evolution is marked by advanced self-learning capabilities, utilizing policy neural networks and sophisticated reinforcement learning techniques for autonomous adaptation. The integration of algorithms like Tree/Graph of Thought with these networks suggests a future where AGI can independently acquire and apply knowledge across diverse domains.

\subsection{Revolution in Autonomous Learning}
\label{subsec:revolution-autonomous-learning}
AGI is anticipated to revolutionize self-learning and exploration \cite{bubeck2023sparks,dou2023towards,fei2022towards,mclean2023risks}. By incorporating methods like PPO, AGI models are positioned to achieve a level of autonomous learning and problem-solving that exceeds the current AI models' dependence on training data, indicating a potential paradigm shift towards reducing the need for frequent retraining and facilitating dynamic adaptation in response to evolving scenarios \cite{alagha2022target,jia2022improving}.

\subsection{Broadening of Cognitive Abilities}
\label{subsec:broadening-cognitive-abilities}
Envisaged to integrate various architectures, AGI could promise a level of general intelligence that replicates the multifaceted nature of human cognition \cite{bubeck2023sparks,simeone2022unknown}. The universal adapter approach, mirroring models like GPT and BERT, could facilitate rapid assimilation of diverse information, positioning AGI as a system capable of performing tasks across multiple domains with an adaptability akin to human intellect \cite{bubeck2023sparks,nair2022bridging}. While AGI's full capabilities remain speculative, current trends suggest its potential application in advanced healthcare diagnostics, which is evidenced by recent breakthroughs in AI-driven predictive medicine models, indicating AGI's potential to revolutionize medical diagnosis and treatment.

\subsection{Elevating Understanding and Interaction}
\label{subsec:elevating-understanding-interaction}
AGI is projected to achieve an unparalleled understanding of human language and socio-emotional subtleties, leveraging algorithms like those in transformer architectures, which would enable AGI to engage in complex, empathetic, and contextually aware interactions, suggesting potential applications that revolutionize how AI systems communicate and interact \cite{bubeck2023sparks,dou2023towards,jarrahi2023artificial}.

\subsection{Advanced Common Sense Reasoning}
\label{subsec:advanced-common-sense}
Symbolic AI and probabilistic reasoning, integrated into AGI, could imbue these systems with an innate grasp of common sense, to bridge the gap between artificial and natural intelligence, enabling AGI to navigate and respond effectively to real-world scenarios with reasoning aligned closely with human thought processes \cite{bubeck2023sparks,edwards2022functional,mccarthy2022artificial}.

\subsection{Holistic Integration of Knowledge}
\label{subsec:holistic-integration-knowledge}
AGI's potential in integrating extensive real-world knowledge, guided by formal verification systems, hints at future capabilities where AGI's outputs are not only accurate but ethically grounded, suggesting AGI's ability for responsible interaction with real-world complexities  \cite{bubeck2023sparks,jarrahi2023artificial}. The projected capabilities of AGI extend to addressing significant global challenges, such as climate change, in which AGI's advanced data analysis and predictive modeling can play a better and more crucial role in environmental monitoring, forecasting climate patterns, and devising sustainable solutions, contributing significantly to global ecological efforts \cite{bubeck2023sparks,fei2022towards,mclean2023risks}.

\subsection{Challenges and Opportunities in AGI Development}
\label{subsec:challenges-opportunities-agi}
The development of AGI encompasses both challenges and opportunities. While AGI promises productivity boosts in creative fields and innovations in cross-modal generation techniques, substantial challenges like data bias, computational efficiency, and ethical implications persist \cite{huang2022overview,mclean2023risks}. These challenges necessitate a balanced approach in AGI development, focusing on data curation, efficient systems, and societal impacts \cite{simeone2022unknown}.

In the context of AGI development, experts from various domains caution against overestimating current AI capabilities and highlight the gap between the theoretical framework of AGI and the practical realities of today's AI \cite{friederich2023symbiosis,mclean2023risks}. The envisioned autonomy and cognitive abilities of AGI separate it from current AI models, suggesting a future where AI systems could perform tasks across various domains without human intervention \cite{bubeck2023sparks}. This development trajectory underscores the importance of ethical considerations and technological breakthroughs in AGI's journey towards becoming a transformative force in society \cite{huang2022overview,mclean2023risks}. While projecting the timeline for achieving true AGI remains speculative, recognizing potential roadblocks is crucial, such as the current limitations in computational power, and the complexity of replicating human-like cognitive abilities. These emphasize the need for sustained research and ethical considerations in the pursuit of AGI, ensuring responsible and conscientious development.

\section{Impact Analysis on Generative AI Research Taxonomy}
\label{sec:impact-analysis-llm}
With the advent of advanced AI developments such as MoE, multimodality, and AGI, the landscape of Generative AI research is undergoing a significant transformation. This section analyzes how these developments are reshaping the research taxonomy in generative AI.

\subsection{Criteria for Impact Analysis}
\label{subsec:criteria-impact-analysis}
The continuously evolving landscape of Generative AI, which instigates transformative changes across various research domains, necessitates a systematic evaluation of these advancements' influence, for which we have established a set of criteria detailed in Table \ref{table:criteria-impact-analysis}, serving as analytical lenses to quantify and categorize the impact, deeply rooted in the dynamic interplay between technological progress and the evolving paradigms of research focus areas. Our analysis framework has been constructed on a gradient scale ranging from emergent to obsolete, reflecting the extent to which areas of Generative AI research are being reshaped. The categorization into five distinct classes allows for a complex assessment, acknowledging that not all areas will be uniformly affected. This multi-tiered approach is informed by historical patterns of technological disruption and the adaptability of scientific inquiry.

At the apex of our evaluative hierarchy, `Emerging Direction' encapsulates the advent of uncharted research vistas, propelled by ongoing AI breakthroughs, which is predicated not on conjecture, but on a historical continuum of AI evolution, where each surge in technological power unfurls new scientific enigmas and avenues \cite{makridakis2017forthcoming,pal2023ai}. `Areas Requiring Redirection' denote research spheres that, though established, find themselves at an inflection point, necessitating a strategic pivot to assimilate emergent AI paradigms and an overhaul of traditional methodologies, akin to the transition from rule-based expert systems to adaptive machine learning frameworks \cite{makridakis2017forthcoming,verma2021artificial}. The `Still Relevant' classification affirms the tenacity of select research domains that, by addressing persistent scientific inquiries or through their inherent malleability, remain impervious to the tides of AI innovation \cite{verma2021artificial}. In contrast, domains categorized as `Likely to Become Redundant' confront potential obsolescence, inviting strategic foresight and resource reallocation to forestall scientific stagnation \cite{budhwar2023human}. Lastly, `Inherently Unresolvable' challenges serve as a sobering reminder of the perpetual dilemmas within AI research that defy resolution, rooted in the complex web of human ethics and cultural diversity, thus anchoring the pursuit of AI within the intractable tapestry of human values and societal imperatives \cite{telkamp2022implications,zhou2023interpretable}.

\begin{table*}[tph!]
	\centering
	\caption{Criteria for Analyzing Impact on Generative AI Research}
	\label{table:criteria-impact-analysis}
	\resizebox{\textwidth}{!}{ 
		\begin{tabular}{|c|p{2cm}|c|p{5cm}|p{7cm}|}
			\hline
			\textbf{Symbol} & \textbf{Criteria} & \textbf{Score}& \textbf{Definition} & \textbf{Justification} \\ \hline
			$\nearrow$ & Emerging Direction & 5 & New research areas expected to arise as a direct consequence of AI advancements. & Emphasizes novel research domains emerging from AI breakthroughs \cite{makridakis2017forthcoming,pal2023ai}. \\ \hline
			$\hookrightarrow$ & Requiring Redirection & 4 & Areas that need to shift focus or methodology to stay relevant with new AI developments. & Technological shifts necessitate reevaluation and redirection in AI research \cite{makridakis2017forthcoming,verma2021artificial}. \\ \hline
			$\leftrightarrow$ & Still Relevant & 3 & Areas where the advancements have minimal or no impact, maintaining their current status and methodologies. & Observes the persistence of certain AI research areas despite technological advancements \cite{verma2021artificial}. \\ \hline
			$\searrow$ & Likely to Become Redundant & 2 & Areas that may lose relevance or become obsolete with the advent of new AI technologies. & Discusses rapid obsolescence in AI methodologies due to new technologies \cite{budhwar2023human}. \\ \hline
			$\bigtriangleup$ & Inherently Unresolvable & 1 & Challenges that may remain unresolved due to complexities like subjective human perspectives and diverse cultural values. & Inherent difficulties in issues such as aligning AI with diverse human values and ethics \cite{telkamp2022implications,zhou2023interpretable}. \\ \hline
		\end{tabular}
	}
\end{table*}

\subsection{Overview of Impact Analysis}
\label{subsec:overview-impact-analysis}
This subsection offers a detailed overview of the impact analysis carried out on the research taxonomy within the realm of generative AI, with a specific focus on recent progress in MoE, multimodality, and AGI, aiming to evaluate the impact of these innovative developments on various facets of generative AI research, ranging from model architecture to sophisticated learning methodologies, and includes both quantitative and qualitative assessments across a multitude of domains and subdomains in LLM research, shedding light on the extent to which each area is influenced by these technological advancements. This evaluation considered factors such as the emergence of new research directions, the necessity for redirection in existing research areas, the continued relevance of certain methodologies, and the potential redundancy of others, and has encapsulated in Table \ref{table:llm-taxonomy-impact}.

\begin{table*}[tp!]
	\centering
	\caption{Impact of MoE, Multimodality, and AGI on Generative AI Research}
	\label{table:llm-taxonomy-impact}
	\resizebox{\textwidth}{!}{
		\begin{tabular}{|p{4.5cm}|p{4cm}|c|c|c|c|}
			\hline
			\textbf{Domain} & \textbf{Subdomain} & \textbf{MoE} & \textbf{Multimodality} & \textbf{AGI} & \textbf{Overall Score} \\ \hline
			Model Architecture & Transformer Models & $\hookrightarrow$ (4) & $\leftrightarrow$ (3) & $\hookrightarrow$ (4) & 11 \\ \cline{2-6}
			
			& Recurrent Neural Networks & $\searrow$ (2) & $\leftrightarrow$ (3) & $\searrow$ (2) & 7 \\ \cline{2-6}
			
			& Mixture of Experts & $\leftrightarrow$ (3) & $\nearrow$ (5) & $\hookrightarrow$ (4) & 12   \\ \cline{2-6}
			
			& Multimodal Models & $\nearrow$ (5) & $\leftrightarrow$ (3) & $\nearrow$ (5) & 13 \\ \hline
			
			Training Techniques & Supervised Learning & $\hookrightarrow$ (4) & $\leftrightarrow$ (3) & $\searrow$ (2) & 9 \\  \cline{2-6}
			
			& Unsupervised Learning & $\hookrightarrow$ (4) & $\leftrightarrow$ (3) & $\hookrightarrow$ (4) & 11 \\ \cline{2-6}
			
			& Reinforcement Learning & $\leftrightarrow$ (3) & $\hookrightarrow$ (4) & $\nearrow$ (5) & 12 \\  \cline{2-6}

			& Transfer Learning & $\leftrightarrow$ (3) & $\nearrow$ (5) & $\hookrightarrow$ (4) & 12 \\ \hline
			
			Application Domains & Natural Language Understanding & $\leftrightarrow$ (3) & $\leftrightarrow$ (3) & $\nearrow$ (5) & 11\\ \cline{2-6}
			
			& Natural Language Generation & $\leftrightarrow$ (3) & $\hookrightarrow$ (4) & $\nearrow$ (5) & 12 \\ \cline{2-6}
			
			& Conversational AI & $\hookrightarrow$ (4) & $\nearrow$ (5) & $\nearrow$ (5) & 14 \\  \cline{2-6}
			
			& Creative AI & $\hookrightarrow$ (4) & $\nearrow$ (5) & $\nearrow$ (5) & 14 \\ \hline
			
			Compliance and Ethical Considerations & Bias Mitigation & $\hookrightarrow$ (4) & $\hookrightarrow$ (4) & $\nearrow$ (5) & 13 \\ \cline{2-6}
			
			& Data Security & $\leftrightarrow$ (3) & $\leftrightarrow$ (3) & $\leftrightarrow$ (3) & 9  \\ \cline{2-6}
			
			& AI Ethics & $\hookrightarrow$ (4) & $\hookrightarrow$ (4) & $\bigtriangleup$ (1) & 9 \\ \cline{2-6}
			
			& Privacy Preservation & $\hookrightarrow$ (4) & $\hookrightarrow$ (4) & $\hookrightarrow$ (4) & 12 \\  \hline
			
			Advanced Learning & Self-supervised Learning & $\hookrightarrow$ (4) & $\nearrow$ (5) & $\leftrightarrow$ (3) & 12 \\  \cline{2-6}
			
			& Meta-learning & $\leftrightarrow$ (3) & $\leftrightarrow$ (3) & $\nearrow$ (5) & 11 \\ \cline{2-6}
			
			& Fine Tuning & $\leftrightarrow$ (3) & $\leftrightarrow$ (3) & $\searrow$ (2) & 8 \\ \cline{2-6}

			& Human Value Alignment & $\bigtriangleup$ (1) & $\bigtriangleup$ (1) & $\bigtriangleup$ (1) & 3    \\ 			\hline
			
			Emerging Trends & Multimodal Learning & $\nearrow$ (5) & $\leftrightarrow$ (3) & $\nearrow$ (5) & 13   \\ \cline{2-6}
			
			& Interactive and Cooperative AI & $\hookrightarrow$ (4) & $\leftrightarrow$ (3) & $\nearrow$ (5) & 12 \\ \cline{2-6}
			
			& AGI Development & $\hookrightarrow$ (4) & $\hookrightarrow$ (4) & $\leftrightarrow$ (3) & 11 \\ \cline{2-6}
			
			& AGI Containment & $\bigtriangleup$ (1) & $\bigtriangleup$ (1) & $\nearrow$ (5) & 7  \\ \hline
		\end{tabular}
	}
\end{table*}

\subsubsection{Impact On Model Architecture}
\label{subsubsec:impact-model-architecture}
Transformer Models have been scored with a redirection requirement ($\hookrightarrow$) of 4 in both MoE and AGI, and a relevance ($\leftrightarrow$) of 3 in multimodality, leading to an overall score of 11. These models, forming the backbone of many current AI architectures, continue to be relevant for handling complex input sequences. However, the emergence of MoE and AGI indicates a shift towards more dynamic and specialized architectures. While transformers remain essential, there is a need for them to evolve and integrate with these advanced systems for enhanced performance and adaptability.

Recurrent Neural Networks (RNNs) are facing a potential decline in relevance, as indicated by their scores: likely to become redundant ($\searrow$) 2 in both MoE and AGI contexts and still relevant ($\leftrightarrow$) 3 in multimodality, totaling a score of 7. Although effective for sequence processing, RNNs are challenged by their limitations in handling long-range dependencies and lower efficiency compared to newer models like transformers. They may retain some relevance in multimodal tasks involving sequential data but are generally overshadowed by more advanced architectures.

The MoE models have scored a consistent relevance ($\leftrightarrow$) of 3 in their own development and a score of 5 ($\nearrow$) in multimodality, combined with a redirection score ($\hookrightarrow$) of 4 in the context of AGI, amounting to an overall score of 12. MoE models are at the forefront of emerging research in multimodality due to their ability to handle diverse data types. For AGI, these models will require adjustments to effectively integrate into systems exhibiting general intelligence, especially in areas beyond their initial specialization.

Multimodal Models have received high scores for emerging research directions ($\nearrow$) of 5 in both MoE and AGI contexts, alongside a score of 3 ($\leftrightarrow$) for current relevance in multimodality, culminating in an overall score of 13. The integration of MoE and the pursuit of AGI are opening new pathways for research in multimodal models. These developments are crucial for enhancing the ability to process and synthesize information from multiple modalities, a key aspect for both specialized and generalized AI systems.

\subsubsection{Impact On Training Techniques}
Supervised Learning has been assigned a redirection score ($\hookrightarrow$) of 4, a relevance score ($\leftrightarrow$) of 3 in multimodality, and a score indicating potential redundancy ($\searrow$) of 2 in the context of AGI, culminating in an overall score of 9. While supervised learning requires adaptation to fit the MoE framework, it remains relevant for multimodal AI models that depend on labeled data. However, with the shift towards more autonomous learning methods in AGI, the dependence on extensive labeled datasets typically associated with supervised learning may diminish, leading to its potential decrease in significance.

Unsupervised Learning scores a redirection requirement ($\hookrightarrow$) of 4 in both MoE and AGI contexts and maintains its relevance ($\leftrightarrow$) with a score of 3 in multimodality, resulting in a total score of 11. In the MoE architecture, unsupervised learning methods may need adjustments, particularly in managing dynamic task allocation. It remains crucial for understanding unlabeled data across various modalities. In AGI, unsupervised learning is expected to evolve beyond traditional techniques, focusing on more advanced self-discovery and intrinsic learning mechanisms.

Reinforcement Learning is rated as still relevant ($\leftrightarrow$) with a score of 3 in MoE, requiring redirection ($\hookrightarrow$) with a score of 4 in multimodality, and identified as an emerging research area ($\nearrow$) with a score of 5 in AGI, giving it a total score of 12. This technique continues to play a significant role in optimizing MoE model structures. In the realm of multimodality, it necessitates a strategic shift to effectively manage complex interactions between different modalities. As for AGI, reinforcement learning is emerging as a crucial area, particularly in the development of autonomous systems that learn from their environment.

Transfer Learning receives a consistent relevance score ($\leftrightarrow$) of 3 in MoE, a high score for emerging research directions ($\nearrow$) of 5 in multimodality, and a redirection requirement ($\hookrightarrow$) of 4 in AGI, accumulating to an overall score of 12. It remains important in the MoE framework for leveraging knowledge across different experts. In multimodal contexts, transfer learning is becoming increasingly crucial as it facilitates the transfer of learning between different modalities. With the evolution of AGI, this technique is expected to undergo significant changes to cater to broader and more generalized knowledge applications.

\subsubsection{Impact On Application Domains}
Natural Language Understanding holds steady relevance ($\leftrightarrow$) with a score of 3 in both MoE and multimodality, and an emerging direction ($\nearrow$) score of 5 in AGI, totaling an overall score of 11. MoE models support the relevance of NLU by enhancing its precision and depth through their ability to handle large, diverse datasets. In multimodal AI, NLU remains a critical component for comprehending language in diverse data formats. With AGI's progress, NLU is expected to undergo significant expansion, moving towards more advanced, human-like comprehension and interpretation capabilities.

Natural Language Generation maintains relevance ($\leftrightarrow$) with a score of 3 in MoE, requires redirection ($\hookrightarrow$) with a score of 4 in multimodality, and is identified as an emerging research area ($\nearrow$) with a score of 5 in AGI, resulting in a total score of 12. MoE's scalability is crucial for enhancing NLG, while in multimodal contexts, NLG may need strategic adjustments to align effectively with other modalities. As AGI evolves, NLG is anticipated to venture into new research domains, especially in creating content that reflects human-like creativity and adaptability.

Conversational AI is marked for redirection ($\hookrightarrow$) with a score of 4 in MoE, emerging research directions ($\nearrow$) with a score of 5 in both multimodality and AGI, accumulating an overall score of 14. While MoE enhances conversational AI, it may require strategic changes to fully utilize MoE's distributed expertise. The integration of multiple modalities opens new avenues for conversational AI, expanding its scope to include various sensory data. The development of AGI is set to bring revolutionary advancements in this domain, paving the way for more autonomous, context-aware, and human-like interactions.

Creative AI scores a redirection requirement ($\hookrightarrow$) of 4 in MoE, and high scores for emerging research directions ($\nearrow$) of 5 in both multimodality and AGI, leading to a total score of 14. In the context of MoE, Creative AI may need to be realigned to capitalize on MoE's capacity for generating novel content. The combination of different modalities in creative AI presents exciting new research opportunities, enabling the creation of more intricate and diverse outputs. As AGI progresses, it is expected to significantly broaden the capabilities of creative AI, potentially surpassing existing boundaries and exploring new realms of creativity.

\subsubsection{Impact On Compliance and Ethical Considerations}
Bias Mitigation in the context of MoE, multimodality, and AGI scores a redirection requirement ($\hookrightarrow$) of 4 in both MoE and multimodality, and an emerging research direction ($\nearrow$) with a score of 5 in AGI, resulting in an overall score of 13. MoE architectures demand a new approach in bias mitigation due to the diversity of expert networks, which could otherwise amplify biases. In multimodal systems, bias mitigation requires novel strategies to address biases in various data types, including non-textual forms like images and audio. With AGI's broad cognitive capabilities, a comprehensive approach towards understanding and addressing biases across diverse domains is emerging as a critical research area.

Data Security maintains a consistent relevance ($\leftrightarrow$) with a score of 3 across MoE, multimodality, and AGI, leading to a total score of 9. The fundamental principles of data security remain crucial despite the advancements in MoE, which may necessitate tailored strategies for its distributed nature. In multimodal AI, the secure handling of diverse data types continues to be of paramount importance. The core tenets of data security are sustained even with the advancement of AGI, though the complexity and scope of security measures are likely to increase.

AI Ethics is marked for redirection ($\hookrightarrow$) with a score of 4 in both MoE and multimodality, and faces inherently unresolvable challenges ($\bigtriangleup$) with a score of 1 in AGI, accumulating a total score of 9. The decision-making processes and transparency of MoE models necessitate a reevaluation of ethical considerations. In multimodal AI, ethical concerns, particularly in the interpretation and use of multimodal data, require new approaches. The ethical challenges in AGI are expected to be complex and involve deep philosophical and societal implications that might be difficult to fully resolve.

Privacy Preservation scores a redirection need ($\hookrightarrow$) of 4 across MoE, multimodality, and AGI, leading to an overall score of 12. The distributed nature of MoE systems requires a reassessment of privacy preservation techniques to handle data processed by multiple experts. Multimodal AI systems, especially those handling sensitive data such as images and sounds, necessitate tailored privacy strategies. With the extensive data processing capabilities of AGI, advanced and potentially new approaches to privacy preservation are called for.

\subsubsection{Impact On Advanced Learning}
In the context of MoE, self-supervised learning requires redirection ($\hookrightarrow$) with a score of 4, signaling the need to adapt to the evolving architecture. Emerging research directions ($\nearrow$) with a score of 5 are identified in multimodality, suggesting the integration of various autonomous data types like text, image, and audio. For AGI, self-supervised learning remains relevant ($\leftrightarrow$) with a score of 3, contributing to the system's autonomy and adaptability, though likely to be integrated with more complex strategies. The overall impact score is 12.

Meta-learning maintains consistent relevance ($\leftrightarrow$) with a score of 3 across MoE and multimodality, aligning well with the dynamic nature of MoE and aiding quick adaptation to varying data types and tasks in multimodal contexts. In AGI, it is marked as an emerging research direction ($\nearrow$) with a score of 5, suggesting novel research in achieving human-like adaptability and learning efficiency. The total score for meta-learning is 11.

Fine tuning continues to be relevant ($\leftrightarrow$) with a score of 3 in both MoE and multimodality, being essential for adapting pre-trained models to specific tasks and tailoring multimodal models. However, in AGI, it is likely to become redundant ($\searrow$) with a score of 2, as AGI aims to develop systems that autonomously understand and learn across a broad range of domains, reducing the need for traditional fine-tuning processes. The overall impact score for fine tuning is 8.

Aligning AI with human values poses inherently unresolvable challenges ($\bigtriangleup$) in all contexts—MoE, multimodality, and AGI—with a score of 1. This reflects the complexity and diversity of tasks MoE models handle, the integration of various data types in multimodal AI, and the broad range of cognitive abilities encompassed by AGI. These factors contribute to the significant ongoing challenges in aligning AI with human values, resulting in a total score of 3.

\subsubsection{Impact On Emerging Trends}
Multimodal learning is marked as an emerging research direction ($\nearrow$) with a score of 5 in both MoE and AGI contexts, reflecting its capacity to integrate various data types such as text, images, and audio. This integration is crucial for specialized tasks in MoE and processing diverse forms of data in AGI. In the realm of multimodality, it remains a core aspect ($\leftrightarrow$) with a score of 3, being essential for ongoing multimodal AI development. The overall impact score is 13.

Interactive and Cooperative AI requires redirection ($\hookrightarrow$) in MoE with a score of 4, as MoE models adapt to include more interactive elements for broader applications. In multimodality, interaction and cooperation continue to be central ($\leftrightarrow$) with a score of 3, especially in fields like robotics and virtual assistants. AGI's evolution includes significant advancements in interactive AI, marking it as an emerging research area ($\nearrow$) with a score of 5. The total score for this trend is 12.

The development of AGI necessitates redirection ($\hookrightarrow$) in both MoE and multimodality, each with a score of 4, indicating the need for more integrated and complex systems. AGI remains at the forefront of its own field ($\leftrightarrow$) with a score of 3, with each breakthrough directly influencing its progress. The overall impact score for AGI development is 11.

AGI containment is identified as a challenge not required to be solved ($\bigtriangleup$) in both MoE and multimodality, with a score of 1, as these areas are not expected to reach the levels of autonomy and complexity associated with AGI. However, as AGI progresses, the emerging need for effective containment strategies is marked ($\nearrow$) with a score of 5, highlighting the importance of ensuring safe and controlled AI deployment. The total impact score is 7.

\section{Emergent Research Priorities in Generative AI}
\label{sec:emergent-research-priorities-agi}
As we are likely to approach the precipice of a new era marked by the advent of Q*, nudging us closer to the realization of usable AGI, the research landscape in generative AI is undergoing a crucial transformation.

\subsection{Emergent Research Priorities in MoE}
\label{subsec:emergent-research-priorities-moe}
The MoE domain is increasingly focusing on two critical areas:

\begin{itemize}
	\item \textbf{Multimodal Models in Model Architecture:} The integration of MoE and AGI is opening new pathways for research in multimodal models. These developments are enhancing the capability to process and synthesize information from multiple modalities, which is crucial for both specialized and generalized AI systems.
	\item \textbf{Multimodal Learning in Emerging Trends:} MoE is at the forefront of multimodal learning, integrating diverse data types like text, images, and audio for specialized tasks. This trend is directly impacting the enhancement of the field.
\end{itemize}

Furthermore, an analysis of funding trends and investment patterns in AI research could indicate a substantial shift towards areas like multimodal models in MoE. This trend, characterized by increased capital flow into fields involving complex data processing and autonomous systems, is shaping the direction of future research priorities. It underscores the growing interest and investment in the potential of generative AI, influencing both academic and industry-led initiatives.

\subsection{Emergent Research Priorities in Multimodality}
\label{subsec:emergent-research-priorities-multimodality}
In the realm of multimodality, several areas are identified as emerging research priorities:

\begin{itemize}
	\item \textbf{MoE in Model Architecture:} MoE models are becoming increasingly relevant for handling diverse data types in multimodal contexts.
	\item \textbf{Transfer Learning in Training Techniques:} Transfer learning is emerging as a key research direction, especially for learning between different modalities.
	\item \textbf{Conversational AI and Creative AI in Application Domains:} Both conversational AI and creative AI are expanding in multimodal contexts, encompassing visual, auditory, and other sensory data integration.
	\item \textbf{Self-Supervised Learning in Advanced Learning:} New research directions in self-supervised learning are emerging, focusing on the integration of various data types autonomously.
\end{itemize}

Additionally, the rise of generative AI, particularly in multimodal contexts, can significantly impact educational curricula and skill development. There is a growing need to update academic programs to include comprehensive AI literacy, with a focus on multimodal AI technologies. This evolution in education is aimed at preparing future professionals to effectively engage with and leverage the advancements in AI, equipping them with the necessary skills to navigate its complexities and innovations.

\subsection{Emergent Research Priorities in AGI}
\label{subsec:emergent-research-priorities-agi}
The AGI domain is witnessing a surge in research priorities across multiple areas:

\begin{itemize}
	\item \textbf{Multimodal Models in Model Architecture:} Similar to MoE, multimodal models are crucial in AGI, enabling deeper and more nuanced understanding.
	\item \textbf{Reinforcement Learning in Training Techniques:} Emerging as a key area in AGI, reinforcement learning focuses on developing autonomous systems learning from their environment.
	\item \textbf{Application Domains:} AGI is extending the boundaries of natural language understanding and generation, conversational AI, and creative AI, with a focus on human-like comprehension and creativity.
	\item \textbf{Bias Mitigation in Compliance and Ethical Considerations:} New directions in bias mitigation are focusing on a comprehensive approach to addressing biases across diverse domains in AGI.
	\item \textbf{Meta-Learning in Advanced Learning:} AGI's pursuit of human-like adaptability is leading to novel research in meta-learning.
	\item \textbf{Emerging Trends:} Multimodal learning, interactive and cooperative AI, and AGI containment strategies are becoming crucial research areas as AGI progresses.
\end{itemize}

In line with these developments in AGI, a noticeable trend in AI research funding and investment patterns is evident. There is a significant inclination towards supporting projects and studies in AGI, particularly in areas such as natural language understanding and generation, and autonomous systems. This funding trend not only mirrors the escalating interest in the capabilities of AGI but also directs the trajectory of future research, shaping both academic exploration and industry-driven projects.

\section{Practical Implications and Limitations of Generative AI Technologies}
\label{sec:practical-implications-limitations}
Generative AI technologies, encompassing MoE, multimodality, and AGI, present unique computational challenges. This section explores the processing power requirements, memory usage, and scalability concerns inherent in these advanced AI models.

\subsection{Computational Complexity and Real-world Applications of Generative AI Technologies}

\subsubsection{Computational Complexity}
Generative AI technologies, encompassing MoE, multimodality, and AGI, present unique computational challenges. This section explores the processing power requirements, memory usage, and scalability concerns inherent in these advanced AI models.
\begin{itemize}
	\item \textbf{Processing Power Requirements}: Advanced generative AI models, including MoE architectures and AGI systems, require significant processing power \cite{zhang2023one}. The demand for GPUs and TPUs is accentuated, particularly when handling complex computations and large datasets typical in multimodal AI applications.
	\item \textbf{Memory Usage in AI Modeling}: A critical challenge in training and deploying large-scale AI models, particularly in multimodal and AGI systems executed on GPUs, lies in the substantial GPU and VRAM requirements. Unlike computer RAM, VRAM often cannot be expanded easily on many platforms, posing significant constraints. Developing strategies for GPU and VRAM optimization and efficient model scaling is thus crucial for the practical deployment of these AI technologies.
	
	\item \textbf{Scalability and Efficiency in AI Deployment}: Addressing scalability challenges in generative AI, especially in MoE and AGI contexts, involves optimizing load management and parallel processing techniques. This is vital for their practical application in fields like healthcare, finance, and education.
\end{itemize}

\subsubsection{Real-world Application Examples of Generative AI Technologies}
The application of generative AI models in real-world scenarios demonstrates their transformative potential and challenges in various sectors.
\begin{itemize}
	\item \textbf{Healthcare}: In healthcare, generative AI facilitates advancements in diagnostic imaging and personalized medicine, but also raises significant concerns regarding data privacy and the potential for misuse of sensitive health information \cite{singhal2023towards}.
	
	\item \textbf{Finance}: The use of AI for fraud detection and algorithmic trading in finance underlines its efficiency and accuracy, while at the same time, it raises ethical concerns, particularly in automated decision-making processes, which may lack transparency and accountability \cite{wu2023bloomberggpt}.
	
	\item \textbf{Education}: Generative AI's role in creating personalized learning experiences offers immense benefits in terms of educational accessibility and tailored instruction. However, it poses challenges in equitable access to technology, potential biases in AI-Generated Content (AIGC), and could reduce demand for human educators. Additionally, there's a growing concern about educators who are against the use of AIGC, fearing it may undermine traditional teaching methodologies and the role of educators.
\end{itemize}

\subsection{Commercial Viability and Industry Solutions in Generative AI Technologies}
\subsubsection{Market Readiness}
Assessing the market readiness of generative AI technologies involves analyzing cost, accessibility, deployment challenges, and user adoption trends.
\begin{itemize}
	\item \textbf{Cost Analysis}: The financial aspects of deploying generative AI, including MoE, multimodality, and AGI, are crucial for market adoption.
	\item \textbf{Accessibility and Deployment}: Integration of these technologies into existing systems and the technical expertise required are key factors influencing their adoption.
	\item \textbf{User Adoption Trends}: Understanding current adoption patterns provides insights into market acceptance and the role of user trust and perceived benefits.
\end{itemize}

\subsubsection{Existing Industry Solutions}
Generative AI is reshaping various industries by offering innovative solutions and altering market dynamics.
\begin{itemize}
	\item \textbf{Sector-Wise Deployment}: The diverse applications of generative AI, from digital content creation to process streamlining, also raise questions about originality and intellectual property rights.
	\item \textbf{Impact on Market Dynamics}: The effect of AI solutions on traditional industry structures and the introduction of novel business models are significant considerations.
	\item \textbf{Challenges and Constraints}: Addressing limitations such as scalability, data management complexity, privacy concerns, and ethical implications is essential for robust governance frameworks.
\end{itemize}

\subsection{Limitations and Future Directions in Generative AI Technologies}
\subsubsection{Technical Limitations}
Identifying and addressing technical limitations in generative AI models is crucial for their advancement and reliability.
\begin{itemize}
	\item \textbf{Contextual Understanding}: Enhancing AI's ability to understand and interpret context, especially in natural language processing and image recognition, is a key area for improvement.
	\item \textbf{Handling Ambiguous Data}: Developing better algorithms for processing ambiguous or incomplete data sets is essential for decision-making accuracy and reliability.
	 \item \textbf{Navigating Human Judgment}: Despite generative AI's accuracy in interpreting policies and procedures, its impact is limited in replacing human judgment. This is especially true in legal and political contexts where decision-makers might selectively use AIGC, leading to biased outcomes. Thus, the effectiveness of generative AI in such scenarios should be realistically assessed.
\end{itemize}

\subsubsection{Future Research Directions to Enhance the Practicality of Generative AI}
Future research in generative AI should focus on addressing current limitations and expanding its practical applications.
\begin{itemize}
	\item \textbf{Improved Contextual Understanding}: Research should aim at developing models with better contextual awareness, particularly in complex natural language and image processing tasks.
	\item \textbf{Robust Handling of Ambiguous Data}: Investigating techniques for effective processing of ambiguous data is vital for advancing the decision-making capabilities of AI models.
	\item \textbf{Ethical Integration of AIGC in Legal and Political Arenas}: Future research should focus on the ethical integration of AI-generated content into legal and political decision-making processes, which involves developing frameworks that utilize AIGC in a supportive role, ensuring it enhances human judgment and contributes to transparency and fairness \cite{henderson2018ethical}. Importantly, researchers should consider the biases and limitations inherent in AI \cite{henderson2018ethical}, alongside the potential for human fallibility, ethical complexities, and possible corruption in these domains. 
\end{itemize}

\section{Impact of Generative AI on Preprints Across Disciplines}
\label{sec:academic-challenges-preprints}
\begin{figure}[t!]
	\begin{tikzpicture}
		\begin{axis}[
			xlabel={Year},
			ylabel={Number of Preprints},
			xmin=2010, xmax=2024,
			ymin=0, 
			xtick={2011,2012,2013,2014,2015,2016,2017,2018,2019,2020,2021,2022,2023},
			xticklabels={2011,2012,2013,2014,2015,2016,2017,2018,2019,2020,2021,2022,2023},
			x tick label style={rotate=45, anchor=center,xshift=-6pt,yshift=-9pt},
			ytick={0,20000,40000,60000,80000,100000,120000},
			yticklabels={0,20k,40k,60k,80k,100k,120k},
			scaled y ticks=false,
			legend pos=north west,
			ymajorgrids=true,
			grid style=dashed,
			width=0.95\columnwidth,
			height=0.8\columnwidth,
			smooth
			]
			
			\addplot[smooth, thick, color=gray, mark=oplus*] coordinates {
				(2011,49842) 
				(2012,52641)
				(2013,55108)
				(2014,56121)
				(2015,58681)
				(2016,61099)
				(2017,61796)
				(2018,64863)
				(2019,67905)
				(2020,73422)
				(2021,74588)
				(2022,74977)
				(2023,90408)
			};
		
			\addplot[smooth, thick, color=red, mark=o] coordinates {
				(2011,18616) 
				(2012,21220)
				(2013,24060)
				(2014,26051)
				(2015,28800)
				(2016,30687)
				(2017,32209)
				(2018,34236)
				(2019,36585)
				(2020,40657)
				(2021,42284)
				(2022,45184)
				(2023,57079)
			};
			
			\addplot[smooth, thick, color=brown, mark=triangle] coordinates {
				(2011,1177) 
				(2012,2162)
				(2013,2352)
				(2014,2861)
				(2015,3468)
				(2016,4338)
				(2017,5830)
				(2018,9715)
				(2019,14290)
				(2020,16155)
				(2021,11353)
				(2022,10452)
				(2023,13196)
			};

			\addplot[smooth, thick, color=cyan, mark=*] coordinates {
				(2011,770) 
				(2012,1103)
				(2013,1667)
				(2014,1001)
				(2015,1017)
				(2016,1728)
				(2017,2543)
				(2018,3894)
				(2019,4545)
				(2020,6600)
				(2021,11609)
				(2022,14606)
				(2023,23717)
			};

			\addplot[smooth, thick, color=blue, mark=x] coordinates {
				(2011,8115) 
				(2012,11283)
				(2013,13606)
				(2014,14999)
				(2015,16734)
				(2016,21429)
				(2017,27715)
				(2018,37427)
				(2019,47911)
				(2020,62134)
				(2021,71393)
				(2022,78066)
				(2023,106753)
			};

			\legend{physics, math, stat, cs.AI, cs}
		\end{axis}
	\end{tikzpicture}
	\caption{Annual preprint submissions to different categories on arXiv.org}
	\label{fig:arxiv-preprints-compare}
\end{figure}
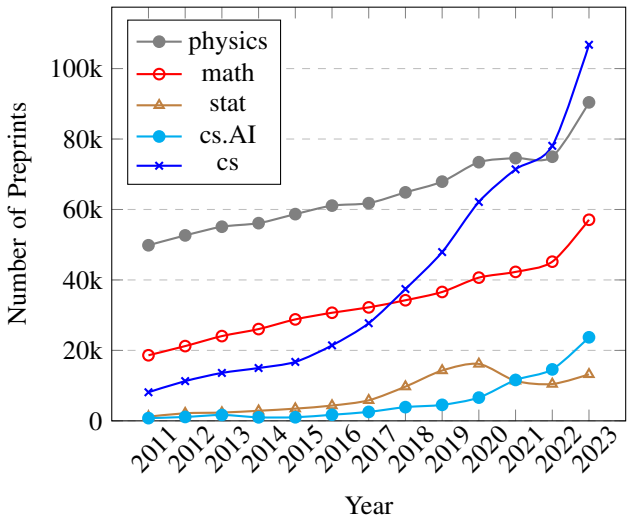

The challenges detailed in this section are not directly related to the knowledge domains within generative AI, but are fueled by the success of Generative AI, particularly the commercialization of ChatGPT. The proliferation of preprints in the field of AI (Fig. \ref{fig:arxiv-preprints-compare}), especially in the cs.AI category on platforms like arXiv, has introduced a set of academic challenges that merit careful consideration and strategic response. The rapid commercialization and adoption of tools such as ChatGPT, as evidenced by over 55,700 entries on Google Scholar mentioning \enquote{ChatGPT} within just one year of its commercialization, exemplify the accelerated pace at which the field is advancing. This rapid development is not mirrored in the traditional peer-review process, which is considerably slower. The peer-review process now appears to be overwhelmed with manuscripts that are either generated with ChatGPT (or other LLMs), or whose writing processes have been significantly accelerated by such LLMs, contributing to a bottleneck in scholarly communication \cite{bin2023use,liu2023ai}. This situation is further compounded by the fact that many journals in disciplines outside of computer science are also experiencing longer review times and higher rates of desk rejections. Additionally, the flourishing trend of manuscripts and preprints, either generated by or significantly expedited using tools like ChatGPT, extends beyond computer science into diverse academic disciplines. This trend presents a looming challenge, potentially overwhelming both the traditional peer-review process and the flourishing preprint ecosystem with a volume of work that may not always adhere to established academic standards.

The sheer volume of preprints has made the task of selecting and scrutinizing research exceedingly demanding. In the current research era, the exploration of scientific literature has become increasingly complex, as knowledge has continued to expand and disseminate exponentially, while concurrently, integrative research efforts attempting to distill these vast literature, attempt to identify and understand a smaller sets of core contributions \cite{siddaway2019systematic}. Thus, the rapid expansion of academic literature across various fields presents a significant challenge for researchers seeking to perform evidence syntheses over the increasingly vast body of available knowledge \cite{landhuis2016scientific}. Furthermore, this explosion in publication volume poses a distinct challenge for literature reviews and surveys, where the human capacity for manually selecting, understanding, and critically evaluating articles is increasingly strained, potentially leading to gaps in synthesizing comprehensive knowledge landscapes. Although reproduction of results is a theoretical possibility, practical constraints such as the lack of technical expertise, computational resources, or access to proprietary datasets hinder rigorous evaluation. This is concerning, as the inability to thoroughly assess preprint research undermines the foundation of scientific reliability and validity. Furthermore, the peer-review system, a cornerstone of academic rigour, is under the threat of being further overwhelmed \cite{bin2023use,chloros2022peer}. The potential consequences are significant, with unvetted preprints possibly perpetuating biases or errors within the scientific community and beyond. The absence of established retraction mechanisms for preprints, akin to those for published articles, exacerbates the risk of persistent dissemination of flawed research.

The academic community is at a crossroads, necessitating an urgent and thoughtful discourse on navigating this emerging \enquote{mess} — a situation that risks spiraling out of control if left unaddressed. In this context, the role of peer review becomes increasingly crucial, as it serves as a critical checkpoint for quality and validity, ensuring that the rapid production of AI research is rigorously studied for scientific accuracy and relevance. However, the current \textit{modus operandi} of traditional peer review does not appear to be sustainable, primarily due to its inability to keep pace with the exponential growth in AI-themed research and Generative-AI-accelerated research submissions, and the increasingly specialized nature of emerging AI topics \cite{bin2023use,liu2023ai}. This situation is compounded by a finite pool of qualified reviewers, leading to delays, potential biases, and a burden on the scholarly community. This reality demands an exploration of new paradigms for peer review and dissemination of research that can keep pace with swift advancements in AI. Innovative models for community-driven vetting processes, enhanced reproducibility checks, and dynamic frameworks for post-publication scrutiny and correction may be necessary. Efforts to incorporate automated tools and AI-assisted review processes could also be explored to alleviate the strain on human reviewers.

\begin{figure*}[t!]
	\centering
	\resizebox{\textwidth}{!}{
		\begin{tikzpicture}[node distance=5cm, auto]
			\tikzstyle{block} = [rectangle, draw, fill=green!20, text width=7em, text centered, rounded corners, minimum height=3em]
			\tikzstyle{line} = [draw, -latex']
			
			\node [block] (preprint) {Preprint Submission};
			\node [block, right of=preprint, node distance=5cm] (communityReview) {Community-Based Review};
			\node [block, right of=communityReview, node distance=5cm] (peerReview) {Formal Peer Review};
			\node [block, right of=peerReview, node distance=5cm] (publication) {Final Publication};
			
			\path [line] (preprint) -- (communityReview);
			\path [line] (communityReview) -- node[text width=2cm, align=center] {Initial\\Validation} (peerReview);
			\path [line] (peerReview) -- node[text width=3cm, align=center] {Rigor and\\Quality\\Assurance} (publication);
			
			\node [text width=3cm, align=center, above of=communityReview, node distance=1.2cm] (communityFeedback) {Rapid Feedback\\(Similar to Product Review Sites)};
			\node [text width=3cm, align=center, above of=peerReview, node distance=1.2cm] (formalAssessment) {In-depth\\Academic Assessment};
			
		\end{tikzpicture}
	}
	\caption{Possible Convergence Between Traditional Peer Review and the Preprint Ecosystem}
	\label{fig:peer_review_preprint_convergence}
\end{figure*}
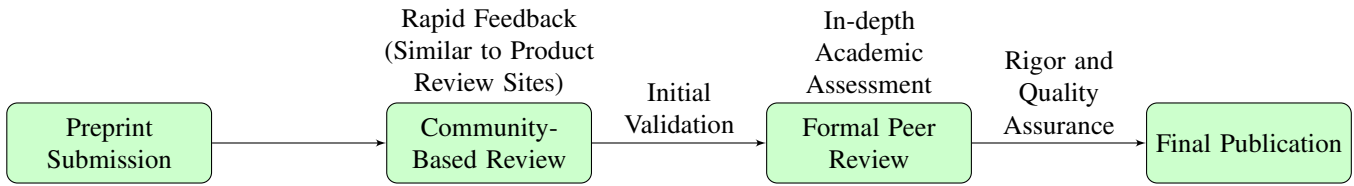

In this rapidly evolving landscape, envision a convergence between the traditional peer review system and the flourishing preprint ecosystem, which could involve creating hybrid models (Fig. \ref{fig:peer_review_preprint_convergence}),  where preprints undergo a preliminary community-based review, harnessing the collective expertise and rapid feedback of the academic community, similar to product review websites and Twitter \cite{allen2022towards}. This approach could provide an initial layer of validation, offering additional insights on issues that may be overlooked by a limited number of peer reviewers. The Editors-in-Chief (EICs) could consider the major criticisms and suggestions of an article from the community-based review, ensuring a more thorough and diverse evaluation. Subsequent, more formal peer review processes could then refine and endorse these preprints for academic rigor and quality assurance. This hybrid model would require robust technological support, possibly leveraging AI and machine learning tools to assist in initial screening and identification of suitable reviewers. The aim would be to establish a seamless continuum from rapid dissemination to validated publication, ensuring both the speed of preprints and the credibility of peer-reviewed research. A balanced approach must be struck to harness the benefits of preprints—such as rapid dissemination of findings and open access—while mitigating their drawbacks. The development of new infrastructure and norms could be instrumental in steering the academic community towards a sustainable model that upholds the integrity and trustworthiness of scientific research in the age of Generative AI.

\section{Conclusions}
\label{sec:conclusions}
This roadmap survey has embarked on an exploration of the transformative trends in generative AI research, particularly focusing on speculated advancements like Q* and the progressive strides towards AGI. Our analysis highlights a crucial paradigm shift, driven by innovations such as MoE, multimodal learning, and the pursuit of AGI. These advancements signal a future where AI systems could significantly extend their capabilities in reasoning, contextual understanding, and creative problem-solving. This study reflects on AI's dual potential to either contribute to or impede global equity and justice. The equitable distribution of AI benefits and its role in decision-making processes raise crucial questions about fairness and inclusivity. It is imperative to thoughtfully integrate AI into societal structures to enhance justice and reduce disparities. Despite these advancements, several open questions and research gaps remain. These include ensuring the ethical alignment of advanced AI systems with human values and societal norms, a challenge compounded by their increasing autonomy. The safety and robustness of AGI systems in diverse environments also remain a significant research gap. Addressing these challenges requires a multidisciplinary approach, incorporating ethical, social, and philosophical perspectives.

Our survey has highlighted key areas for future interdisciplinary research in AI, emphasizing the integration of ethical, sociological, and technical perspectives. This approach will foster collaborative research, bridging the gap between technological advancement and societal needs, ensuring that AI development is aligned with human values and global welfare. The roles of MoE, multimodal, and AGI in reshaping generative AI have been identified as significant, as their advancements can enhance model performance and versatility, and pave the way for future research in areas like ethical AI alignment and AGI. As we forge ahead, the balance between AI advancements and human creativity is not just a goal but a necessity, ensuring AI's role as a complementary force that amplifies our capacity to innovate and solve complex challenges. Our responsibility is to guide these advancements towards enriching the human experience, aligning technological progress with ethical standards and societal well-being.

\section*{Disclaimer}
The authors hereby declare no conflict of interest.

\section*{Abbreviations}
\begin{table}[H]
	\centering
	\begin{tabular}{ll}
		AGI & Artificial General Intelligence \\
		
		AI & Artificial Intelligence \\
		AIGC & AI-generated content\\
		BERT & Bidirectional Encoder Representations from Transformers \\
		CCPA & California Consumer Privacy Act \\
		DQN & Deep Q-Networks \\
		EU & European Union \\
		GAN & Generative Adversarial Network \\
		GDPR & General Data Protection Regulation \\
		GPT & Generative Pre-trained Transformers \\
		GPU & Graphics Processing Unit  \\
		LIDAR & Light Detection and Ranging \\
		LLM & Large Language Model \\
		LSTM & Long Short-Term Memory  \\
		MCTS & Monte Carlo Tree Search \\
		ML & Machine Learning \\
		MoE & Mixture of Experts \\
		NLG & Natural Language Generation \\
		NLP & Natural Language Processing \\
		NLU & Natural Language Understanding \\
		NN & Neural Network \\
		PPO & Proximal Policy Optimization \\
		RNNs & Recurrent Neural Networks \\
		VNN & Value Neural Network  \\
		VRAM & Video Random Access Memory
	\end{tabular}%
\end{table}

\bibliographystyle{IEEEtran}
\bibliography{Q-star}

\begin{thebibliography}{100}
\providecommand{\url}[1]{#1}
\csname url@samestyle\endcsname
\providecommand{\newblock}{\relax}
\providecommand{\bibinfo}[2]{#2}
\providecommand{\BIBentrySTDinterwordspacing}{\spaceskip=0pt\relax}
\providecommand{\BIBentryALTinterwordstretchfactor}{4}
\providecommand{\BIBentryALTinterwordspacing}{\spaceskip=\fontdimen2\font plus
\BIBentryALTinterwordstretchfactor\fontdimen3\font minus
  \fontdimen4\font\relax}
\providecommand{\BIBforeignlanguage}[2]{{%
\expandafter\ifx\csname l@#1\endcsname\relax
\typeout{** WARNING: IEEEtran.bst: No hyphenation pattern has been}%
\typeout{** loaded for the language `#1'. Using the pattern for}%
\typeout{** the default language instead.}%
\else
\language=\csname l@#1\endcsname
\fi
#2}}
\providecommand{\BIBdecl}{\relax}
\BIBdecl

\bibitem{turing1950computing}
A.~Turing, ``Computing machinery and intelligence,'' \emph{Mind}, vol.~59, no.
  236, p. 433, 1950.

\bibitem{mcdermott1976artificial}
D.~McDermott, ``Artificial intelligence meets natural stupidity,'' \emph{Acm
  Sigart Bulletin}, no.~57, pp. 4--9, 1976.

\bibitem{minsky1961steps}
M.~Minsky, ``Steps toward artificial intelligence,'' \emph{Proceedings of the
  IRE}, vol.~49, no.~1, pp. 8--30, 1961.

\bibitem{lecun2015deep}
Y.~LeCun, Y.~Bengio, and G.~Hinton, ``Deep learning,'' \emph{nature}, vol. 521,
  no. 7553, pp. 436--444, 2015.

\bibitem{minsky1969introduction}
M.~Minsky and S.~Papert, ``An introduction to computational geometry,''
  \emph{Cambridge tiass., HIT}, vol. 479, no. 480, p. 104, 1969.

\bibitem{rumelhart1986learning}
D.~E. Rumelhart, G.~E. Hinton, and R.~J. Williams, ``Learning representations
  by back-propagating errors,'' \emph{nature}, vol. 323, no. 6088, pp.
  533--536, 1986.

\bibitem{lee2023multimodality}
G.-G. Lee, L.~Shi, E.~Latif, Y.~Gao, A.~Bewersdorf, M.~Nyaaba, S.~Guo, Z.~Wu,
  Z.~Liu, H.~Wang \emph{et~al.}, ``Multimodality of ai for education: Towards
  artificial general intelligence,'' \emph{arXiv preprint arXiv:2312.06037},
  2023.

\bibitem{maddigan2023chat2vis}
P.~Maddigan and T.~Susnjak, ``Chat2vis: Generating data visualisations via
  natural language using chatgpt, codex and gpt-3 large language models,''
  \emph{IEEE Access}, 2023.

\bibitem{mcintosh2023culturally}
T.~R. McIntosh, T.~Liu, T.~Susnjak, P.~Watters, A.~Ng, and M.~N. Halgamuge, ``A
  culturally sensitive test to evaluate nuanced gpt hallucination,'' \emph{IEEE
  Transactions on Artificial Intelligence}, vol.~1, no.~01, pp. 1--13, 2023.

\bibitem{morris2023levels}
M.~R. Morris, J.~Sohl-dickstein, N.~Fiedel, T.~Warkentin, A.~Dafoe, A.~Faust,
  C.~Farabet, and S.~Legg, ``Levels of agi: Operationalizing progress on the
  path to agi,'' \emph{arXiv preprint arXiv:2311.02462}, 2023.

\bibitem{schuett2023towards}
J.~Schuett, N.~Dreksler, M.~Anderljung, D.~McCaffary, L.~Heim, E.~Bluemke, and
  B.~Garfinkel, ``Towards best practices in agi safety and governance: A survey
  of expert opinion,'' \emph{arXiv preprint arXiv:2305.07153}, 2023.

\bibitem{shuai2017multidimensional}
X.~Shuai, J.~Rollins, I.~Moulinier, T.~Custis, M.~Edmunds, and F.~Schilder, ``A
  multidimensional investigation of the effects of publication retraction on
  scholarly impact,'' \emph{Journal of the Association for Information Science
  and Technology}, vol.~68, no.~9, pp. 2225--2236, 2017.

\bibitem{vaswani2017attention}
A.~Vaswani, N.~Shazeer, N.~Parmar, J.~Uszkoreit, L.~Jones, A.~N. Gomez,
  {\L}.~Kaiser, and I.~Polosukhin, ``Attention is all you need,''
  \emph{Advances in neural information processing systems}, vol.~30, 2017.

\bibitem{radford2018improving}
A.~Radford, K.~Narasimhan, T.~Salimans, I.~Sutskever \emph{et~al.}, ``Improving
  language understanding by generative pre-training,'' 2018.

\bibitem{huang2022overview}
C.~Huang, Z.~Zhang, B.~Mao, and X.~Yao, ``An overview of artificial
  intelligence ethics,'' \emph{IEEE Transactions on Artificial Intelligence},
  2022.

\bibitem{besanccon2021open}
L.~Besan{\c{c}}on, N.~Peiffer-Smadja, C.~Segalas, H.~Jiang, P.~Masuzzo,
  C.~Smout, E.~Billy, M.~Deforet, and C.~Leyrat, ``Open science saves lives:
  lessons from the covid-19 pandemic,'' \emph{BMC Medical Research
  Methodology}, vol.~21, no.~1, pp. 1--18, 2021.

\bibitem{triggle2022requiem}
C.~R. Triggle, R.~MacDonald, D.~J. Triggle, and D.~Grierson, ``Requiem for
  impact factors and high publication charges,'' \emph{Accountability in
  Research}, vol.~29, no.~3, pp. 133--164, 2022.

\bibitem{mcintosh2021ransomware}
T.~McIntosh, A.~Kayes, Y.-P.~P. Chen, A.~Ng, and P.~Watters, ``Ransomware
  mitigation in the modern era: A comprehensive review, research challenges,
  and future directions,'' \emph{ACM Computing Surveys (CSUR)}, vol.~54, no.~9,
  pp. 1--36, 2021.

\bibitem{mcintosh2023harnessing}
T.~McIntosh, T.~Liu, T.~Susnjak, H.~Alavizadeh, A.~Ng, R.~Nowrozy, and
  P.~Watters, ``Harnessing gpt-4 for generation of cybersecurity grc policies:
  A focus on ransomware attack mitigation,'' \emph{Computers \& Security}, vol.
  134, p. 103424, 2023.

\bibitem{bao2022vlmo}
H.~Bao, W.~Wang, L.~Dong, Q.~Liu, O.~K. Mohammed, K.~Aggarwal, S.~Som, S.~Piao,
  and F.~Wei, ``Vlmo: Unified vision-language pre-training with
  mixture-of-modality-experts,'' \emph{Advances in Neural Information
  Processing Systems}, vol.~35, pp. 32\,897--32\,912, 2022.

\bibitem{du2022glam}
N.~Du, Y.~Huang, A.~M. Dai, S.~Tong, D.~Lepikhin, Y.~Xu, M.~Krikun, Y.~Zhou,
  A.~W. Yu, O.~Firat \emph{et~al.}, ``Glam: Efficient scaling of language
  models with mixture-of-experts,'' in \emph{International Conference on
  Machine Learning}.\hskip 1em plus 0.5em minus 0.4em\relax PMLR, 2022, pp.
  5547--5569.

\bibitem{masoudnia2014mixture}
S.~Masoudnia and R.~Ebrahimpour, ``Mixture of experts: a literature survey,''
  \emph{Artificial Intelligence Review}, vol.~42, pp. 275--293, 2014.

\bibitem{riquelme2021scaling}
C.~Riquelme, J.~Puigcerver, B.~Mustafa, M.~Neumann, R.~Jenatton,
  A.~Susano~Pinto, D.~Keysers, and N.~Houlsby, ``Scaling vision with sparse
  mixture of experts,'' \emph{Advances in Neural Information Processing
  Systems}, vol.~34, pp. 8583--8595, 2021.

\bibitem{yuksel2012twenty}
S.~E. Yuksel, J.~N. Wilson, and P.~D. Gader, ``Twenty years of mixture of
  experts,'' \emph{IEEE transactions on neural networks and learning systems},
  vol.~23, no.~8, pp. 1177--1193, 2012.

\bibitem{zhang2019learning}
L.~Zhang, S.~Huang, W.~Liu, and D.~Tao, ``Learning a mixture of
  granularity-specific experts for fine-grained categorization,'' in
  \emph{Proceedings of the IEEE/CVF International Conference on Computer
  Vision}, 2019, pp. 8331--8340.

\bibitem{martin2022multimodality}
D.~Martin, S.~Malpica, D.~Gutierrez, B.~Masia, and A.~Serrano, ``Multimodality
  in vr: A survey,'' \emph{ACM Computing Surveys (CSUR)}, vol.~54, no. 10s, pp.
  1--36, 2022.

\bibitem{sun2023generative}
Q.~Sun, Q.~Yu, Y.~Cui, F.~Zhang, X.~Zhang, Y.~Wang, H.~Gao, J.~Liu, T.~Huang,
  and X.~Wang, ``Generative pretraining in multimodality,'' \emph{arXiv
  preprint arXiv:2307.05222}, 2023.

\bibitem{wei2022mvp}
L.~Wei, L.~Xie, W.~Zhou, H.~Li, and Q.~Tian, ``Mvp: Multimodality-guided visual
  pre-training,'' in \emph{European Conference on Computer Vision}.\hskip 1em
  plus 0.5em minus 0.4em\relax Springer, 2022, pp. 337--353.

\bibitem{wu2023menet}
J.~Wu, W.~Zhou, X.~Qian, J.~Lei, L.~Yu, and T.~Luo, ``Menet: Lightweight
  multimodality enhancement network for detecting salient objects in
  rgb-thermal images,'' \emph{Neurocomputing}, vol. 527, pp. 119--129, 2023.

\bibitem{ye2023mplug}
Q.~Ye, H.~Xu, G.~Xu, J.~Ye, M.~Yan, Y.~Zhou, J.~Wang, A.~Hu, P.~Shi, Y.~Shi
  \emph{et~al.}, ``mplug-owl: Modularization empowers large language models
  with multimodality,'' \emph{arXiv preprint arXiv:2304.14178}, 2023.

\bibitem{lagrandeur2021safe}
K.~LaGrandeur, ``How safe is our reliance on ai, and should we regulate it?''
  \emph{AI and Ethics}, vol.~1, pp. 93--99, 2021.

\bibitem{mclean2023risks}
S.~McLean, G.~J. Read, J.~Thompson, C.~Baber, N.~A. Stanton, and P.~M. Salmon,
  ``The risks associated with artificial general intelligence: A systematic
  review,'' \emph{Journal of Experimental \& Theoretical Artificial
  Intelligence}, vol.~35, no.~5, pp. 649--663, 2023.

\bibitem{dwivedi2021artificial}
Y.~K. Dwivedi, L.~Hughes, E.~Ismagilova, G.~Aarts, C.~Coombs, T.~Crick,
  Y.~Duan, R.~Dwivedi, J.~Edwards, A.~Eirug, V.~Galanos, P.~V. Ilavarasan,
  M.~Janssen, P.~Jones, A.~K. Kar, H.~Kizgin, B.~Kronemann, B.~Lal, B.~Lucini,
  R.~Medaglia, K.~{Le Meunier-FitzHugh}, L.~C. {Le Meunier-FitzHugh}, S.~Misra,
  E.~Mogaji, S.~K. Sharma, J.~B. Singh, V.~Raghavan, R.~Raman, N.~P. Rana,
  S.~Samothrakis, J.~Spencer, K.~Tamilmani, A.~Tubadji, P.~Walton, and M.~D.
  Williams, ``Artificial intelligence (ai): Multidisciplinary perspectives on
  emerging challenges, opportunities, and agenda for research, practice and
  policy,'' \emph{International Journal of Information Management}, vol.~57, p.
  101994, 2021.

\bibitem{gabriel2020artificial}
I.~Gabriel, ``Artificial intelligence, values, and alignment,'' \emph{Minds and
  Machines}, vol.~30, pp. 411--437, 2020.

\bibitem{shaban2022applied}
A.~Shaban-Nejad, M.~Michalowski, S.~Bianco, J.~S. Brownstein, D.~L. Buckeridge,
  and R.~L. Davis, ``Applied artificial intelligence in healthcare: Listening
  to the winds of change in a post-covid-19 world,'' pp. 1969--1971, 2022.

\bibitem{ji2023survey}
Z.~Ji, N.~Lee, R.~Frieske, T.~Yu, D.~Su, Y.~Xu, E.~Ishii, Y.~J. Bang,
  A.~Madotto, and P.~Fung, ``Survey of hallucination in natural language
  generation,'' \emph{ACM Computing Surveys}, vol.~55, no.~12, pp. 1--38, 2023.

\bibitem{min2023recent}
B.~Min, H.~Ross, E.~Sulem, A.~P.~B. Veyseh, T.~H. Nguyen, O.~Sainz, E.~Agirre,
  I.~Heintz, and D.~Roth, ``Recent advances in natural language processing via
  large pre-trained language models: A survey,'' \emph{ACM Computing Surveys},
  vol.~56, no.~2, pp. 1--40, 2023.

\bibitem{li2023halueval}
J.~Li, X.~Cheng, W.~X. Zhao, J.-Y. Nie, and J.-R. Wen, ``Halueval: A
  large-scale hallucination evaluation benchmark for large language models,''
  in \emph{Proceedings of the 2023 Conference on Empirical Methods in Natural
  Language Processing}, 2023, pp. 6449--6464.

\bibitem{weidinger2021ethical}
L.~Weidinger, J.~Mellor, M.~Rauh, C.~Griffin, J.~Uesato, P.-S. Huang, M.~Cheng,
  M.~Glaese, B.~Balle, A.~Kasirzadeh \emph{et~al.}, ``Ethical and social risks
  of harm from language models,'' \emph{arXiv preprint arXiv:2112.04359}, 2021.

\bibitem{zhiheng2023safety}
X.~Zhiheng, Z.~Rui, and G.~Tao, ``Safety and ethical concerns of large language
  models,'' in \emph{Proceedings of the 22nd Chinese National Conference on
  Computational Linguistics (Volume 4: Tutorial Abstracts)}, 2023, pp. 9--16.

\bibitem{brown1992class}
P.~F. Brown, V.~J. Della~Pietra, P.~V. Desouza, J.~C. Lai, and R.~L. Mercer,
  ``Class-based n-gram models of natural language,'' \emph{Computational
  linguistics}, vol.~18, no.~4, pp. 467--480, 1992.

\bibitem{katz1987estimation}
S.~Katz, ``Estimation of probabilities from sparse data for the language model
  component of a speech recognizer,'' \emph{IEEE transactions on acoustics,
  speech, and signal processing}, vol.~35, no.~3, pp. 400--401, 1987.

\bibitem{kneser1995improved}
R.~Kneser and H.~Ney, ``Improved backing-off for m-gram language modeling,'' in
  \emph{1995 international conference on acoustics, speech, and signal
  processing}, vol.~1.\hskip 1em plus 0.5em minus 0.4em\relax IEEE, 1995, pp.
  181--184.

\bibitem{kuhn1990cache}
R.~Kuhn and R.~De~Mori, ``A cache-based natural language model for speech
  recognition,'' \emph{IEEE transactions on pattern analysis and machine
  intelligence}, vol.~12, no.~6, pp. 570--583, 1990.

\bibitem{ney1994structuring}
H.~Ney, U.~Essen, and R.~Kneser, ``On structuring probabilistic dependences in
  stochastic language modelling,'' \emph{Computer Speech \& Language}, vol.~8,
  no.~1, pp. 1--38, 1994.

\bibitem{hochreiter1997long}
S.~Hochreiter and J.~Schmidhuber, ``Long short-term memory,'' \emph{Neural
  computation}, vol.~9, no.~8, pp. 1735--1780, 1997.

\bibitem{nammous2019natural}
M.~K. Nammous and K.~Saeed, ``Natural language processing: speaker, language,
  and gender identification with lstm,'' \emph{Advanced Computing and Systems
  for Security: Volume Eight}, pp. 143--156, 2019.

\bibitem{wei2017research}
D.~Wei, B.~Wang, G.~Lin, D.~Liu, Z.~Dong, H.~Liu, and Y.~Liu, ``Research on
  unstructured text data mining and fault classification based on rnn-lstm with
  malfunction inspection report,'' \emph{Energies}, vol.~10, no.~3, p. 406,
  2017.

\bibitem{yao2018improved}
L.~Yao and Y.~Guan, ``An improved lstm structure for natural language
  processing,'' in \emph{2018 IEEE International Conference of Safety Produce
  Informatization (IICSPI)}.\hskip 1em plus 0.5em minus 0.4em\relax IEEE, 2018,
  pp. 565--569.

\bibitem{ouyang2022training}
L.~Ouyang, J.~Wu, X.~Jiang, D.~Almeida, C.~Wainwright, P.~Mishkin, C.~Zhang,
  S.~Agarwal, K.~Slama, A.~Ray \emph{et~al.}, ``Training language models to
  follow instructions with human feedback,'' \emph{Advances in Neural
  Information Processing Systems}, vol.~35, pp. 27\,730--27\,744, 2022.

\bibitem{susnjak2023beyond}
T.~Susnjak, ``Beyond predictive learning analytics modelling and onto
  explainable artificial intelligence with prescriptive analytics and
  chatgpt,'' \emph{International Journal of Artificial Intelligence in
  Education}, pp. 1--31, 2023.

\bibitem{susnjak2023towards}
T.~Susnjak, E.~Griffin, M.~McCutcheon, and K.~Potter, ``Towards clinical
  prediction with transparency: An explainable ai approach to survival
  modelling in residential aged care,'' \emph{arXiv preprint arXiv:2312.00271},
  2023.

\bibitem{yang2023large}
R.~Yang, T.~F. Tan, W.~Lu, A.~J. Thirunavukarasu, D.~S.~W. Ting, and N.~Liu,
  ``Large language models in health care: Development, applications, and
  challenges,'' \emph{Health Care Science}, vol.~2, no.~4, pp. 255--263, 2023.

\bibitem{baidoo2023education}
D.~Baidoo-Anu and L.~O. Ansah, ``Education in the era of generative artificial
  intelligence (ai): Understanding the potential benefits of chatgpt in
  promoting teaching and learning,'' \emph{Journal of AI}, vol.~7, no.~1, pp.
  52--62, 2023.

\bibitem{susnjak2022chatgpt}
T.~Susnjak, ``Chatgpt: The end of online exam integrity?'' \emph{arXiv preprint
  arXiv:2212.09292}, 2022.

\bibitem{tlili2023if}
A.~Tlili, B.~Shehata, M.~A. Adarkwah, A.~Bozkurt, D.~T. Hickey, R.~Huang, and
  B.~Agyemang, ``What if the devil is my guardian angel: Chatgpt as a case
  study of using chatbots in education,'' \emph{Smart Learning Environments},
  vol.~10, no.~1, p.~15, 2023.

\bibitem{alafnan2023chatgpt}
M.~A. AlAfnan, S.~Dishari, M.~Jovic, and K.~Lomidze, ``Chatgpt as an
  educational tool: Opportunities, challenges, and recommendations for
  communication, business writing, and composition courses,'' \emph{Journal of
  Artificial Intelligence and Technology}, vol.~3, no.~2, pp. 60--68, 2023.

\bibitem{george2023review}
A.~S. George and A.~H. George, ``A review of chatgpt ai's impact on several
  business sectors,'' \emph{Partners Universal International Innovation
  Journal}, vol.~1, no.~1, pp. 9--23, 2023.

\bibitem{hadfield2023regulatory}
G.~K. Hadfield and J.~Clark, ``Regulatory markets: The future of ai
  governance,'' \emph{arXiv preprint arXiv:2304.04914}, 2023.

\bibitem{bakker2022fine}
M.~Bakker, M.~Chadwick, H.~Sheahan, M.~Tessler, L.~Campbell-Gillingham,
  J.~Balaguer, N.~McAleese, A.~Glaese, J.~Aslanides, M.~Botvinick
  \emph{et~al.}, ``Fine-tuning language models to find agreement among humans
  with diverse preferences,'' \emph{Advances in Neural Information Processing
  Systems}, vol.~35, pp. 38\,176--38\,189, 2022.

\bibitem{hu2023llm}
Z.~Hu, Y.~Lan, L.~Wang, W.~Xu, E.-P. Lim, R.~K.-W. Lee, L.~Bing, and S.~Poria,
  ``Llm-adapters: An adapter family for parameter-efficient fine-tuning of
  large language models,'' \emph{arXiv preprint arXiv:2304.01933}, 2023.

\bibitem{liu2022few}
H.~Liu, D.~Tam, M.~Muqeeth, J.~Mohta, T.~Huang, M.~Bansal, and C.~A. Raffel,
  ``Few-shot parameter-efficient fine-tuning is better and cheaper than
  in-context learning,'' \emph{Advances in Neural Information Processing
  Systems}, vol.~35, pp. 1950--1965, 2022.

\bibitem{zheng2023learn}
H.~Zheng, L.~Shen, A.~Tang, Y.~Luo, H.~Hu, B.~Du, and D.~Tao, ``Learn from
  model beyond fine-tuning: A survey,'' \emph{arXiv preprint arXiv:2310.08184},
  2023.

\bibitem{manakul2023selfcheckgpt}
P.~Manakul, A.~Liusie, and M.~J. Gales, ``Selfcheckgpt: Zero-resource black-box
  hallucination detection for generative large language models,'' \emph{arXiv
  preprint arXiv:2303.08896}, 2023.

\bibitem{martino2023knowledge}
A.~Martino, M.~Iannelli, and C.~Truong, ``Knowledge injection to counter large
  language model (llm) hallucination,'' in \emph{European Semantic Web
  Conference}.\hskip 1em plus 0.5em minus 0.4em\relax Springer, 2023, pp.
  182--185.

\bibitem{yao2023llm}
J.-Y. Yao, K.-P. Ning, Z.-H. Liu, M.-N. Ning, and L.~Yuan, ``Llm lies:
  Hallucinations are not bugs, but features as adversarial examples,''
  \emph{arXiv preprint arXiv:2310.01469}, 2023.

\bibitem{zhang2023siren}
Y.~Zhang, Y.~Li, L.~Cui, D.~Cai, L.~Liu, T.~Fu, X.~Huang, E.~Zhao, Y.~Zhang,
  Y.~Chen \emph{et~al.}, ``Siren's song in the ai ocean: A survey on
  hallucination in large language models,'' \emph{arXiv preprint
  arXiv:2309.01219}, 2023.

\bibitem{ji2023beavertails}
J.~Ji, M.~Liu, J.~Dai, X.~Pan, C.~Zhang, C.~Bian, R.~Sun, Y.~Wang, and Y.~Yang,
  ``Beavertails: Towards improved safety alignment of llm via a
  human-preference dataset,'' \emph{arXiv preprint arXiv:2307.04657}, 2023.

\bibitem{liu2023trustworthy}
Y.~Liu, Y.~Yao, J.-F. Ton, X.~Zhang, R.~G.~H. Cheng, Y.~Klochkov, M.~F. Taufiq,
  and H.~Li, ``Trustworthy llms: a survey and guideline for evaluating large
  language models' alignment,'' \emph{arXiv preprint arXiv:2308.05374}, 2023.

\bibitem{wang2023aligning}
Y.~Wang, W.~Zhong, L.~Li, F.~Mi, X.~Zeng, W.~Huang, L.~Shang, X.~Jiang, and
  Q.~Liu, ``Aligning large language models with human: A survey,'' \emph{arXiv
  preprint arXiv:2307.12966}, 2023.

\bibitem{sun2023principle}
Z.~Sun, Y.~Shen, Q.~Zhou, H.~Zhang, Z.~Chen, D.~Cox, Y.~Yang, and C.~Gan,
  ``Principle-driven self-alignment of language models from scratch with
  minimal human supervision,'' \emph{arXiv preprint arXiv:2305.03047}, 2023.

\bibitem{wolf2023fundamental}
Y.~Wolf, N.~Wies, Y.~Levine, and A.~Shashua, ``Fundamental limitations of
  alignment in large language models,'' \emph{arXiv preprint arXiv:2304.11082},
  2023.

\bibitem{dang2022prompt}
H.~Dang, L.~Mecke, F.~Lehmann, S.~Goller, and D.~Buschek, ``How to prompt?
  opportunities and challenges of zero-and few-shot learning for human-ai
  interaction in creative applications of generative models,'' \emph{arXiv
  preprint arXiv:2209.01390}, 2022.

\bibitem{ma2021template}
R.~Ma, X.~Zhou, T.~Gui, Y.~Tan, L.~Li, Q.~Zhang, and X.~Huang, ``Template-free
  prompt tuning for few-shot ner,'' \emph{arXiv preprint arXiv:2109.13532},
  2021.

\bibitem{qin2021lfpt5}
C.~Qin and S.~Joty, ``Lfpt5: A unified framework for lifelong few-shot language
  learning based on prompt tuning of t5,'' \emph{arXiv preprint
  arXiv:2110.07298}, 2021.

\bibitem{wang2022trustworthy}
S.~Wang, L.~Tang, A.~Majety, J.~F. Rousseau, G.~Shih, Y.~Ding, and Y.~Peng,
  ``Trustworthy assertion classification through prompting,'' \emph{Journal of
  biomedical informatics}, vol. 132, p. 104139, 2022.

\bibitem{fan2023grammargpt}
Y.~Fan, F.~Jiang, P.~Li, and H.~Li, ``Grammargpt: Exploring open-source llms
  for native chinese grammatical error correction with supervised
  fine-tuning,'' in \emph{CCF International Conference on Natural Language
  Processing and Chinese Computing}.\hskip 1em plus 0.5em minus 0.4em\relax
  Springer, 2023, pp. 69--80.

\bibitem{liga2023fine}
D.~Liga and L.~Robaldo, ``Fine-tuning gpt-3 for legal rule classification,''
  \emph{Computer Law \& Security Review}, vol.~51, p. 105864, 2023.

\bibitem{liu2023improving}
Y.~Liu, A.~Singh, C.~D. Freeman, J.~D. Co-Reyes, and P.~J. Liu, ``Improving
  large language model fine-tuning for solving math problems,'' \emph{arXiv
  preprint arXiv:2310.10047}, 2023.

\bibitem{talat2022you}
Z.~Talat, A.~N{\'e}v{\'e}ol, S.~Biderman, M.~Clinciu, M.~Dey, S.~Longpre,
  S.~Luccioni, M.~Masoud, M.~Mitchell, D.~Radev \emph{et~al.}, ``You reap what
  you sow: On the challenges of bias evaluation under multilingual settings,''
  in \emph{Proceedings of BigScience Episode\# 5--Workshop on Challenges \&
  Perspectives in Creating Large Language Models}, 2022, pp. 26--41.

\bibitem{liu2023hessian}
Y.~Liu, S.~Yu, and T.~Lin, ``Hessian regularization of deep neural networks: A
  novel approach based on stochastic estimators of hessian trace,''
  \emph{Neurocomputing}, vol. 536, pp. 13--20, 2023.

\bibitem{lu2021confidence}
Y.~Lu, Y.~Bo, and W.~He, ``Confidence adaptive regularization for deep learning
  with noisy labels,'' \emph{arXiv preprint arXiv:2108.08212}, 2021.

\bibitem{pereyra2017regularizing}
G.~Pereyra, G.~Tucker, J.~Chorowski, {\L}.~Kaiser, and G.~Hinton,
  ``Regularizing neural networks by penalizing confident output
  distributions,'' \emph{arXiv preprint arXiv:1701.06548}, 2017.

\bibitem{chen2022redeeming}
E.~Chen, Z.-W. Hong, J.~Pajarinen, and P.~Agrawal, ``Redeeming intrinsic
  rewards via constrained optimization,'' \emph{Advances in Neural Information
  Processing Systems}, vol.~35, pp. 4996--5008, 2022.

\bibitem{jiang2022stable}
Y.~Jiang, Z.~Li, M.~Tan, S.~Wei, G.~Zhang, Z.~Guan, and B.~Han, ``A stable
  block adjustment method without ground control points using bound constrained
  optimization,'' \emph{International Journal of Remote Sensing}, vol.~43,
  no.~12, pp. 4708--4722, 2022.

\bibitem{kachuee2022constrained}
M.~Kachuee and S.~Lee, ``Constrained policy optimization for controlled
  self-learning in conversational ai systems,'' \emph{arXiv preprint
  arXiv:2209.08429}, 2022.

\bibitem{song2023surrogate}
Z.~Song, H.~Wang, and Y.~Jin, ``A surrogate-assisted evolutionary framework
  with regions of interests-based data selection for expensive constrained
  optimization,'' \emph{IEEE Transactions on Systems, Man, and Cybernetics:
  Systems}, 2023.

\bibitem{yu2022structure}
J.~Yu, T.~Xu, Y.~Rong, J.~Huang, and R.~He, ``Structure-aware conditional
  variational auto-encoder for constrained molecule optimization,''
  \emph{Pattern Recognition}, vol. 126, p. 108581, 2022.

\bibitem{butlin2021ai}
P.~Butlin, ``Ai alignment and human reward,'' in \emph{Proceedings of the 2021
  AAAI/ACM Conference on AI, Ethics, and Society}, 2021, pp. 437--445.

\bibitem{faal2023reward}
F.~Faal, K.~Schmitt, and J.~Y. Yu, ``Reward modeling for mitigating toxicity in
  transformer-based language models,'' \emph{Applied Intelligence}, vol.~53,
  no.~7, pp. 8421--8435, 2023.

\bibitem{leike2018scalable}
J.~Leike, D.~Krueger, T.~Everitt, M.~Martic, V.~Maini, and S.~Legg, ``Scalable
  agent alignment via reward modeling: a research direction,'' \emph{arXiv
  preprint arXiv:1811.07871}, 2018.

\bibitem{li2023tool}
L.~Li, Y.~Chai, S.~Wang, Y.~Sun, H.~Tian, N.~Zhang, and H.~Wu, ``Tool-augmented
  reward modeling,'' \emph{arXiv preprint arXiv:2310.01045}, 2023.

\bibitem{barreto2023generative}
F.~Barreto, L.~Moharkar, M.~Shirodkar, V.~Sarode, S.~Gonsalves, and A.~Johns,
  ``Generative artificial intelligence: Opportunities and challenges of large
  language models,'' in \emph{International Conference on Intelligent Computing
  and Networking}.\hskip 1em plus 0.5em minus 0.4em\relax Springer, 2023, pp.
  545--553.

\bibitem{chen2023octavius}
Z.~Chen, Z.~Wang, Z.~Wang, H.~Liu, Z.~Yin, S.~Liu, L.~Sheng, W.~Ouyang,
  Y.~Qiao, and J.~Shao, ``Octavius: Mitigating task interference in mllms via
  moe,'' \emph{arXiv preprint arXiv:2311.02684}, 2023.

\bibitem{dun2023sweeping}
C.~Dun, M.~D. C.~H. Garcia, G.~Zheng, A.~H. Awadallah, A.~Kyrillidis, and
  R.~Sim, ``Sweeping heterogeneity with smart mops: Mixture of prompts for llm
  task adaptation,'' \emph{arXiv preprint arXiv:2310.02842}, 2023.

\bibitem{naveed2023comprehensive}
H.~Naveed, A.~U. Khan, S.~Qiu, M.~Saqib, S.~Anwar, M.~Usman, N.~Barnes, and
  A.~Mian, ``A comprehensive overview of large language models,'' \emph{arXiv
  preprint arXiv:2307.06435}, 2023.

\bibitem{xue2023repeat}
F.~Xue, Y.~Fu, W.~Zhou, Z.~Zheng, and Y.~You, ``To repeat or not to repeat:
  Insights from scaling llm under token-crisis,'' \emph{arXiv preprint
  arXiv:2305.13230}, 2023.

\bibitem{nowaz2023patch}
M.~Nowaz Rabbani~Chowdhury, S.~Zhang, M.~Wang, S.~Liu, and P.-Y. Chen,
  ``Patch-level routing in mixture-of-experts is provably sample-efficient for
  convolutional neural networks,'' \emph{arXiv e-prints}, pp. arXiv--2306,
  2023.

\bibitem{peng2023sparse}
J.~Peng, K.~Zhou, R.~Zhou, T.~Hartvigsen, Y.~Zhang, Z.~Wang, and T.~Chen,
  ``Sparse moe as a new treatment: Addressing forgetting, fitting, learning
  issues in multi-modal multi-task learning,'' in \emph{Conference on Parsimony
  and Learning (Recent Spotlight Track)}, 2023.

\bibitem{santos2023memory}
C.~N.~d. Santos, J.~Lee-Thorp, I.~Noble, C.-C. Chang, and D.~Uthus, ``Memory
  augmented language models through mixture of word experts,'' \emph{arXiv
  preprint arXiv:2311.10768}, 2023.

\bibitem{wang2023language}
W.~Wang, G.~Ma, Y.~Li, and B.~Du, ``Language-routing mixture of experts for
  multilingual and code-switching speech recognition,'' \emph{arXiv preprint
  arXiv:2307.05956}, 2023.

\bibitem{zhao2023sparse}
X.~Zhao, X.~Chen, Y.~Cheng, and T.~Chen, ``Sparse moe with language guided
  routing for multilingual machine translation,'' in \emph{Conference on
  Parsimony and Learning (Recent Spotlight Track)}, 2023.

\bibitem{huang2023multi}
W.~Huang, H.~Zhang, P.~Peng, and H.~Wang, ``Multi-gate mixture-of-expert
  combined with synthetic minority over-sampling technique for multimode
  imbalanced fault diagnosis,'' in \emph{2023 26th International Conference on
  Computer Supported Cooperative Work in Design (CSCWD)}.\hskip 1em plus 0.5em
  minus 0.4em\relax IEEE, 2023, pp. 456--461.

\bibitem{liu2023diversifying}
B.~Liu, L.~Ding, L.~Shen, K.~Peng, Y.~Cao, D.~Cheng, and D.~Tao, ``Diversifying
  the mixture-of-experts representation for language models with orthogonal
  optimizer,'' \emph{arXiv preprint arXiv:2310.09762}, 2023.

\bibitem{wang2023prophet}
W.~Wang, Z.~Lai, S.~Li, W.~Liu, K.~Ge, Y.~Liu, A.~Shen, and D.~Li, ``Prophet:
  Fine-grained load balancing for parallel training of large-scale moe
  models,'' in \emph{2023 IEEE International Conference on Cluster Computing
  (CLUSTER)}.\hskip 1em plus 0.5em minus 0.4em\relax IEEE, 2023, pp. 82--94.

\bibitem{yao2023enhancing}
X.~Yao, S.~Liang, S.~Han, and H.~Huang, ``Enhancing molecular property
  prediction via mixture of collaborative experts,'' \emph{arXiv preprint
  arXiv:2312.03292}, 2023.

\bibitem{xiao2021adversarial}
Z.~Xiao, Y.~Jiang, G.~Tang, L.~Liu, S.~Xu, Y.~Xiao, and W.~Yan, ``Adversarial
  mixture of experts with category hierarchy soft constraint,'' in \emph{2021
  IEEE 37th International Conference on Data Engineering (ICDE)}.\hskip 1em
  plus 0.5em minus 0.4em\relax IEEE, 2021, pp. 2453--2463.

\bibitem{agbese2023implementing}
M.~Agbese, R.~Mohanani, A.~Khan, and P.~Abrahamsson, ``Implementing ai ethics:
  Making sense of the ethical requirements,'' in \emph{Proceedings of the 27th
  International Conference on Evaluation and Assessment in Software
  Engineering}, 2023, pp. 62--71.

\bibitem{chen2022towards}
Z.~Chen, Y.~Deng, Y.~Wu, Q.~Gu, and Y.~Li, ``Towards understanding the
  mixture-of-experts layer in deep learning,'' \emph{Advances in neural
  information processing systems}, vol.~35, pp. 23\,049--23\,062, 2022.

\bibitem{zhou2022mixture}
Y.~Zhou, T.~Lei, H.~Liu, N.~Du, Y.~Huang, V.~Zhao, A.~M. Dai, Q.~V. Le,
  J.~Laudon \emph{et~al.}, ``Mixture-of-experts with expert choice routing,''
  \emph{Advances in Neural Information Processing Systems}, vol.~35, pp.
  7103--7114, 2022.

\bibitem{guha2023ai}
N.~Guha, C.~Lawrence, L.~A. Gailmard, K.~Rodolfa, F.~Surani, R.~Bommasani,
  I.~Raji, M.-F. Cu{\'e}llar, C.~Honigsberg, P.~Liang \emph{et~al.}, ``Ai
  regulation has its own alignment problem: The technical and institutional
  feasibility of disclosure, registration, licensing, and auditing,''
  \emph{George Washington Law Review, Forthcoming}, 2023.

\bibitem{gemini2023gemini}
\BIBentryALTinterwordspacing
{Gemini Team, Google}, ``Gemini: A family of highly capable multimodal
  models,'' 2023, accessed: 17 December 2023. [Online]. Available:
  \url{https://storage.googleapis.com/deepmind-media/gemini/gemini\_1\_report.pdf}
\BIBentrySTDinterwordspacing

\bibitem{acosta2022multimodal}
J.~N. Acosta, G.~J. Falcone, P.~Rajpurkar, and E.~J. Topol, ``Multimodal
  biomedical ai,'' \emph{Nature Medicine}, vol.~28, no.~9, pp. 1773--1784,
  2022.

\bibitem{qi2023limitation}
S.~Qi, Z.~Cao, J.~Rao, L.~Wang, J.~Xiao, and X.~Wang, ``What is the limitation
  of multimodal llms? a deeper look into multimodal llms through prompt
  probing,'' \emph{Information Processing \& Management}, vol.~60, no.~6, p.
  103510, 2023.

\bibitem{xu2020applications}
B.~Xu, D.~Kocyigit, R.~Grimm, B.~P. Griffin, and F.~Cheng, ``Applications of
  artificial intelligence in multimodality cardiovascular imaging: a
  state-of-the-art review,'' \emph{Progress in cardiovascular diseases},
  vol.~63, no.~3, pp. 367--376, 2020.

\bibitem{birhane2021multimodal}
A.~Birhane, V.~U. Prabhu, and E.~Kahembwe, ``Multimodal datasets: misogyny,
  pornography, and malignant stereotypes,'' \emph{arXiv preprint
  arXiv:2110.01963}, 2021.

\bibitem{li2021multimodal}
Y.~Li, W.~Li, N.~Li, X.~Qiu, and K.~B. Manokaran, ``Multimodal information
  interaction and fusion for the parallel computing system using ai
  techniques,'' \emph{International Journal of High Performance Systems
  Architecture}, vol.~10, no. 3-4, pp. 185--196, 2021.

\bibitem{zhang2020multimodal}
C.~Zhang, Z.~Yang, X.~He, and L.~Deng, ``Multimodal intelligence:
  Representation learning, information fusion, and applications,'' \emph{IEEE
  Journal of Selected Topics in Signal Processing}, vol.~14, no.~3, pp.
  478--493, 2020.

\bibitem{qiao2022initial}
H.~Qiao, V.~Liu, and L.~Chilton, ``Initial images: using image prompts to
  improve subject representation in multimodal ai generated art,'' in
  \emph{Proceedings of the 14th Conference on Creativity and Cognition}, 2022,
  pp. 15--28.

\bibitem{stewart2021multimodal}
A.~E. Stewart, Z.~Keirn, and S.~K. D’Mello, ``Multimodal modeling of
  collaborative problem-solving facets in triads,'' \emph{User Modeling and
  User-Adapted Interaction}, pp. 1--39, 2021.

\bibitem{xue2023ulip}
L.~Xue, N.~Yu, S.~Zhang, J.~Li, R.~Mart{\'\i}n-Mart{\'\i}n, J.~Wu, C.~Xiong,
  R.~Xu, J.~C. Niebles, and S.~Savarese, ``Ulip-2: Towards scalable multimodal
  pre-training for 3d understanding,'' \emph{arXiv preprint arXiv:2305.08275},
  2023.

\bibitem{yan2022scalability}
L.~Yan, L.~Zhao, D.~Gasevic, and R.~Martinez-Maldonado, ``Scalability,
  sustainability, and ethicality of multimodal learning analytics,'' in
  \emph{LAK22: 12th international learning analytics and knowledge conference},
  2022, pp. 13--23.

\bibitem{liu2022artificial}
Y.~Liu-Thompkins, S.~Okazaki, and H.~Li, ``Artificial empathy in marketing
  interactions: Bridging the human-ai gap in affective and social customer
  experience,'' \emph{Journal of the Academy of Marketing Science}, vol.~50,
  no.~6, pp. 1198--1218, 2022.

\bibitem{rahman2023new}
M.~S. Rahman, S.~Bag, M.~A. Hossain, F.~A. M.~A. Fattah, M.~O. Gani, and N.~P.
  Rana, ``The new wave of ai-powered luxury brands online shopping experience:
  The role of digital multisensory cues and customers’ engagement,''
  \emph{Journal of Retailing and Consumer Services}, vol.~72, p. 103273, 2023.

\bibitem{sachdeva2023rank2tell}
E.~Sachdeva, N.~Agarwal, S.~Chundi, S.~Roelofs, J.~Li, B.~Dariush, C.~Choi, and
  M.~Kochenderfer, ``Rank2tell: A multimodal driving dataset for joint
  importance ranking and reasoning,'' \emph{arXiv preprint arXiv:2309.06597},
  2023.

\bibitem{cui2023survey}
C.~Cui, Y.~Ma, X.~Cao, W.~Ye, Y.~Zhou, K.~Liang, J.~Chen, J.~Lu, Z.~Yang, K.-D.
  Liao \emph{et~al.}, ``A survey on multimodal large language models for
  autonomous driving,'' \emph{arXiv preprint arXiv:2311.12320}, 2023.

\bibitem{temsamani2022multimodal}
A.~B. Temsamani, A.~K. Chavali, W.~Vervoort, T.~Tuytelaars, G.~Radevski,
  H.~Van~Hamme, K.~Mets, M.~Hutsebaut-Buysse, T.~De~Schepper, and S.~Latr{\'e},
  ``A multimodal ai approach for intuitively instructable autonomous systems: a
  case study of an autonomous off-highway vehicle,'' in \emph{The Eighteenth
  International Conference on Autonomic and Autonomous Systems, ICAS 2022, May
  22-26, 2022, Venice, Italy}, 2022, pp. 31--39.

\bibitem{lee2022something}
J.~Lee and S.~Y. Shin, ``Something that they never said: Multimodal
  disinformation and source vividness in understanding the power of ai-enabled
  deepfake news,'' \emph{Media Psychology}, vol.~25, no.~4, pp. 531--546, 2022.

\bibitem{muppalla2023integrating}
S.~Muppalla, S.~Jia, and S.~Lyu, ``Integrating audio-visual features for
  multimodal deepfake detection,'' \emph{arXiv preprint arXiv:2310.03827},
  2023.

\bibitem{kumar2022privacy}
S.~Kumar, M.~K. Chaube, S.~N. Nenavath, S.~K. Gupta, and S.~K. Tetarave,
  ``Privacy preservation and security challenges: a new frontier multimodal
  machine learning research,'' \emph{International Journal of Sensor Networks},
  vol.~39, no.~4, pp. 227--245, 2022.

\bibitem{marchang2022assistive}
J.~Marchang and A.~Di~Nuovo, ``Assistive multimodal robotic system (amrsys):
  security and privacy issues, challenges, and possible solutions,''
  \emph{Applied Sciences}, vol.~12, no.~4, p. 2174, 2022.

\bibitem{pena2023human}
A.~Pe{\~n}a, I.~Serna, A.~Morales, J.~Fierrez, A.~Ortega, A.~Herrarte,
  M.~Alcantara, and J.~Ortega-Garcia, ``Human-centric multimodal machine
  learning: Recent advances and testbed on ai-based recruitment,'' \emph{SN
  Computer Science}, vol.~4, no.~5, p. 434, 2023.

\bibitem{wolfe2022american}
R.~Wolfe and A.~Caliskan, ``American== white in multimodal language-and-image
  ai,'' in \emph{Proceedings of the 2022 AAAI/ACM Conference on AI, Ethics, and
  Society}, 2022, pp. 800--812.

\bibitem{wolfe2023contrastive}
R.~Wolfe, Y.~Yang, B.~Howe, and A.~Caliskan, ``Contrastive language-vision ai
  models pretrained on web-scraped multimodal data exhibit sexual
  objectification bias,'' in \emph{Proceedings of the 2023 ACM Conference on
  Fairness, Accountability, and Transparency}, 2023, pp. 1174--1185.

\bibitem{afshar2022development}
M.~Afshar, B.~Sharma, D.~Dligach, M.~Oguss, R.~Brown, N.~Chhabra, H.~M.
  Thompson, T.~Markossian, C.~Joyce, M.~M. Churpek \emph{et~al.}, ``Development
  and multimodal validation of a substance misuse algorithm for referral to
  treatment using artificial intelligence (smart-ai): a retrospective deep
  learning study,'' \emph{The Lancet Digital Health}, vol.~4, no.~6, pp.
  e426--e435, 2022.

\bibitem{alwahaby2022evidence}
H.~Alwahaby, M.~Cukurova, Z.~Papamitsiou, and M.~Giannakos, ``The evidence of
  impact and ethical considerations of multimodal learning analytics: A
  systematic literature review,'' \emph{The Multimodal Learning Analytics
  Handbook}, pp. 289--325, 2022.

\bibitem{miao2023dao}
Q.~Miao, W.~Zheng, Y.~Lv, M.~Huang, W.~Ding, and F.-Y. Wang, ``Dao to hanoi via
  desci: Ai paradigm shifts from alphago to chatgpt,'' \emph{IEEE/CAA Journal
  of Automatica Sinica}, vol.~10, no.~4, pp. 877--897, 2023.

\bibitem{rong2022roadmap}
Y.~Rong, ``Roadmap of alphago to alphastar: Problems and challenges,'' in
  \emph{2nd International Conference on Artificial Intelligence, Automation,
  and High-Performance Computing (AIAHPC 2022)}, vol. 12348.\hskip 1em plus
  0.5em minus 0.4em\relax SPIE, 2022, pp. 904--914.

\bibitem{gao2022data}
Y.~Gao, M.~Zhou, D.~Liu, Z.~Yan, S.~Zhang, and D.~N. Metaxas, ``A data-scalable
  transformer for medical image segmentation: architecture, model efficiency,
  and benchmark,'' \emph{arXiv preprint arXiv:2203.00131}, 2022.

\bibitem{peebles2023scalable}
W.~Peebles and S.~Xie, ``Scalable diffusion models with transformers,'' in
  \emph{Proceedings of the IEEE/CVF International Conference on Computer
  Vision}, 2023, pp. 4195--4205.

\bibitem{pope2023efficiently}
R.~Pope, S.~Douglas, A.~Chowdhery, J.~Devlin, J.~Bradbury, J.~Heek, K.~Xiao,
  S.~Agrawal, and J.~Dean, ``Efficiently scaling transformer inference,''
  \emph{Proceedings of Machine Learning and Systems}, vol.~5, 2023.

\bibitem{ding2022convolutional}
Y.~Ding and M.~Jia, ``Convolutional transformer: An enhanced attention
  mechanism architecture for remaining useful life estimation of bearings,''
  \emph{IEEE Transactions on Instrumentation and Measurement}, vol.~71, pp.
  1--10, 2022.

\bibitem{ding2022novel}
Y.~Ding, M.~Jia, Q.~Miao, and Y.~Cao, ``A novel time--frequency transformer
  based on self--attention mechanism and its application in fault diagnosis of
  rolling bearings,'' \emph{Mechanical Systems and Signal Processing}, vol.
  168, p. 108616, 2022.

\bibitem{wang2022shift}
G.~Wang, Y.~Zhao, C.~Tang, C.~Luo, and W.~Zeng, ``When shift operation meets
  vision transformer: An extremely simple alternative to attention mechanism,''
  in \emph{Proceedings of the AAAI Conference on Artificial Intelligence},
  vol.~36, no.~2, 2022, pp. 2423--2430.

\bibitem{cai2023efficientvit}
H.~Cai, J.~Li, M.~Hu, C.~Gan, and S.~Han, ``Efficientvit: Lightweight
  multi-scale attention for high-resolution dense prediction,'' in
  \emph{Proceedings of the IEEE/CVF International Conference on Computer
  Vision}, 2023, pp. 17\,302--17\,313.

\bibitem{liu2023efficientvit}
X.~Liu, H.~Peng, N.~Zheng, Y.~Yang, H.~Hu, and Y.~Yuan, ``Efficientvit: Memory
  efficient vision transformer with cascaded group attention,'' in
  \emph{Proceedings of the IEEE/CVF Conference on Computer Vision and Pattern
  Recognition}, 2023, pp. 14\,420--14\,430.

\bibitem{li2023modified}
Y.~Li, Q.~Fan, H.~Huang, Z.~Han, and Q.~Gu, ``A modified yolov8 detection
  network for uav aerial image recognition,'' \emph{Drones}, vol.~7, no.~5, p.
  304, 2023.

\bibitem{talaat2023improved}
F.~M. Talaat and H.~ZainEldin, ``An improved fire detection approach based on
  yolo-v8 for smart cities,'' \emph{Neural Computing and Applications},
  vol.~35, no.~28, pp. 20\,939--20\,954, 2023.

\bibitem{tamang2023enhancing}
S.~Tamang, B.~Sen, A.~Pradhan, K.~Sharma, and V.~K. Singh, ``Enhancing covid-19
  safety: Exploring yolov8 object detection for accurate face mask
  classification,'' \emph{International Journal of Intelligent Systems and
  Applications in Engineering}, vol.~11, no.~2, pp. 892--897, 2023.

\bibitem{lu2022battery}
J.~Lu, R.~Xiong, J.~Tian, C.~Wang, C.-W. Hsu, N.-T. Tsou, F.~Sun, and J.~Li,
  ``Battery degradation prediction against uncertain future conditions with
  recurrent neural network enabled deep learning,'' \emph{Energy Storage
  Materials}, vol.~50, pp. 139--151, 2022.

\bibitem{onan2022bidirectional}
A.~Onan, ``Bidirectional convolutional recurrent neural network architecture
  with group-wise enhancement mechanism for text sentiment classification,''
  \emph{Journal of King Saud University-Computer and Information Sciences},
  vol.~34, no.~5, pp. 2098--2117, 2022.

\bibitem{shan2022success}
F.~Shan, X.~He, D.~J. Armaghani, P.~Zhang, and D.~Sheng, ``Success and
  challenges in predicting tbm penetration rate using recurrent neural
  networks,'' \emph{Tunnelling and Underground Space Technology}, vol. 130, p.
  104728, 2022.

\bibitem{sridhar2022optimal}
C.~Sridhar, P.~K. Pareek, R.~Kalidoss, S.~S. Jamal, P.~K. Shukla, S.~J. Nuagah
  \emph{et~al.}, ``Optimal medical image size reduction model creation using
  recurrent neural network and genpsowvq,'' \emph{Journal of Healthcare
  Engineering}, vol. 2022, 2022.

\bibitem{zhu2022application}
J.~Zhu, Q.~Jiang, Y.~Shen, C.~Qian, F.~Xu, and Q.~Zhu, ``Application of
  recurrent neural network to mechanical fault diagnosis: A review,''
  \emph{Journal of Mechanical Science and Technology}, vol.~36, no.~2, pp.
  527--542, 2022.

\bibitem{lin2023segrnn}
S.~Lin, W.~Lin, W.~Wu, F.~Zhao, R.~Mo, and H.~Zhang, ``Segrnn: Segment
  recurrent neural network for long-term time series forecasting,'' \emph{arXiv
  preprint arXiv:2308.11200}, 2023.

\bibitem{wei2022extracting}
Z.~Wei, X.~Zhang, and M.~Sun, ``Extracting weighted finite automata from
  recurrent neural networks for natural languages,'' in \emph{International
  Conference on Formal Engineering Methods}.\hskip 1em plus 0.5em minus
  0.4em\relax Springer, 2022, pp. 370--385.

\bibitem{bonassi2022recurrent}
F.~Bonassi, M.~Farina, J.~Xie, and R.~Scattolini, ``On recurrent neural
  networks for learning-based control: recent results and ideas for future
  developments,'' \emph{Journal of Process Control}, vol. 114, pp. 92--104,
  2022.

\bibitem{guo2023viewrefer}
Z.~Guo, Y.~Tang, R.~Zhang, D.~Wang, Z.~Wang, B.~Zhao, and X.~Li, ``Viewrefer:
  Grasp the multi-view knowledge for 3d visual grounding,'' in
  \emph{Proceedings of the IEEE/CVF International Conference on Computer
  Vision}, 2023, pp. 15\,372--15\,383.

\bibitem{pan2023baeformer}
C.~Pan, Y.~He, J.~Peng, Q.~Zhang, W.~Sui, and Z.~Zhang, ``Baeformer:
  Bi-directional and early interaction transformers for bird's eye view
  semantic segmentation,'' in \emph{Proceedings of the IEEE/CVF Conference on
  Computer Vision and Pattern Recognition}, 2023, pp. 9590--9599.

\bibitem{xu2023multimodal}
P.~Xu, X.~Zhu, and D.~A. Clifton, ``Multimodal learning with transformers: A
  survey,'' \emph{IEEE Transactions on Pattern Analysis and Machine
  Intelligence}, 2023.

\bibitem{molenaar2023measuring}
I.~Molenaar, S.~de~Mooij, R.~Azevedo, M.~Bannert, S.~J{\"a}rvel{\"a}, and
  D.~Ga{\v{s}}evi{\'c}, ``Measuring self-regulated learning and the role of ai:
  Five years of research using multimodal multichannel data,'' \emph{Computers
  in Human Behavior}, vol. 139, p. 107540, 2023.

\bibitem{steyaert2023multimodal}
S.~Steyaert, M.~Pizurica, D.~Nagaraj, P.~Khandelwal, T.~Hernandez-Boussard,
  A.~J. Gentles, and O.~Gevaert, ``Multimodal data fusion for cancer biomarker
  discovery with deep learning,'' \emph{Nature Machine Intelligence}, vol.~5,
  no.~4, pp. 351--362, 2023.

\bibitem{rani2023self}
V.~Rani, S.~T. Nabi, M.~Kumar, A.~Mittal, and K.~Kumar, ``Self-supervised
  learning: A succinct review,'' \emph{Archives of Computational Methods in
  Engineering}, vol.~30, no.~4, pp. 2761--2775, 2023.

\bibitem{schiappa2023self}
M.~C. Schiappa, Y.~S. Rawat, and M.~Shah, ``Self-supervised learning for
  videos: A survey,'' \emph{ACM Computing Surveys}, vol.~55, no. 13s, pp.
  1--37, 2023.

\bibitem{yu2023self}
J.~Yu, H.~Yin, X.~Xia, T.~Chen, J.~Li, and Z.~Huang, ``Self-supervised learning
  for recommender systems: A survey,'' \emph{IEEE Transactions on Knowledge and
  Data Engineering}, 2023.

\bibitem{bharti2023label}
V.~Bharti, A.~Kumar, V.~Purohit, R.~Singh, A.~K. Singh, and S.~K. Singh, ``A
  label efficient semi self-supervised learning framework for iot devices in
  industrial process,'' \emph{IEEE Transactions on Industrial Informatics},
  2023.

\bibitem{sam2023losses}
D.~Sam and J.~Z. Kolter, ``Losses over labels: Weakly supervised learning via
  direct loss construction,'' in \emph{Proceedings of the AAAI Conference on
  Artificial Intelligence}, vol.~37, no.~8, 2023, pp. 9695--9703.

\bibitem{wang2023t5}
M.~Wang, P.~Xie, Y.~Du, and X.~Hu, ``T5-based model for abstractive
  summarization: A semi-supervised learning approach with consistency loss
  functions,'' \emph{Applied Sciences}, vol.~13, no.~12, p. 7111, 2023.

\bibitem{li2022unsupervised}
Q.~Li, X.~Peng, Y.~Qiao, and Q.~Hao, ``Unsupervised person re-identification
  with multi-label learning guided self-paced clustering,'' \emph{Pattern
  Recognition}, vol. 125, p. 108521, 2022.

\bibitem{nancy2022deep}
P.~Nancy, H.~Pallathadka, M.~Naved, K.~Kaliyaperumal, K.~Arumugam, and
  V.~Garchar, ``Deep learning and machine learning based efficient framework
  for image based plant disease classification and detection,'' in \emph{2022
  International Conference on Advanced Computing Technologies and Applications
  (ICACTA)}.\hskip 1em plus 0.5em minus 0.4em\relax IEEE, 2022, pp. 1--6.

\bibitem{an2022ensemble}
P.~An, Z.~Wang, and C.~Zhang, ``Ensemble unsupervised autoencoders and gaussian
  mixture model for cyberattack detection,'' \emph{Information Processing \&
  Management}, vol.~59, no.~2, p. 102844, 2022.

\bibitem{yan2023hybrid}
S.~Yan, H.~Shao, Y.~Xiao, B.~Liu, and J.~Wan, ``Hybrid robust convolutional
  autoencoder for unsupervised anomaly detection of machine tools under
  noises,'' \emph{Robotics and Computer-Integrated Manufacturing}, vol.~79, p.
  102441, 2023.

\bibitem{ayanoglu2022machine}
E.~Ayanoglu, K.~Davaslioglu, and Y.~E. Sagduyu, ``Machine learning in nextg
  networks via generative adversarial networks,'' \emph{IEEE Transactions on
  Cognitive Communications and Networking}, vol.~8, no.~2, pp. 480--501, 2022.

\bibitem{yan2022physical}
K.~Yan, X.~Chen, X.~Zhou, Z.~Yan, and J.~Ma, ``Physical model informed fault
  detection and diagnosis of air handling units based on transformer generative
  adversarial network,'' \emph{IEEE Transactions on Industrial Informatics},
  vol.~19, no.~2, pp. 2192--2199, 2022.

\bibitem{zhou2023hybrid}
N.-R. Zhou, T.-F. Zhang, X.-W. Xie, and J.-Y. Wu, ``Hybrid quantum--classical
  generative adversarial networks for image generation via learning discrete
  distribution,'' \emph{Signal Processing: Image Communication}, vol. 110, p.
  116891, 2023.

\bibitem{ladosz2022exploration}
P.~Ladosz, L.~Weng, M.~Kim, and H.~Oh, ``Exploration in deep reinforcement
  learning: A survey,'' \emph{Information Fusion}, vol.~85, pp. 1--22, 2022.

\bibitem{matsuo2022deep}
Y.~Matsuo, Y.~LeCun, M.~Sahani, D.~Precup, D.~Silver, M.~Sugiyama, E.~Uchibe,
  and J.~Morimoto, ``Deep learning, reinforcement learning, and world models,''
  \emph{Neural Networks}, vol. 152, pp. 267--275, 2022.

\bibitem{bertoin2022look}
D.~Bertoin, A.~Zouitine, M.~Zouitine, and E.~Rachelson, ``Look where you look!
  saliency-guided q-networks for generalization in visual reinforcement
  learning,'' \emph{Advances in Neural Information Processing Systems},
  vol.~35, pp. 30\,693--30\,706, 2022.

\bibitem{hafiz2022survey}
A.~Hafiz, ``A survey of deep q-networks used for reinforcement learning: State
  of the art,'' \emph{Intelligent Communication Technologies and Virtual Mobile
  Networks: Proceedings of ICICV 2022}, pp. 393--402, 2022.

\bibitem{hafiz2023reinforcement}
A.~Hafiz, M.~Hassaballah, A.~Alqahtani, S.~Alsubai, and M.~A. Hameed,
  ``Reinforcement learning with an ensemble of binary action deep q-networks.''
  \emph{Computer Systems Science \& Engineering}, vol.~46, no.~3, 2023.

\bibitem{alagha2022target}
A.~Alagha, S.~Singh, R.~Mizouni, J.~Bentahar, and H.~Otrok, ``Target
  localization using multi-agent deep reinforcement learning with proximal
  policy optimization,'' \emph{Future Generation Computer Systems}, vol. 136,
  pp. 342--357, 2022.

\bibitem{hassan20223to}
S.~S. Hassan, Y.~M. Park, Y.~K. Tun, W.~Saad, Z.~Han, and C.~S. Hong, ``3to:
  Thz-enabled throughput and trajectory optimization of uavs in 6g networks by
  proximal policy optimization deep reinforcement learning,'' in \emph{ICC
  2022-IEEE International Conference on Communications}.\hskip 1em plus 0.5em
  minus 0.4em\relax IEEE, 2022, pp. 5712--5718.

\bibitem{jayant2022model}
A.~K. Jayant and S.~Bhatnagar, ``Model-based safe deep reinforcement learning
  via a constrained proximal policy optimization algorithm,'' \emph{Advances in
  Neural Information Processing Systems}, vol.~35, pp. 24\,432--24\,445, 2022.

\bibitem{lin2023reinforcement}
B.~Lin, ``Reinforcement learning and bandits for speech and language
  processing: Tutorial, review and outlook,'' \emph{Expert Systems with
  Applications}, p. 122254, 2023.

\bibitem{luo2023human}
B.~Luo, Z.~Wu, F.~Zhou, and B.-C. Wang, ``Human-in-the-loop reinforcement
  learning in continuous-action space,'' \emph{IEEE Transactions on Neural
  Networks and Learning Systems}, 2023.

\bibitem{raza2022designing}
A.~Raza, K.~P. Tran, L.~Koehl, and S.~Li, ``Designing ecg monitoring healthcare
  system with federated transfer learning and explainable ai,''
  \emph{Knowledge-Based Systems}, vol. 236, p. 107763, 2022.

\bibitem{siahpour2022novel}
S.~Siahpour, X.~Li, and J.~Lee, ``A novel transfer learning approach in
  remaining useful life prediction for incomplete dataset,'' \emph{IEEE
  Transactions on Instrumentation and Measurement}, vol.~71, pp. 1--11, 2022.

\bibitem{guo2022transfer}
Z.~Guo, K.~Lin, X.~Chen, and C.-Y. Chit, ``Transfer learning for angle of
  arrivals estimation in massive mimo system,'' in \emph{2022 IEEE/CIC
  International Conference on Communications in China (ICCC)}.\hskip 1em plus
  0.5em minus 0.4em\relax IEEE, 2022, pp. 506--511.

\bibitem{liu2022adaptive}
S.~Liu, Y.~Lu, P.~Zheng, H.~Shen, and J.~Bao, ``Adaptive reconstruction of
  digital twins for machining systems: A transfer learning approach,''
  \emph{Robotics and Computer-Integrated Manufacturing}, vol.~78, p. 102390,
  2022.

\bibitem{liu2023logiqa}
H.~Liu, J.~Liu, L.~Cui, Z.~Teng, N.~Duan, M.~Zhou, and Y.~Zhang, ``Logiqa
  2.0—an improved dataset for logical reasoning in natural language
  understanding,'' \emph{IEEE/ACM Transactions on Audio, Speech, and Language
  Processing}, 2023.

\bibitem{meng2022generating}
Y.~Meng, J.~Huang, Y.~Zhang, and J.~Han, ``Generating training data with
  language models: Towards zero-shot language understanding,'' \emph{Advances
  in Neural Information Processing Systems}, vol.~35, pp. 462--477, 2022.

\bibitem{samant2022framework}
R.~M. Samant, M.~R. Bachute, S.~Gite, and K.~Kotecha, ``Framework for deep
  learning-based language models using multi-task learning in natural language
  understanding: A systematic literature review and future directions,''
  \emph{IEEE Access}, vol.~10, pp. 17\,078--17\,097, 2022.

\bibitem{weld2022survey}
H.~Weld, X.~Huang, S.~Long, J.~Poon, and S.~C. Han, ``A survey of joint intent
  detection and slot filling models in natural language understanding,''
  \emph{ACM Computing Surveys}, vol.~55, no.~8, pp. 1--38, 2022.

\bibitem{ajmal2023natural}
S.~Ajmal, A.~A.~I. Ahmed, and C.~Jalota, ``Natural language processing in
  improving information retrieval and knowledge discovery in healthcare
  conversational agents,'' \emph{Journal of Artificial Intelligence and Machine
  Learning in Management}, vol.~7, no.~1, pp. 34--47, 2023.

\bibitem{montejo2022current}
A.~Montejo-R{\'a}ez and S.~M. Jim{\'e}nez-Zafra, ``Current approaches and
  applications in natural language processing,'' \emph{Applied Sciences},
  vol.~12, no.~10, p. 4859, 2022.

\bibitem{vijayan2022language}
K.~Vijayan, O.~Anand, and A.~Sahaj, ``Language-agnostic text processing for
  information extraction,'' in \emph{CS \& IT Conference Proceedings}, vol.~12,
  no.~23.\hskip 1em plus 0.5em minus 0.4em\relax CS \& IT Conference
  Proceedings, 2022.

\bibitem{manning2022human}
C.~D. Manning, ``Human language understanding \& reasoning,'' \emph{Daedalus},
  vol. 151, no.~2, pp. 127--138, 2022.

\bibitem{peng2023gpt}
W.~Peng, D.~Xu, T.~Xu, J.~Zhang, and E.~Chen, ``Are gpt embeddings useful for
  ads and recommendation?'' in \emph{International Conference on Knowledge
  Science, Engineering and Management}.\hskip 1em plus 0.5em minus 0.4em\relax
  Springer, 2023, pp. 151--162.

\bibitem{erdem2022neural}
E.~Erdem, M.~Kuyu, S.~Yagcioglu, A.~Frank, L.~Parcalabescu, B.~Plank, A.~Babii,
  O.~Turuta, A.~Erdem, I.~Calixto \emph{et~al.}, ``Neural natural language
  generation: A survey on multilinguality, multimodality, controllability and
  learning,'' \emph{Journal of Artificial Intelligence Research}, vol.~73, pp.
  1131--1207, 2022.

\bibitem{qian2022controllable}
J.~Qian, L.~Dong, Y.~Shen, F.~Wei, and W.~Chen, ``Controllable natural language
  generation with contrastive prefixes,'' \emph{arXiv preprint
  arXiv:2202.13257}, 2022.

\bibitem{rashkin2023measuring}
H.~Rashkin, V.~Nikolaev, M.~Lamm, L.~Aroyo, M.~Collins, D.~Das, S.~Petrov,
  G.~S. Tomar, I.~Turc, and D.~Reitter, ``Measuring attribution in natural
  language generation models,'' \emph{Computational Linguistics}, pp. 1--64,
  2023.

\bibitem{pandey2023natural}
A.~K. Pandey and S.~S. Roy, ``Natural language generation using sequential
  models: A survey,'' \emph{Neural Processing Letters}, pp. 1--34, 2023.

\bibitem{khan2022automatic}
J.~Y. Khan and G.~Uddin, ``Automatic code documentation generation using
  gpt-3,'' in \emph{Proceedings of the 37th IEEE/ACM International Conference
  on Automated Software Engineering}, 2022, pp. 1--6.

\bibitem{dwivedi2023so}
Y.~K. Dwivedi, N.~Kshetri, L.~Hughes, E.~L. Slade, A.~Jeyaraj, A.~K. Kar, A.~M.
  Baabdullah, A.~Koohang, V.~Raghavan, M.~Ahuja \emph{et~al.}, ``“so what if
  chatgpt wrote it?” multidisciplinary perspectives on opportunities,
  challenges and implications of generative conversational ai for research,
  practice and policy,'' \emph{International Journal of Information
  Management}, vol.~71, p. 102642, 2023.

\bibitem{fu2022learning}
T.~Fu, S.~Gao, X.~Zhao, J.-r. Wen, and R.~Yan, ``Learning towards
  conversational ai: A survey,'' \emph{AI Open}, vol.~3, pp. 14--28, 2022.

\bibitem{ji2023systematic}
H.~Ji, I.~Han, and Y.~Ko, ``A systematic review of conversational ai in
  language education: Focusing on the collaboration with human teachers,''
  \emph{Journal of Research on Technology in Education}, vol.~55, no.~1, pp.
  48--63, 2023.

\bibitem{wan2023biasasker}
Y.~Wan, W.~Wang, P.~He, J.~Gu, H.~Bai, and M.~R. Lyu, ``Biasasker: Measuring
  the bias in conversational ai system,'' in \emph{Proceedings of the 31st ACM
  Joint European Software Engineering Conference and Symposium on the
  Foundations of Software Engineering}, 2023, pp. 515--527.

\bibitem{kusal2022ai}
S.~Kusal, S.~Patil, J.~Choudrie, K.~Kotecha, S.~Mishra, and A.~Abraham,
  ``Ai-based conversational agents: A scoping review from technologies to
  future directions,'' \emph{IEEE Access}, 2022.

\bibitem{xiao2023seeing}
Z.~Xiao, ``Seeing us through machines: designing and building conversational ai
  to understand humans,'' Ph.D. dissertation, University of Illinois at
  Urbana-Champaign, 2023.

\bibitem{ko2023large}
H.-K. Ko, G.~Park, H.~Jeon, J.~Jo, J.~Kim, and J.~Seo, ``Large-scale
  text-to-image generation models for visual artists’ creative works,'' in
  \emph{Proceedings of the 28th International Conference on Intelligent User
  Interfaces}, 2023, pp. 919--933.

\bibitem{pearson2023rise}
A.~Pearson, ``The rise of crealtives: Using ai to enable and speed up the
  creative process,'' \emph{Journal of AI, Robotics \& Workplace Automation},
  vol.~2, no.~2, pp. 101--114, 2023.

\bibitem{rezwana2023designing}
J.~Rezwana and M.~L. Maher, ``Designing creative ai partners with cofi: A
  framework for modeling interaction in human-ai co-creative systems,''
  \emph{ACM Transactions on Computer-Human Interaction}, vol.~30, no.~5, pp.
  1--28, 2023.

\bibitem{sharma2023generative}
S.~Sharma and S.~Bvuma, ``Generative adversarial networks (gans) for creative
  applications: Exploring art and music generation,'' \emph{International
  Journal of Multidisciplinary Innovation and Research Methodology, ISSN:
  2960-2068}, vol.~2, no.~4, pp. 29--33, 2023.

\bibitem{attard2023ethics}
B.~Attard-Frost, A.~De~los R{\'\i}os, and D.~R. Walters, ``The ethics of ai
  business practices: a review of 47 ai ethics guidelines,'' \emph{AI and
  Ethics}, vol.~3, no.~2, pp. 389--406, 2023.

\bibitem{gardner2022ethical}
A.~Gardner, A.~L. Smith, A.~Steventon, E.~Coughlan, and M.~Oldfield, ``Ethical
  funding for trustworthy ai: proposals to address the responsibilities of
  funders to ensure that projects adhere to trustworthy ai practice,'' \emph{AI
  and Ethics}, pp. 1--15, 2022.

\bibitem{schuett2023three}
J.~Schuett, ``Three lines of defense against risks from ai,'' \emph{AI \&
  SOCIETY}, pp. 1--15, 2023.

\bibitem{sloane2022german}
M.~Sloane and J.~Zakrzewski, ``German ai start-ups and “ai ethics”: Using a
  social practice lens for assessing and implementing socio-technical
  innovation,'' in \emph{Proceedings of the 2022 ACM Conference on Fairness,
  Accountability, and Transparency}, 2022, pp. 935--947.

\bibitem{vasconcelos2018modeling}
M.~Vasconcelos, C.~Cardonha, and B.~Gon{\c{c}}alves, ``Modeling epistemological
  principles for bias mitigation in ai systems: an illustration in hiring
  decisions,'' in \emph{Proceedings of the 2018 AAAI/ACM Conference on AI,
  Ethics, and Society}, 2018, pp. 323--329.

\bibitem{yang2022enhancing}
Y.~Yang, A.~Gupta, J.~Feng, P.~Singhal, V.~Yadav, Y.~Wu, P.~Natarajan,
  V.~Hedau, and J.~Joo, ``Enhancing fairness in face detection in computer
  vision systems by demographic bias mitigation,'' in \emph{Proceedings of the
  2022 AAAI/ACM Conference on AI, Ethics, and Society}, 2022, pp. 813--822.

\bibitem{schwartz2022towards}
R.~Schwartz, A.~Vassilev, K.~Greene, L.~Perine, A.~Burt, P.~Hall \emph{et~al.},
  ``Towards a standard for identifying and managing bias in artificial
  intelligence,'' \emph{NIST special publication}, vol. 1270, no. 10.6028,
  2022.

\bibitem{guo2021detecting}
W.~Guo and A.~Caliskan, ``Detecting emergent intersectional biases:
  Contextualized word embeddings contain a distribution of human-like biases,''
  in \emph{Proceedings of the 2021 AAAI/ACM Conference on AI, Ethics, and
  Society}, 2021, pp. 122--133.

\bibitem{kong2022intersectionally}
Y.~Kong, ``Are “intersectionally fair” ai algorithms really fair to women
  of color? a philosophical analysis,'' in \emph{Proceedings of the 2022 ACM
  Conference on Fairness, Accountability, and Transparency}, 2022, pp.
  485--494.

\bibitem{tan2019assessing}
Y.~C. Tan and L.~E. Celis, ``Assessing social and intersectional biases in
  contextualized word representations,'' \emph{Advances in neural information
  processing systems}, vol.~32, 2019.

\bibitem{cheng2021causal}
L.~Cheng, A.~Mosallanezhad, P.~Sheth, and H.~Liu, ``Causal learning for
  socially responsible ai,'' \emph{arXiv preprint arXiv:2104.12278}, 2021.

\bibitem{correa2019identification}
J.~D. Correa, J.~Tian, and E.~Bareinboim, ``Identification of causal effects in
  the presence of selection bias,'' in \emph{Proceedings of the AAAI Conference
  on Artificial Intelligence}, vol.~33, no.~01, 2019, pp. 2744--2751.

\bibitem{ghai2022d}
B.~Ghai and K.~Mueller, ``D-bias: a causality-based human-in-the-loop system
  for tackling algorithmic bias,'' \emph{IEEE Transactions on Visualization and
  Computer Graphics}, vol.~29, no.~1, pp. 473--482, 2022.

\bibitem{yan2020silva}
J.~N. Yan, Z.~Gu, H.~Lin, and J.~M. Rzeszotarski, ``Silva: Interactively
  assessing machine learning fairness using causality,'' in \emph{Proceedings
  of the 2020 chi conference on human factors in computing systems}, 2020, pp.
  1--13.

\bibitem{bertino2021ai}
E.~Bertino, M.~Kantarcioglu, C.~G. Akcora, S.~Samtani, S.~Mittal, and M.~Gupta,
  ``Ai for security and security for ai,'' in \emph{Proceedings of the Eleventh
  ACM Conference on Data and Application Security and Privacy}, 2021, pp.
  333--334.

\bibitem{susanto2021data}
H.~Susanto, L.~F. Yie, D.~Rosiyadi, A.~I. Basuki, and D.~Setiana, ``Data
  security for connected governments and organisations: Managing automation and
  artificial intelligence,'' in \emph{Web 2.0 and cloud technologies for
  implementing connected government}.\hskip 1em plus 0.5em minus 0.4em\relax
  IGI Global, 2021, pp. 229--251.

\bibitem{dilmaghani2019privacy}
S.~Dilmaghani, M.~R. Brust, G.~Danoy, N.~Cassagnes, J.~Pecero, and P.~Bouvry,
  ``Privacy and security of big data in ai systems: A research and standards
  perspective,'' in \emph{2019 IEEE International Conference on Big Data (Big
  Data)}.\hskip 1em plus 0.5em minus 0.4em\relax IEEE, 2019, pp. 5737--5743.

\bibitem{mcintosh2022intercepting}
T.~McIntosh, ``Intercepting ransomware attacks with staged event-driven access
  control,'' Ph.D. dissertation, La Trobe, 2022.

\bibitem{mcintosh2023applying}
T.~McIntosh, A.~Kayes, Y.-P.~P. Chen, A.~Ng, and P.~Watters, ``Applying staged
  event-driven access control to combat ransomware,'' \emph{Computers \&
  Security}, vol. 128, p. 103160, 2023.

\bibitem{hummel2021data}
P.~Hummel, M.~Braun, M.~Tretter, and P.~Dabrock, ``Data sovereignty: A
  review,'' \emph{Big Data \& Society}, vol.~8, no.~1, p. 2053951720982012,
  2021.

\bibitem{lukings2022data}
M.~Lukings and A.~Habibi~Lashkari, ``Data sovereignty,'' in \emph{Understanding
  Cybersecurity Law in Data Sovereignty and Digital Governance: An Overview
  from a Legal Perspective}.\hskip 1em plus 0.5em minus 0.4em\relax Springer,
  2022, pp. 1--38.

\bibitem{hickok2021lessons}
M.~Hickok, ``Lessons learned from ai ethics principles for future actions,''
  \emph{AI and Ethics}, vol.~1, no.~1, pp. 41--47, 2021.

\bibitem{zhou2022ai}
J.~Zhou and F.~Chen, ``Ai ethics: From principles to practice,'' \emph{AI \&
  SOCIETY}, pp. 1--11, 2022.

\bibitem{kroll2021outlining}
J.~A. Kroll, ``Outlining traceability: A principle for operationalizing
  accountability in computing systems,'' in \emph{Proceedings of the 2021 ACM
  Conference on Fairness, Accountability, and Transparency}, 2021, pp.
  758--771.

\bibitem{oseni2021security}
A.~Oseni, N.~Moustafa, H.~Janicke, P.~Liu, Z.~Tari, and A.~Vasilakos,
  ``Security and privacy for artificial intelligence: Opportunities and
  challenges,'' \emph{arXiv preprint arXiv:2102.04661}, 2021.

\bibitem{stahl2018ethics}
B.~C. Stahl and D.~Wright, ``Ethics and privacy in ai and big data:
  Implementing responsible research and innovation,'' \emph{IEEE Security \&
  Privacy}, vol.~16, no.~3, pp. 26--33, 2018.

\bibitem{ma2023trusted}
C.~Ma, J.~Li, K.~Wei, B.~Liu, M.~Ding, L.~Yuan, Z.~Han, and H.~V. Poor,
  ``Trusted ai in multiagent systems: An overview of privacy and security for
  distributed learning,'' \emph{Proceedings of the IEEE}, vol. 111, no.~9, pp.
  1097--1132, 2023.

\bibitem{song2020analyzing}
M.~Song, Z.~Wang, Z.~Zhang, Y.~Song, Q.~Wang, J.~Ren, and H.~Qi, ``Analyzing
  user-level privacy attack against federated learning,'' \emph{IEEE Journal on
  Selected Areas in Communications}, vol.~38, no.~10, pp. 2430--2444, 2020.

\bibitem{misra2020self}
I.~Misra and L.~v.~d. Maaten, ``Self-supervised learning of pretext-invariant
  representations,'' in \emph{Proceedings of the IEEE/CVF conference on
  computer vision and pattern recognition}, 2020, pp. 6707--6717.

\bibitem{zhai2019s4l}
X.~Zhai, A.~Oliver, A.~Kolesnikov, and L.~Beyer, ``S4l: Self-supervised
  semi-supervised learning,'' in \emph{Proceedings of the IEEE/CVF
  international conference on computer vision}, 2019, pp. 1476--1485.

\bibitem{chen2019self}
T.~Chen, X.~Zhai, M.~Ritter, M.~Lucic, and N.~Houlsby, ``Self-supervised gans
  via auxiliary rotation loss,'' in \emph{Proceedings of the IEEE/CVF
  conference on computer vision and pattern recognition}, 2019, pp.
  12\,154--12\,163.

\bibitem{jenni2018self}
S.~Jenni and P.~Favaro, ``Self-supervised feature learning by learning to spot
  artifacts,'' in \emph{Proceedings of the IEEE Conference on Computer Vision
  and Pattern Recognition}, 2018, pp. 2733--2742.

\bibitem{patel2021lt}
P.~Patel, N.~Kumari, M.~Singh, and B.~Krishnamurthy, ``Lt-gan: Self-supervised
  gan with latent transformation detection,'' in \emph{Proceedings of the
  IEEE/CVF winter conference on applications of computer vision}, 2021, pp.
  3189--3198.

\bibitem{chen2020simple}
T.~Chen, S.~Kornblith, M.~Norouzi, and G.~Hinton, ``A simple framework for
  contrastive learning of visual representations,'' in \emph{International
  conference on machine learning}.\hskip 1em plus 0.5em minus 0.4em\relax PMLR,
  2020, pp. 1597--1607.

\bibitem{he2020momentum}
K.~He, H.~Fan, Y.~Wu, S.~Xie, and R.~Girshick, ``Momentum contrast for
  unsupervised visual representation learning,'' in \emph{Proceedings of the
  IEEE/CVF conference on computer vision and pattern recognition}, 2020, pp.
  9729--9738.

\bibitem{liu2021tera}
A.~T. Liu, S.-W. Li, and H.-y. Lee, ``Tera: Self-supervised learning of
  transformer encoder representation for speech,'' \emph{IEEE/ACM Transactions
  on Audio, Speech, and Language Processing}, vol.~29, pp. 2351--2366, 2021.

\bibitem{pang2022masked}
Y.~Pang, W.~Wang, F.~E. Tay, W.~Liu, Y.~Tian, and L.~Yuan, ``Masked
  autoencoders for point cloud self-supervised learning,'' in \emph{European
  conference on computer vision}.\hskip 1em plus 0.5em minus 0.4em\relax
  Springer, 2022, pp. 604--621.

\bibitem{hospedales2021meta}
T.~Hospedales, A.~Antoniou, P.~Micaelli, and A.~Storkey, ``Meta-learning in
  neural networks: A survey,'' \emph{IEEE transactions on pattern analysis and
  machine intelligence}, vol.~44, no.~9, pp. 5149--5169, 2021.

\bibitem{vilalta2002perspective}
R.~Vilalta and Y.~Drissi, ``A perspective view and survey of meta-learning,''
  \emph{Artificial intelligence review}, vol.~18, pp. 77--95, 2002.

\bibitem{al2021data}
M.~Al-Shedivat, L.~Li, E.~Xing, and A.~Talwalkar, ``On data efficiency of
  meta-learning,'' in \emph{International Conference on Artificial Intelligence
  and Statistics}.\hskip 1em plus 0.5em minus 0.4em\relax PMLR, 2021, pp.
  1369--1377.

\bibitem{hu2021task}
Y.~Hu, R.~Liu, X.~Li, D.~Chen, and Q.~Hu, ``Task-sequencing meta learning for
  intelligent few-shot fault diagnosis with limited data,'' \emph{IEEE
  Transactions on Industrial Informatics}, vol.~18, no.~6, pp. 3894--3904,
  2021.

\bibitem{baik2021meta}
S.~Baik, J.~Choi, H.~Kim, D.~Cho, J.~Min, and K.~M. Lee, ``Meta-learning with
  task-adaptive loss function for few-shot learning,'' in \emph{Proceedings of
  the IEEE/CVF international conference on computer vision}, 2021, pp.
  9465--9474.

\bibitem{chen2021meta}
Y.~Chen, Z.~Liu, H.~Xu, T.~Darrell, and X.~Wang, ``Meta-baseline: Exploring
  simple meta-learning for few-shot learning,'' in \emph{Proceedings of the
  IEEE/CVF international conference on computer vision}, 2021, pp. 9062--9071.

\bibitem{jamal2019task}
M.~A. Jamal and G.-J. Qi, ``Task agnostic meta-learning for few-shot
  learning,'' in \emph{Proceedings of the IEEE/CVF Conference on Computer
  Vision and Pattern Recognition}, 2019, pp. 11\,719--11\,727.

\bibitem{behnia2022ew}
R.~Behnia, M.~R. Ebrahimi, J.~Pacheco, and B.~Padmanabhan, ``Ew-tune: A
  framework for privately fine-tuning large language models with differential
  privacy,'' in \emph{2022 IEEE International Conference on Data Mining
  Workshops (ICDMW)}.\hskip 1em plus 0.5em minus 0.4em\relax IEEE, 2022, pp.
  560--566.

\bibitem{wei2021finetuned}
J.~Wei, M.~Bosma, V.~Y. Zhao, K.~Guu, A.~W. Yu, B.~Lester, N.~Du, A.~M. Dai,
  and Q.~V. Le, ``Finetuned language models are zero-shot learners,''
  \emph{arXiv preprint arXiv:2109.01652}, 2021.

\bibitem{kuang2023federatedscope}
W.~Kuang, B.~Qian, Z.~Li, D.~Chen, D.~Gao, X.~Pan, Y.~Xie, Y.~Li, B.~Ding, and
  J.~Zhou, ``Federatedscope-llm: A comprehensive package for fine-tuning large
  language models in federated learning,'' \emph{arXiv preprint
  arXiv:2309.00363}, 2023.

\bibitem{nguyen2023efficient}
M.~Nguyen, K.~Kishan, T.~Nguyen, A.~Chadha, and T.~Vu, ``Efficient fine-tuning
  large language models for knowledge-aware response planning,'' in \emph{Joint
  European Conference on Machine Learning and Knowledge Discovery in
  Databases}.\hskip 1em plus 0.5em minus 0.4em\relax Springer, 2023, pp.
  593--611.

\bibitem{engelbach2023fine}
M.~Engelbach, D.~Klau, F.~Scheerer, J.~Drawehn, and M.~Kintz, ``Fine-tuning and
  aligning question answering models for complex information extraction
  tasks,'' \emph{arXiv preprint arXiv:2309.14805}, 2023.

\bibitem{nguyen2023fine}
T.~T. Nguyen, C.~Wilson, and J.~Dalins, ``Fine-tuning llama 2 large language
  models for detecting online sexual predatory chats and abusive texts,''
  \emph{arXiv preprint arXiv:2308.14683}, 2023.

\bibitem{zhou2023regionblip}
Q.~Zhou, C.~Yu, S.~Zhang, S.~Wu, Z.~Wang, and F.~Wang, ``Regionblip: A unified
  multi-modal pre-training framework for holistic and regional comprehension,''
  \emph{arXiv preprint arXiv:2308.02299}, 2023.

\bibitem{arnold2017value}
T.~Arnold and D.~Kasenberg, ``Value alignment or misalignment - what will keep
  systems accountable?'' in \emph{AAAI Workshop on AI, Ethics, and Society},
  2017.

\bibitem{gabriel2021challenge}
I.~Gabriel and V.~Ghazavi, ``The challenge of value alignment: From fairer
  algorithms to ai safety,'' \emph{arXiv preprint arXiv:2101.06060}, 2021.

\bibitem{nyholm2023responsibility}
S.~Nyholm, ``Responsibility gaps, value alignment, and meaningful human control
  over artificial intelligence,'' in \emph{Risk and responsibility in
  context}.\hskip 1em plus 0.5em minus 0.4em\relax Routledge, 2023, pp.
  191--213.

\bibitem{wu2023next}
S.~Wu, H.~Fei, L.~Qu, W.~Ji, and T.-S. Chua, ``Next-gpt: Any-to-any multimodal
  llm,'' \emph{arXiv preprint arXiv:2309.05519}, 2023.

\bibitem{bayoudh2021survey}
K.~Bayoudh, R.~Knani, F.~Hamdaoui, and A.~Mtibaa, ``A survey on deep multimodal
  learning for computer vision: advances, trends, applications, and datasets,''
  \emph{The Visual Computer}, pp. 1--32, 2021.

\bibitem{hu2019scalable}
P.~Hu, L.~Zhen, D.~Peng, and P.~Liu, ``Scalable deep multimodal learning for
  cross-modal retrieval,'' in \emph{Proceedings of the 42nd international ACM
  SIGIR conference on research and development in information retrieval}, 2019,
  pp. 635--644.

\bibitem{rahate2022multimodal}
A.~Rahate, R.~Walambe, S.~Ramanna, and K.~Kotecha, ``Multimodal co-learning:
  Challenges, applications with datasets, recent advances and future
  directions,'' \emph{Information Fusion}, vol.~81, pp. 203--239, 2022.

\bibitem{che2023multimodal}
L.~Che, J.~Wang, Y.~Zhou, and F.~Ma, ``Multimodal federated learning: A
  survey,'' \emph{Sensors}, vol.~23, no.~15, p. 6986, 2023.

\bibitem{liang2021multibench}
P.~P. Liang, Y.~Lyu, X.~Fan, Z.~Wu, Y.~Cheng, J.~Wu, L.~Chen, P.~Wu, M.~A. Lee,
  Y.~Zhu \emph{et~al.}, ``Multibench: Multiscale benchmarks for multimodal
  representation learning,'' \emph{arXiv preprint arXiv:2107.07502}, 2021.

\bibitem{ashktorab2020human}
Z.~Ashktorab, Q.~V. Liao, C.~Dugan, J.~Johnson, Q.~Pan, W.~Zhang, S.~Kumaravel,
  and M.~Campbell, ``Human-ai collaboration in a cooperative game setting:
  Measuring social perception and outcomes,'' \emph{Proceedings of the ACM on
  Human-Computer Interaction}, vol.~4, no. CSCW2, pp. 1--20, 2020.

\bibitem{esmaeilzadeh2021patients}
P.~Esmaeilzadeh, T.~Mirzaei, and S.~Dharanikota, ``Patients’ perceptions
  toward human--artificial intelligence interaction in health care:
  experimental study,'' \emph{Journal of medical Internet research}, vol.~23,
  no.~11, p. e25856, 2021.

\bibitem{nazar2021systematic}
M.~Nazar, M.~M. Alam, E.~Yafi, and M.~M. Su’ud, ``A systematic review of
  human--computer interaction and explainable artificial intelligence in
  healthcare with artificial intelligence techniques,'' \emph{IEEE Access},
  vol.~9, pp. 153\,316--153\,348, 2021.

\bibitem{rajawat2021robotic}
A.~S. Rajawat, R.~Rawat, K.~Barhanpurkar, R.~N. Shaw, and A.~Ghosh, ``Robotic
  process automation with increasing productivity and improving product quality
  using artificial intelligence and machine learning,'' in \emph{Artificial
  Intelligence for Future Generation Robotics}.\hskip 1em plus 0.5em minus
  0.4em\relax Elsevier, 2021, pp. 1--13.

\bibitem{mohseni2021multidisciplinary}
S.~Mohseni, N.~Zarei, and E.~D. Ragan, ``A multidisciplinary survey and
  framework for design and evaluation of explainable ai systems,'' \emph{ACM
  Transactions on Interactive Intelligent Systems (TiiS)}, vol.~11, no. 3-4,
  pp. 1--45, 2021.

\bibitem{buehler2020theory}
M.~C. Buehler and T.~H. Weisswange, ``Theory of mind based communication for
  human agent cooperation,'' in \emph{2020 IEEE International Conference on
  Human-Machine Systems (ICHMS)}.\hskip 1em plus 0.5em minus 0.4em\relax IEEE,
  2020, pp. 1--6.

\bibitem{ccelikok2019interactive}
M.~M. {\c{C}}elikok, T.~Peltola, P.~Daee, and S.~Kaski, ``Interactive ai with a
  theory of mind,'' \emph{arXiv preprint arXiv:1912.05284}, 2019.

\bibitem{dafoe2020open}
A.~Dafoe, E.~Hughes, Y.~Bachrach, T.~Collins, K.~R. McKee, J.~Z. Leibo,
  K.~Larson, and T.~Graepel, ``Open problems in cooperative ai,'' \emph{arXiv
  preprint arXiv:2012.08630}, 2020.

\bibitem{bubeck2023sparks}
S.~Bubeck, V.~Chandrasekaran, R.~Eldan, J.~Gehrke, E.~Horvitz, E.~Kamar,
  P.~Lee, Y.~T. Lee, Y.~Li, S.~Lundberg \emph{et~al.}, ``Sparks of artificial
  general intelligence: Early experiments with gpt-4,'' \emph{arXiv preprint
  arXiv:2303.12712}, 2023.

\bibitem{fei2022towards}
N.~Fei, Z.~Lu, Y.~Gao, G.~Yang, Y.~Huo, J.~Wen, H.~Lu, R.~Song, X.~Gao,
  T.~Xiang \emph{et~al.}, ``Towards artificial general intelligence via a
  multimodal foundation model,'' \emph{Nature Communications}, vol.~13, no.~1,
  p. 3094, 2022.

\bibitem{williams2021understanding}
R.~Williams and R.~Yampolskiy, ``Understanding and avoiding ai failures: A
  practical guide,'' \emph{Philosophies}, vol.~6, no.~3, p.~53, 2021.

\bibitem{fedus2022switch}
W.~Fedus, B.~Zoph, and N.~Shazeer, ``Switch transformers: Scaling to trillion
  parameter models with simple and efficient sparsity,'' \emph{The Journal of
  Machine Learning Research}, vol.~23, no.~1, pp. 5232--5270, 2022.

\bibitem{shen2023mixture}
S.~Shen, L.~Hou, Y.~Zhou, N.~Du, S.~Longpre, J.~Wei, H.~W. Chung, B.~Zoph,
  W.~Fedus, X.~Chen \emph{et~al.}, ``Mixture-of-experts meets instruction
  tuning: A winning combination for large language models,'' \emph{arXiv
  preprint arXiv:2305.14705}, 2023.

\bibitem{rajbhandari2022deepspeed}
S.~Rajbhandari, C.~Li, Z.~Yao, M.~Zhang, R.~Y. Aminabadi, A.~A. Awan,
  J.~Rasley, and Y.~He, ``Deepspeed-moe: Advancing mixture-of-experts inference
  and training to power next-generation ai scale,'' in \emph{International
  Conference on Machine Learning}.\hskip 1em plus 0.5em minus 0.4em\relax PMLR,
  2022, pp. 18\,332--18\,346.

\bibitem{shen2022se}
L.~Shen, Z.~Wu, W.~Gong, H.~Hao, Y.~Bai, H.~Wu, X.~Wu, J.~Bian, H.~Xiong, D.~Yu
  \emph{et~al.}, ``Se-moe: A scalable and efficient mixture-of-experts
  distributed training and inference system,'' \emph{arXiv preprint
  arXiv:2205.10034}, 2022.

\bibitem{hwang2023tutel}
C.~Hwang, W.~Cui, Y.~Xiong, Z.~Yang, Z.~Liu, H.~Hu, Z.~Wang, R.~Salas, J.~Jose,
  P.~Ram \emph{et~al.}, ``Tutel: Adaptive mixture-of-experts at scale,''
  \emph{Proceedings of Machine Learning and Systems}, vol.~5, 2023.

\bibitem{wang2022adamix}
Y.~Wang, S.~Mukherjee, X.~Liu, J.~Gao, A.~H. Awadallah, and J.~Gao, ``Adamix:
  Mixture-of-adapter for parameter-efficient tuning of large language models,''
  \emph{arXiv preprint arXiv:2205.12410}, vol.~1, no.~2, p.~4, 2022.

\bibitem{chen2023sparse}
T.~Chen, Z.~Zhang, A.~Jaiswal, S.~Liu, and Z.~Wang, ``Sparse moe as the new
  dropout: Scaling dense and self-slimmable transformers,'' \emph{arXiv
  preprint arXiv:2303.01610}, 2023.

\bibitem{zhu2022multi}
H.~Zhu, B.~He, and X.~Zhang, ``Multi-gate mixture-of-experts stacked
  autoencoders for quality prediction in blast furnace ironmaking,'' \emph{ACS
  omega}, vol.~7, no.~45, pp. 41\,296--41\,303, 2022.

\bibitem{chi2022representation}
Z.~Chi, L.~Dong, S.~Huang, D.~Dai, S.~Ma, B.~Patra, S.~Singhal, P.~Bajaj,
  X.~Song, X.-L. Mao \emph{et~al.}, ``On the representation collapse of sparse
  mixture of experts,'' \emph{Advances in Neural Information Processing
  Systems}, vol.~35, pp. 34\,600--34\,613, 2022.

\bibitem{gupta2022sparsely}
S.~Gupta, S.~Mukherjee, K.~Subudhi, E.~Gonzalez, D.~Jose, A.~H. Awadallah, and
  J.~Gao, ``Sparsely activated mixture-of-experts are robust multi-task
  learners,'' \emph{arXiv preprint arXiv:2204.07689}, 2022.

\bibitem{dikkala2023benefits}
N.~Dikkala, N.~Ghosh, R.~Meka, R.~Panigrahy, N.~Vyas, and X.~Wang, ``On the
  benefits of learning to route in mixture-of-experts models,'' in
  \emph{Proceedings of the 2023 Conference on Empirical Methods in Natural
  Language Processing}, 2023, pp. 9376--9396.

\bibitem{dryden2022spatial}
N.~Dryden and T.~Hoefler, ``Spatial mixture-of-experts,'' \emph{Advances in
  Neural Information Processing Systems}, vol.~35, pp. 11\,697--11\,713, 2022.

\bibitem{you2022speechmoe2}
Z.~You, S.~Feng, D.~Su, and D.~Yu, ``Speechmoe2: Mixture-of-experts model with
  improved routing,'' in \emph{ICASSP 2022-2022 IEEE International Conference
  on Acoustics, Speech and Signal Processing (ICASSP)}.\hskip 1em plus 0.5em
  minus 0.4em\relax IEEE, 2022, pp. 7217--7221.

\bibitem{puigcerver2022adversarial}
J.~Puigcerver, R.~Jenatton, C.~Riquelme, P.~Awasthi, and S.~Bhojanapalli, ``On
  the adversarial robustness of mixture of experts,'' \emph{Advances in Neural
  Information Processing Systems}, vol.~35, pp. 9660--9671, 2022.

\bibitem{li2023accelerating}
J.~Li, Y.~Jiang, Y.~Zhu, C.~Wang, and H.~Xu, ``Accelerating distributed
  $\{$MoE$\}$ training and inference with lina,'' in \emph{2023 USENIX Annual
  Technical Conference (USENIX ATC 23)}, 2023, pp. 945--959.

\bibitem{wu2022residual}
L.~Wu, M.~Liu, Y.~Chen, D.~Chen, X.~Dai, and L.~Yuan, ``Residual mixture of
  experts,'' \emph{arXiv preprint arXiv:2204.09636}, 2022.

\bibitem{zoph2022designing}
B.~Zoph, I.~Bello, S.~Kumar, N.~Du, Y.~Huang, J.~Dean, N.~Shazeer, and
  W.~Fedus, ``Designing effective sparse expert models,'' \emph{arXiv preprint
  arXiv:2202.08906}, vol.~2, 2022.

\bibitem{zoph2022st}
------, ``St-moe: Designing stable and transferable sparse expert models,''
  \emph{arXiv preprint arXiv:2202.08906}, 2022.

\bibitem{chow2022mixture}
Y.~Chow, A.~Tulepbergenov, O.~Nachum, M.~Ryu, M.~Ghavamzadeh, and C.~Boutilier,
  ``A mixture-of-expert approach to rl-based dialogue management,'' \emph{arXiv
  preprint arXiv:2206.00059}, 2022.

\bibitem{fan2022m3vit}
Z.~Fan, R.~Sarkar, Z.~Jiang, T.~Chen, K.~Zou, Y.~Cheng, C.~Hao, Z.~Wang
  \emph{et~al.}, ``M$^3$vit: Mixture-of-experts vision transformer for
  efficient multi-task learning with model-accelerator co-design,''
  \emph{Advances in Neural Information Processing Systems}, vol.~35, pp.
  28\,441--28\,457, 2022.

\bibitem{zadouri2023pushing}
T.~Zadouri, A.~{\"U}st{\"u}n, A.~Ahmadian, B.~Ermi{\c{s}}, A.~Locatelli, and
  S.~Hooker, ``Pushing mixture of experts to the limit: Extremely parameter
  efficient moe for instruction tuning,'' \emph{arXiv preprint
  arXiv:2309.05444}, 2023.

\bibitem{zhu2022uni}
J.~Zhu, X.~Zhu, W.~Wang, X.~Wang, H.~Li, X.~Wang, and J.~Dai,
  ``Uni-perceiver-moe: Learning sparse generalist models with conditional
  moes,'' \emph{Advances in Neural Information Processing Systems}, vol.~35,
  pp. 2664--2678, 2022.

\bibitem{dou2023towards}
F.~Dou, J.~Ye, G.~Yuan, Q.~Lu, W.~Niu, H.~Sun, L.~Guan, G.~Lu, G.~Mai, N.~Liu
  \emph{et~al.}, ``Towards artificial general intelligence (agi) in the
  internet of things (iot): Opportunities and challenges,'' \emph{arXiv
  preprint arXiv:2309.07438}, 2023.

\bibitem{jia2022improving}
Z.~Jia, X.~Li, Z.~Ling, S.~Liu, Y.~Wu, and H.~Su, ``Improving policy
  optimization with generalist-specialist learning,'' in \emph{International
  Conference on Machine Learning}.\hskip 1em plus 0.5em minus 0.4em\relax PMLR,
  2022, pp. 10\,104--10\,119.

\bibitem{simeone2022unknown}
M.~Simeone, ``Unknown future, repeated present: A narrative-centered analysis
  of long-term ai discourse,'' \emph{Humanist Studies \& the Digital Age},
  vol.~7, no.~1, 2022.

\bibitem{nair2022bridging}
A.~Nair and F.~Banaei-Kashani, ``Bridging the gap between artificial
  intelligence and artificial general intelligence: A ten commandment framework
  for human-like intelligence,'' \emph{arXiv preprint arXiv:2210.09366}, 2022.

\bibitem{jarrahi2023artificial}
M.~H. Jarrahi, D.~Askay, A.~Eshraghi, and P.~Smith, ``Artificial intelligence
  and knowledge management: A partnership between human and ai,''
  \emph{Business Horizons}, vol.~66, no.~1, pp. 87--99, 2023.

\bibitem{edwards2022functional}
D.~J. Edwards, C.~McEnteggart, and Y.~Barnes-Holmes, ``A functional contextual
  account of background knowledge in categorization: Implications for
  artificial general intelligence and cognitive accounts of general
  knowledge,'' \emph{Frontiers in Psychology}, vol.~13, p. 745306, 2022.

\bibitem{mccarthy2022artificial}
J.~McCarthy, ``Artificial intelligence, logic, and formalising common sense,''
  \emph{Machine Learning and the City: Applications in Architecture and Urban
  Design}, pp. 69--90, 2022.

\bibitem{friederich2023symbiosis}
S.~Friederich, ``Symbiosis, not alignment, as the goal for liberal democracies
  in the transition to artificial general intelligence,'' \emph{AI and Ethics},
  pp. 1--10, 2023.

\bibitem{makridakis2017forthcoming}
S.~Makridakis, ``The forthcoming artificial intelligence (ai) revolution: Its
  impact on society and firms,'' \emph{Futures}, vol.~90, pp. 46--60, 2017.

\bibitem{pal2023ai}
S.~Pal, K.~Kumari, S.~Kadam, and A.~Saha, ``The ai revolution,'' \emph{IARA
  Publication}, 2023.

\bibitem{verma2021artificial}
S.~Verma, R.~Sharma, S.~Deb, and D.~Maitra, ``Artificial intelligence in
  marketing: Systematic review and future research direction,''
  \emph{International Journal of Information Management Data Insights}, vol.~1,
  no.~1, p. 100002, 2021.

\bibitem{budhwar2023human}
P.~Budhwar, S.~Chowdhury, G.~Wood, H.~Aguinis, G.~J. Bamber, J.~R. Beltran,
  P.~Boselie, F.~Lee~Cooke, S.~Decker, A.~DeNisi \emph{et~al.}, ``Human
  resource management in the age of generative artificial intelligence:
  Perspectives and research directions on chatgpt,'' \emph{Human Resource
  Management Journal}, vol.~33, no.~3, pp. 606--659, 2023.

\bibitem{telkamp2022implications}
J.~B. Telkamp and M.~H. Anderson, ``The implications of diverse human moral
  foundations for assessing the ethicality of artificial intelligence,''
  \emph{Journal of Business Ethics}, vol. 178, no.~4, pp. 961--976, 2022.

\bibitem{zhou2023interpretable}
X.~Zhou, C.~Liu, L.~Zhai, Z.~Jia, C.~Guan, and Y.~Liu, ``Interpretable and
  robust ai in eeg systems: A survey,'' \emph{arXiv preprint arXiv:2304.10755},
  2023.

\bibitem{zhang2023one}
C.~Zhang, C.~Zhang, C.~Li, Y.~Qiao, S.~Zheng, S.~K. Dam, M.~Zhang, J.~U. Kim,
  S.~T. Kim, J.~Choi \emph{et~al.}, ``One small step for generative ai, one
  giant leap for agi: A complete survey on chatgpt in aigc era,'' \emph{arXiv
  preprint arXiv:2304.06488}, 2023.

\bibitem{singhal2023towards}
K.~Singhal, T.~Tu, J.~Gottweis, R.~Sayres, E.~Wulczyn, L.~Hou, K.~Clark,
  S.~Pfohl, H.~Cole-Lewis, D.~Neal \emph{et~al.}, ``Towards expert-level
  medical question answering with large language models,'' \emph{arXiv preprint
  arXiv:2305.09617}, 2023.

\bibitem{wu2023bloomberggpt}
S.~Wu, O.~Irsoy, S.~Lu, V.~Dabravolski, M.~Dredze, S.~Gehrmann, P.~Kambadur,
  D.~Rosenberg, and G.~Mann, ``Bloomberggpt: A large language model for
  finance,'' \emph{arXiv preprint arXiv:2303.17564}, 2023.

\bibitem{henderson2018ethical}
P.~Henderson, K.~Sinha, N.~Angelard-Gontier, N.~R. Ke, G.~Fried, R.~Lowe, and
  J.~Pineau, ``Ethical challenges in data-driven dialogue systems,'' in
  \emph{Proceedings of the 2018 AAAI/ACM Conference on AI, Ethics, and
  Society}, 2018, pp. 123--129.

\bibitem{bin2023use}
S.~A. Bin-Nashwan, M.~Sadallah, and M.~Bouteraa, ``Use of chatgpt in academia:
  Academic integrity hangs in the balance,'' \emph{Technology in Society},
  vol.~75, p. 102370, 2023.

\bibitem{liu2023ai}
N.~Liu, A.~Brown \emph{et~al.}, ``Ai increases the pressure to overhaul the
  scientific peer review process. comment on “artificial intelligence can
  generate fraudulent but authentic-looking scientific medical articles:
  Pandora’s box has been opened”,'' \emph{J Med Internet Res}, vol.~25, p.
  e50591, 2023.

\bibitem{siddaway2019systematic}
A.~P. Siddaway, A.~M. Wood, and L.~V. Hedges, ``How to do a systematic review:
  a best practice guide for conducting and reporting narrative reviews,
  meta-analyses, and meta-syntheses,'' \emph{Annual review of psychology},
  vol.~70, pp. 747--770, 2019.

\bibitem{landhuis2016scientific}
E.~Landhuis, ``Scientific literature: Information overload,'' \emph{Nature},
  vol. 535, no. 7612, pp. 457--458, 2016.

\bibitem{chloros2022peer}
G.~D. Chloros, V.~P. Giannoudis, and P.~V. Giannoudis, ``Peer-reviewing in
  surgical journals: revolutionize or perish?'' \emph{Annals of surgery}, vol.
  275, no.~1, pp. e82--e90, 2022.

\bibitem{allen2022towards}
K.-A. Allen, J.~Reardon, Y.~Lu, D.~V. Smith, E.~Rainsford, and L.~Walsh,
  ``Towards improving peer review: Crowd-sourced insights from twitter,''
  \emph{Journal of university teaching \& learning practice}, vol.~19, no.~3,
  p.~02, 2022.

\end{thebibliography}

\vfill

%
%




\end{document}